\documentclass[lettersize,journal]{IEEEtran}
\usepackage{amsmath,amsfonts}
\usepackage{algorithmic}
\usepackage{algorithm}
\usepackage{array}
\usepackage[caption=false,font=normalsize,labelfont=sf,textfont=sf]{subfig}
\usepackage{textcomp}
\usepackage{stfloats}
\usepackage{url}
\usepackage{verbatim}
\usepackage{graphicx}
\usepackage{cite}
\usepackage{algorithm}
\usepackage{algorithmic}
\usepackage{multirow}
\usepackage{lineno}
\usepackage{amsmath}
\usepackage{multirow}

\begin{document}

\title{Learning Multiple Representations with Inconsistency-Guided Detail Regularization for Mask-Guided Matting}

\author{Weihao Jiang, Zhaozhi Xie, Yuxiang Lu,~\IEEEmembership{Student Member,~IEEE,} Longjie Qi, Jingyong Cai,  Hiroyuki Uchiyama, Bin Chen, Yue Ding,~\IEEEmembership{Member,~IEEE,} and Hongtao Lu,~\IEEEmembership{Member,~IEEE}  
\thanks{Weihao Jiang, Zhaozhi Xie, Yuxiang Lu, Longjie Qi, Yue Ding, and Hongtao Lu are with the Department of Computer Science and Engineering, Shanghai Jiao Tong
University, Shanghai 200240, China (email: jiangweihao@sjtu.edu.cn; xiezhzh@sjtu.edu.cn; luyuxiang\_2018@sjtu.edu.cn; qilongjie@sjtu.edu.cn; dingyue@sjtu.edu.cn; htlu@sjtu.edu.cn)\\
Jingyong Cai, Hiroyuki Uchiyama, and  Bin Chen are with the OPPO Japan Research Center, Japan }
}

\markboth{Journal of \LaTeX\ Class Files,~Vol.~14, No.~8, August~2021}%
{Shell \MakeLowercase{\textit{et al.}}: A Sample Article Using IEEEtran.cls for IEEE Journals}


\maketitle

\begin{abstract}
Mask-guided matting networks  have achieved significant improvements and have shown great potential in practical applications in recent years.  However, simply learning matting representation from synthetic and lack-of-real-world-diversity matting data, these approaches  tend to overfit low-level details in wrong regions, lack generalization to objects with complex structures and real-world scenes such as shadows, as well as suffer from interference of background lines or textures. To address these challenges, in this paper, we propose a novel auxiliary learning framework for mask-guided matting models, incorporating  three auxiliary tasks: semantic segmentation, edge detection, and background line detection besides matting, to learn different and effective representations from different types of data and annotations. Our framework and model introduce the following key aspects: (1) to learn real-world adaptive semantic representation for objects with diverse and complex structures under real-world scenes, we introduce extra semantic segmentation and edge detection tasks on more diverse real-world data with segmentation annotations; (2) to avoid overfitting on low-level details, we propose a module to utilize the inconsistency between learned segmentation and matting representations to regularize detail refinement; (3) we propose a novel background line detection task into our auxiliary learning framework, to suppress interference of background lines or textures. In addition, we propose a high-quality matting benchmark,  Plant-Mat, to evaluate matting methods on complex structures. Extensively quantitative and qualitative results show that our approach outperforms state-of-the-art mask-guided methods.  
\end{abstract}


\section{Introduction}
\IEEEPARstart{I}{mage}  alpha matting is an important computer vision task that predicts an alpha matte representing the opacity of foreground objects to precisely cut them out in an image. It has many applications in computational photography and image or video processing,  editing, and compositing~\cite{wang2008image,deepmatting,tmm_unscreen,tmm_render,tmm_recomp,zitnick2007stereo}. Alpha matting tasks usually model the natural image $I$ as a convex
combination of a foreground image $F$ and a background image $B$ at each pixel $i$, as shown below:
\begin{equation}
	I_i ={\alpha}_i F_i + (1-{\alpha}_i) B_i, {\alpha}_i \in [0,1],
	\label{alphaeq}
\end{equation}
where ${\alpha}_i$ is the value of the alpha matte at pixel $i$. 

\begin{figure}[t]
  \centering
  \footnotesize
  \setlength{\tabcolsep}{0.0pt} 
  
  \begin{tabular}{ccc}
      \includegraphics[trim={0 0cm 0 2.65cm},clip,width=0.15\textwidth]{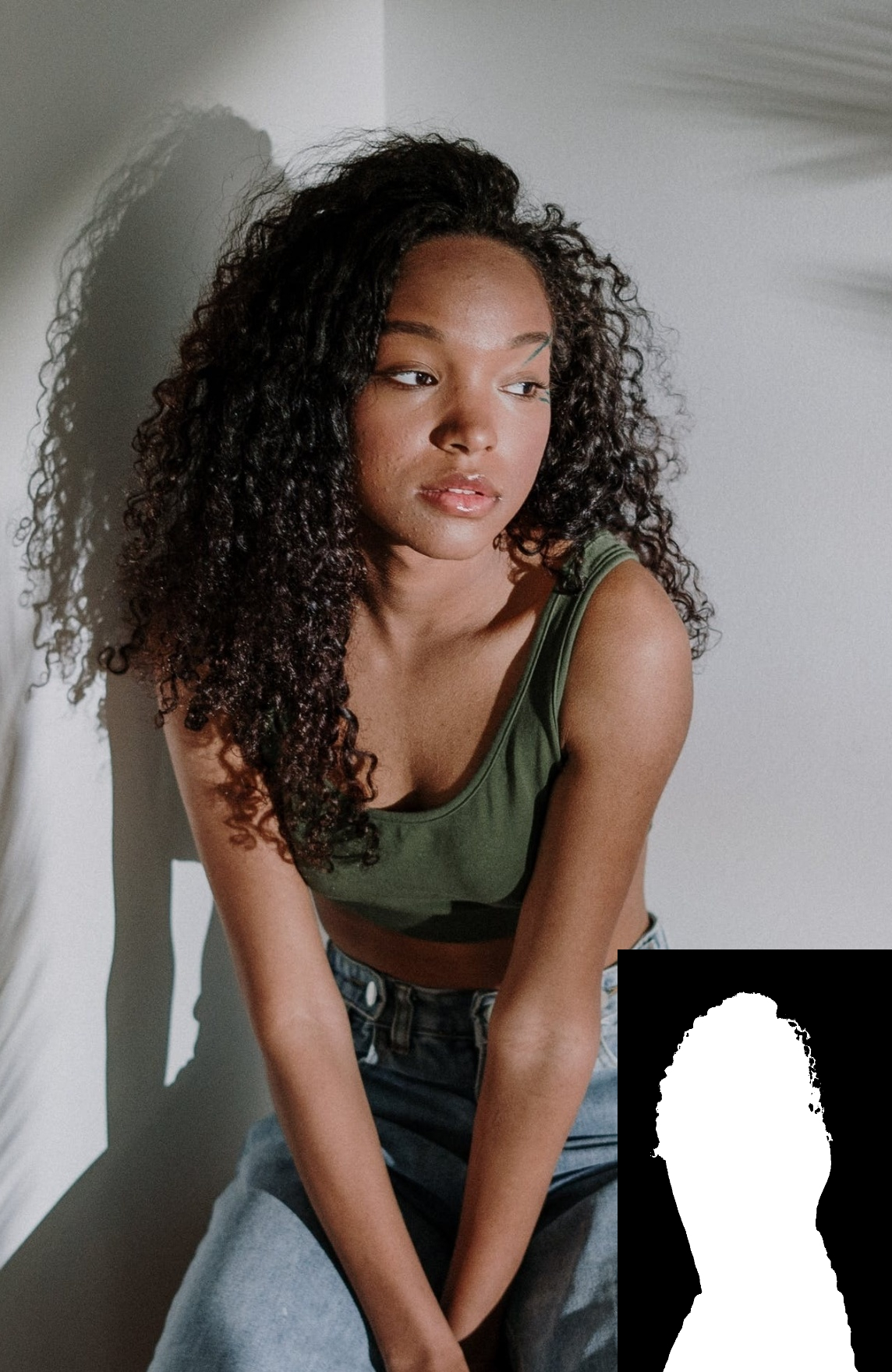} &
      \includegraphics[trim={0 0cm 0 2.65cm},clip,width=0.15\textwidth]{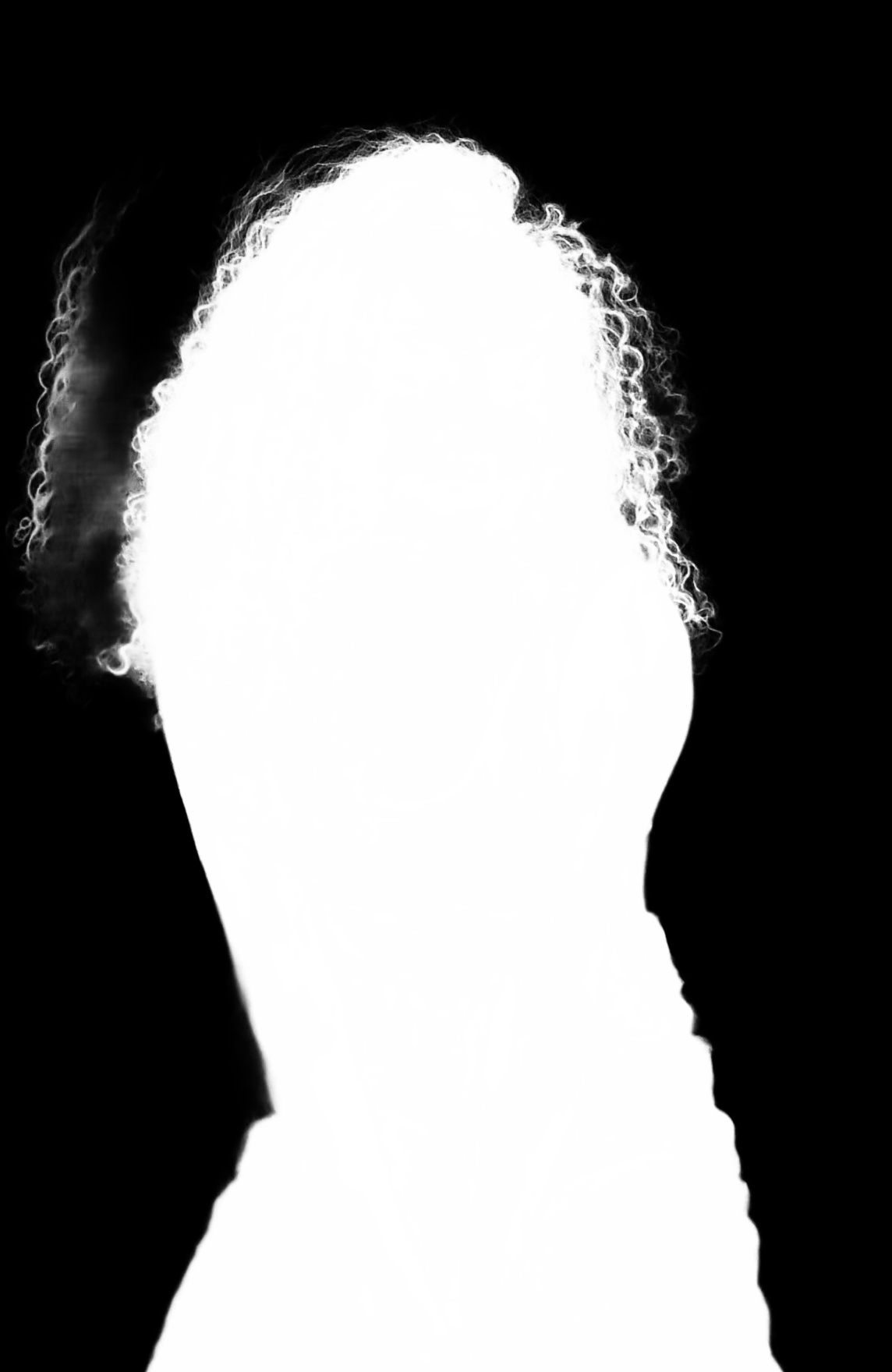} &
      \includegraphics[trim={0 0cm 0 2.65cm},clip,width=0.15\textwidth]{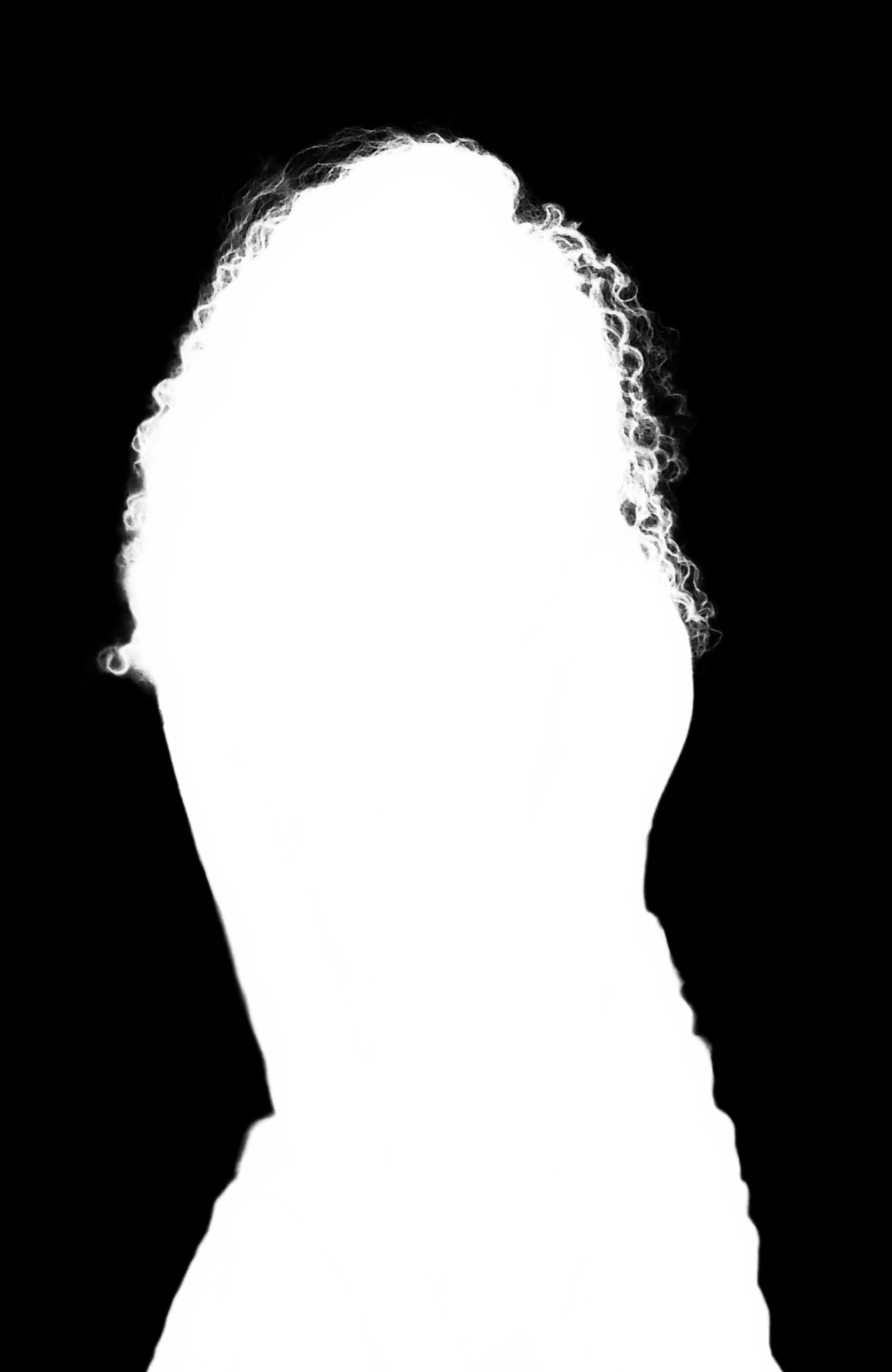} \\
      
      \includegraphics[width=0.15\textwidth]{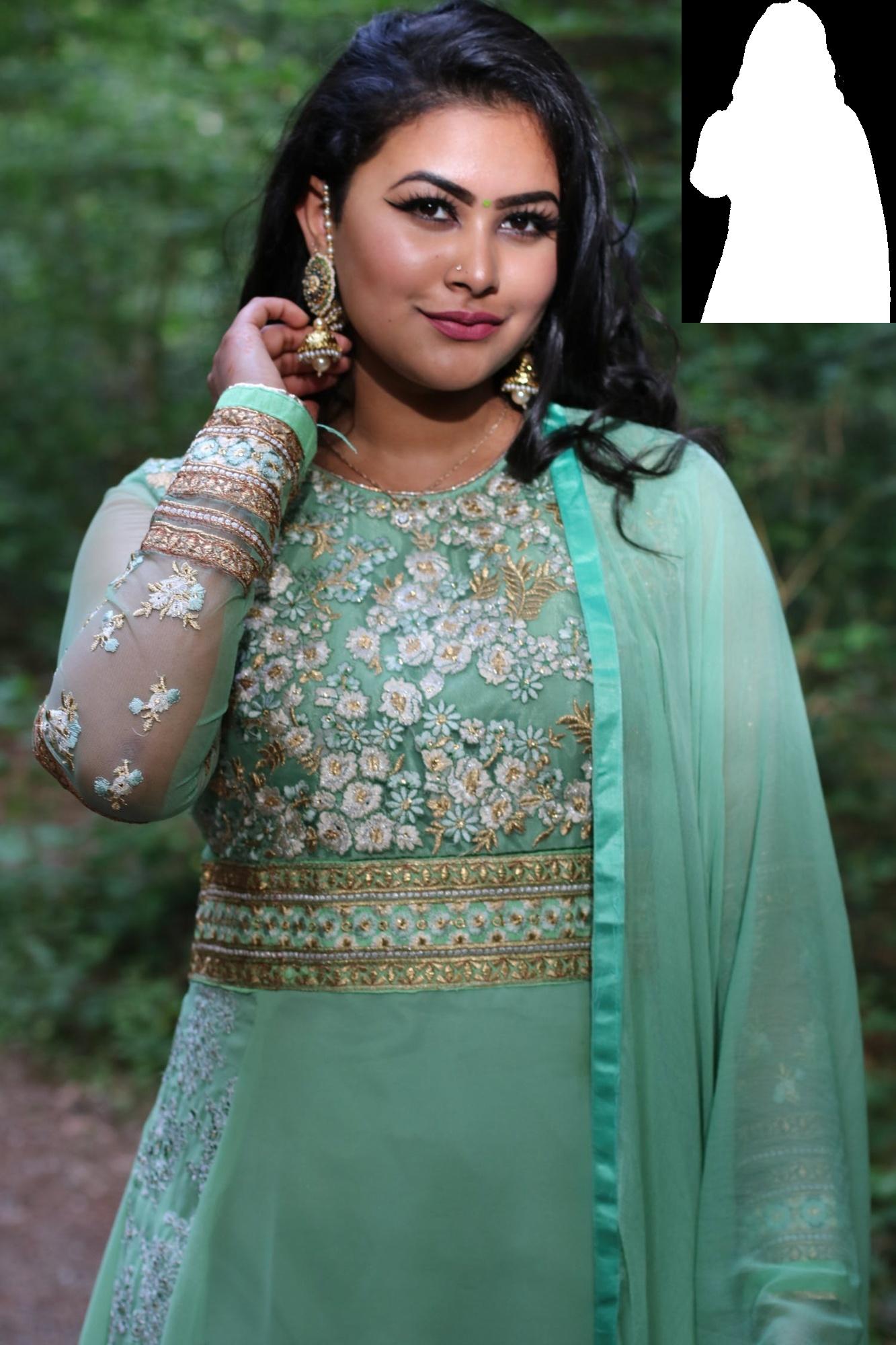} &
      \includegraphics[width=0.15\textwidth]{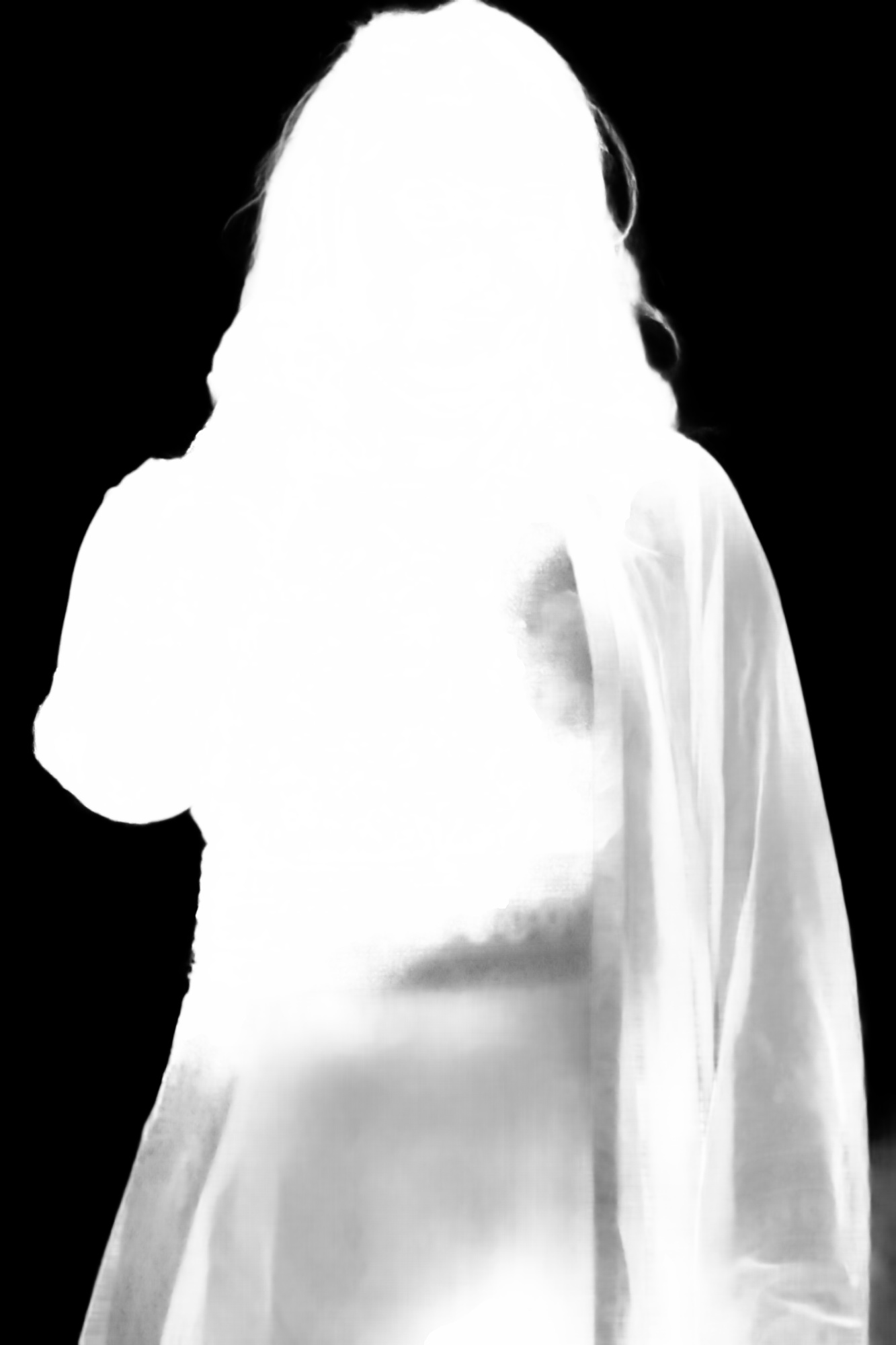} &
      \includegraphics[width=0.15\textwidth]{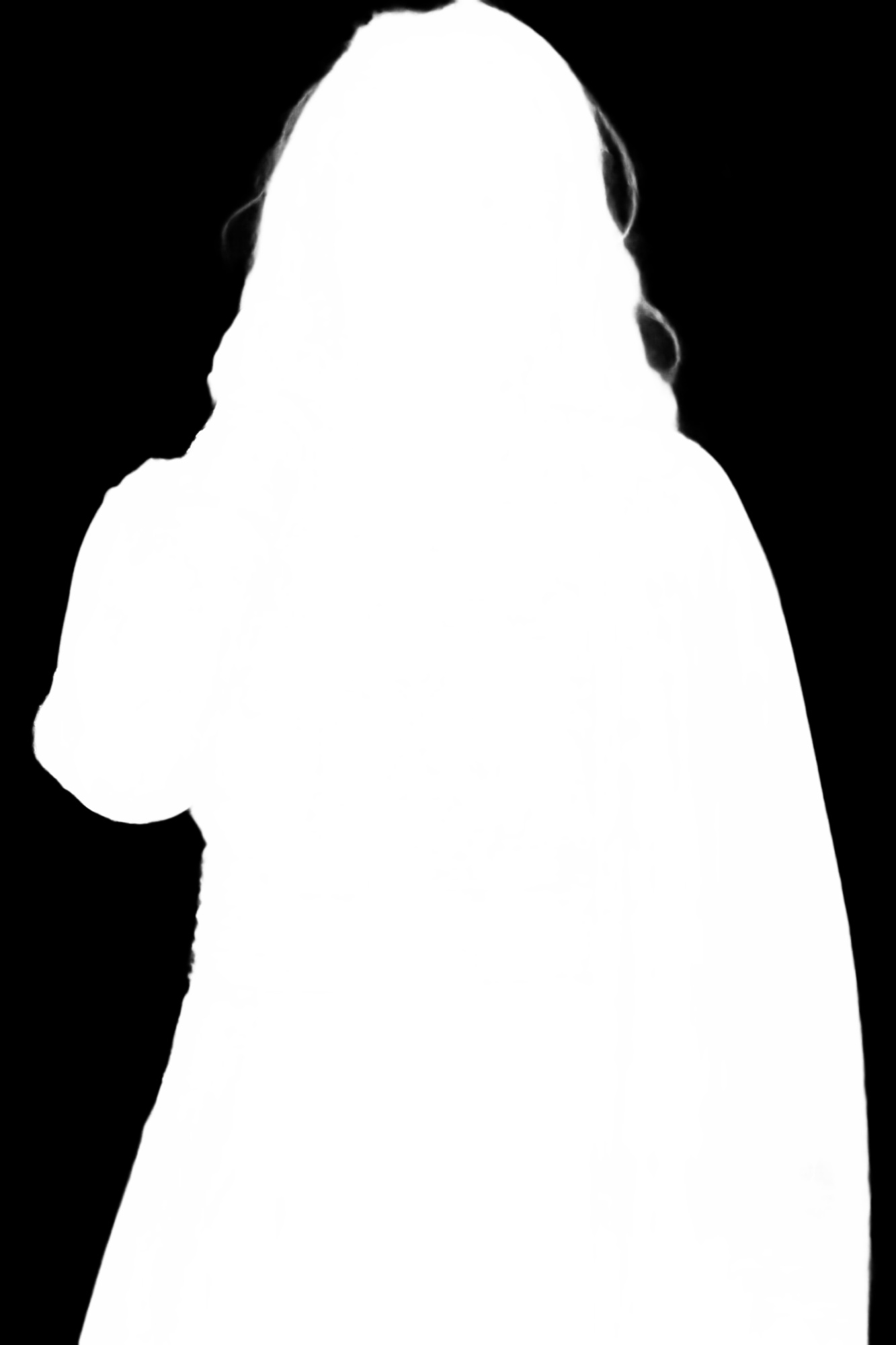} \\
      
      \includegraphics[width=0.15\textwidth]{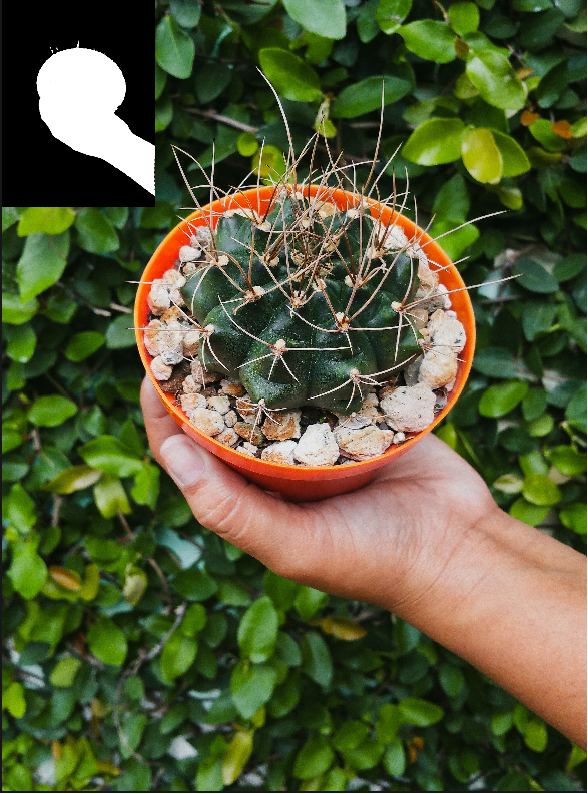} &
      \includegraphics[width=0.15\textwidth]{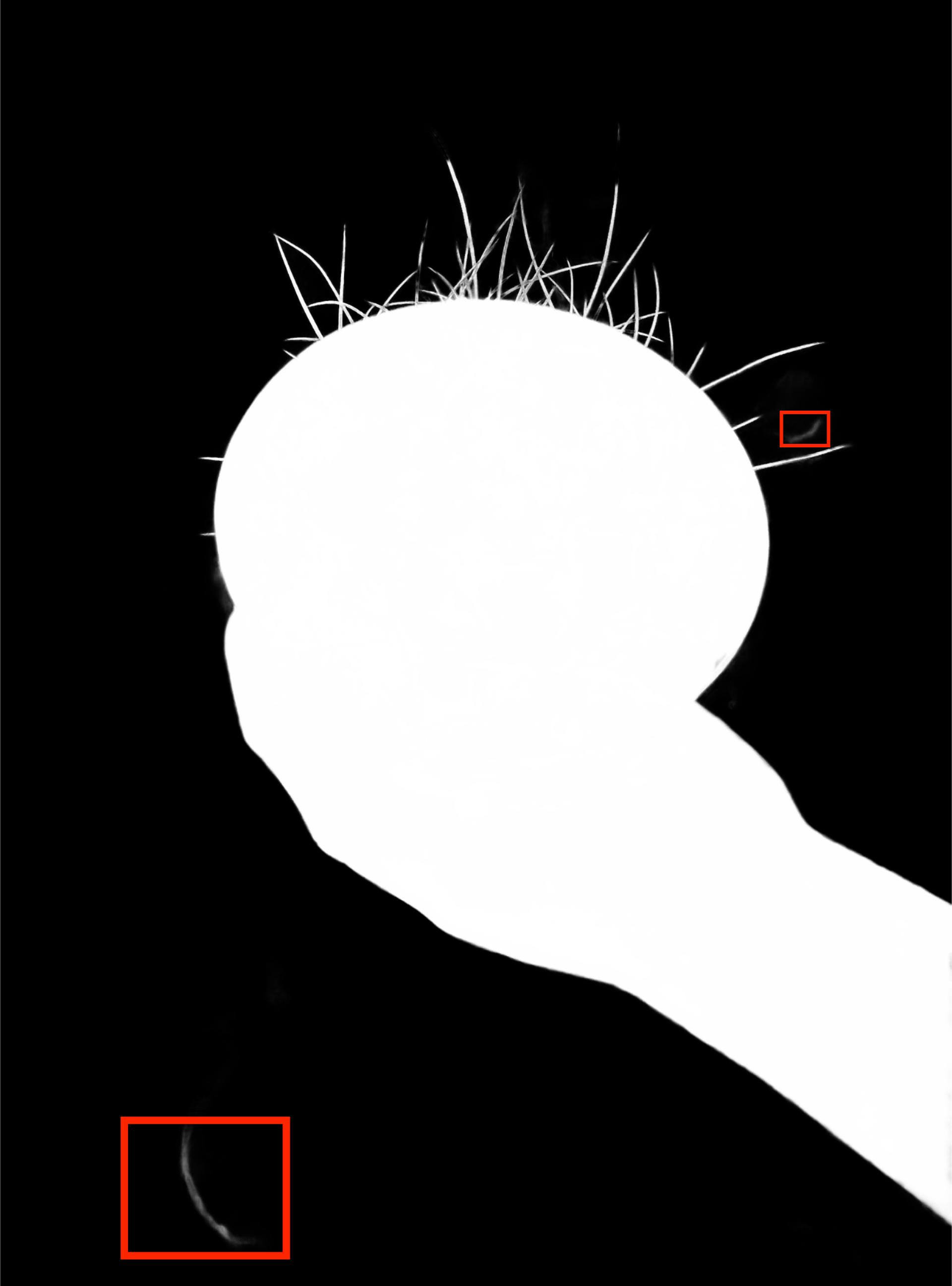} &
      \includegraphics[width=0.15\textwidth]{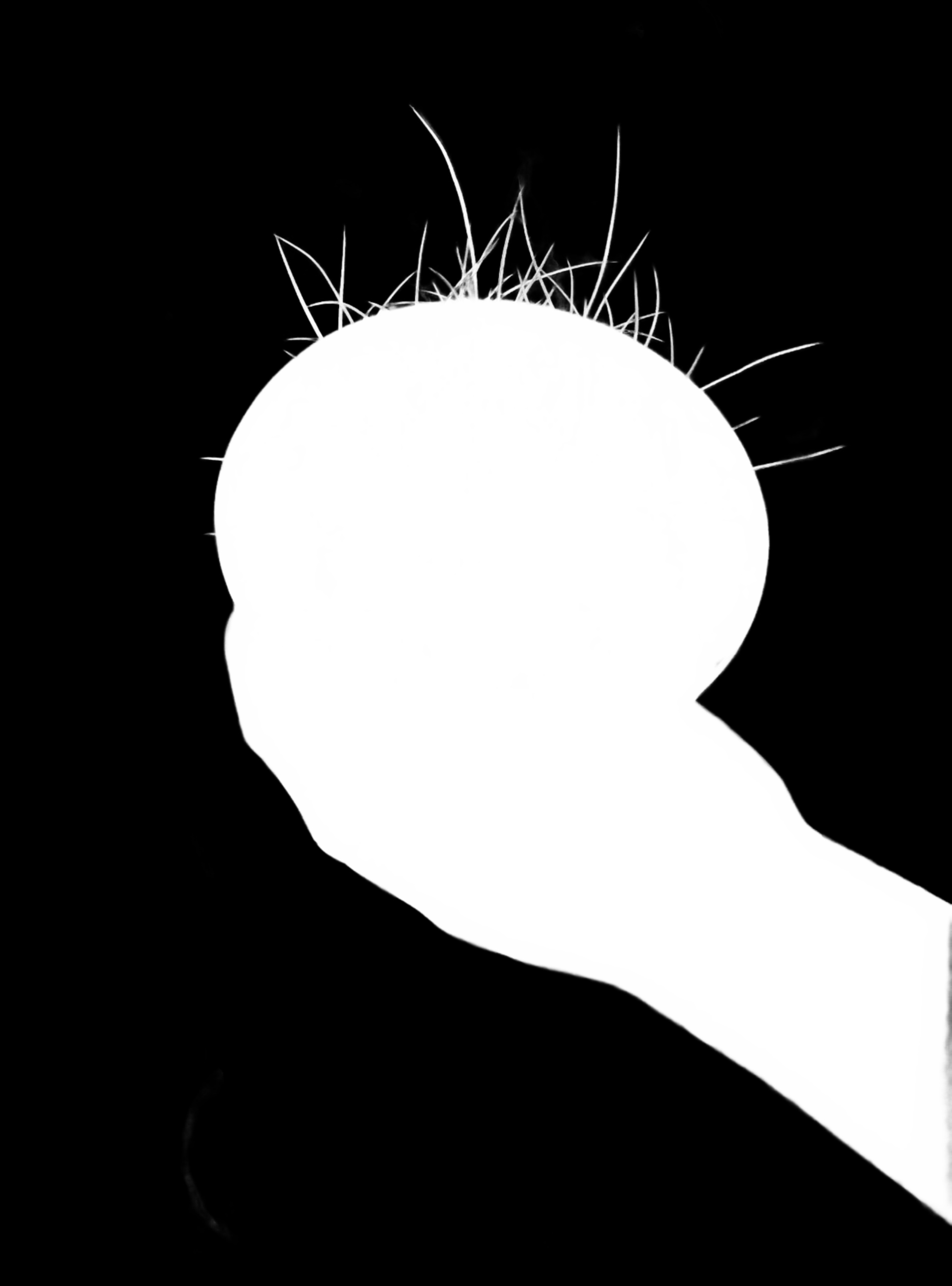} \\
    \end{tabular}
    \caption{Qualitative comparisons between MGMatting~\cite{mgm_ref} and Ours. From left to right, the input image and a binary guidance mask, MGMatting, Ours.}
    \label{fig:illu_vis}
  \vspace{-2.5em} 
\end{figure}

The Eq.~\ref{alphaeq} is highly ill-posed~\cite{closed_24}. Therefore, many traditional methods~\cite{bayes_12,opt,global_18,shared_mat,closed_24,knn_10,large_kernel} and deep learning based methods~\cite{pii,gca,a2u,matteformer_ref,sim,adamatting,fba,lfpnet,sim_ijcv,hdmatt,samplenet,alphagan,rmat_ref} utilize trimaps  as guidance to reduce the solution space. In recent years, methods like   MGMatting~\cite{mgm_ref} and IGF~\cite{igf_ref} propose  mask-guided matting frameworks, which only need an easily obtained coarse mask instead of a complex trimap. Since fine matting data is labor-intensive in data selection and annotation, deep matting methods composite finely annotated foreground on various background images to train models for objects under diverse scenes. However, these training samples still lack real-world diversity. Although MGMatting has an elaborate network and uses a strong training data augmentation like Context-aware matting~\cite{context} did on composited data to adapt the real-world application, and promote the mask-guided task, it still suffers from a few problems. Due to composited data or hard-to-attain and lack-of-real-world-diversity data, MGMatting is hard to generalize to real-world scenes such as shadows shown in Row 1, Fig.~\ref{fig:illu_vis} and complex real-world foreground structures such as elongated cactus spines shown in Row 3, Fig~\ref{fig:illu_vis}. Due to overfitting on low-level details, MGMatting refines low-level details in the wrong regions (SARI's textures on a woman's body) instead of the correct sparse hairs in Row 2, Fig.~\ref{fig:illu_vis}. Last but not least, MGMatting also struggles to suppress interference of background lines or textures such as the two red boxes in Row 3, Fig.~\ref{fig:illu_vis}.

To address these challenges for mask-guided matting, we propose a novel auxiliary learning framework to properly learn real-world adaptive semantic representation and background line aware representation from different types of data (composited and real-world) and annotations (matting, segmentation, background line), and propose an inconsistency-guided detail regularization module to regularize detail refinement.

First, to adapt to diverse and complex object structures in real world, we introduce a real-world adaptive semantic representation to mask-guided networks through auxiliary semantic segmentation learning on diverse real-world data. Although the composited matting data provides precise alpha mattes strictly following Eq.~\ref{alphaeq} and can train matting models with detailed predictions, the real-world data provides real scenes such as shadows that can not be provided by the composited data. As the real-world data with coarse segmentation masks is much easier to attain than data with fine matting alpha mattes, it can provide training data with diverse and complex objects under real-world scenes. Different from matting alpha mattes, semantic segmentation masks focus more on representing the high-level semantic foreground regions instead of low-level details. Therefore, instead of naively using the segmentation data to supervise matting heads,  we set an extra segmentation head on proper high-level feature maps in our matting network to learn the high-level semantics of real-world objects. In addition, we also add an extra edge detection head on proper high-resolution feature maps to learn real-world object boundaries and semantic contours. In this way, our network learns real-world adaptive semantic representation $Se$  for diverse and complex real-world objects, besides the matting representation $Ma$ with richer low-level details learned from the matting task, and adapts better to diverse and complex structures and real-world scenes. 
\begin{figure}[t]
  \centering
  \footnotesize
  \setlength{\tabcolsep}{0.0pt} 

  \begin{tabular}{cccc}
    \includegraphics[trim={0 2.55cm 0 0cm},clip,width=0.112\textwidth]{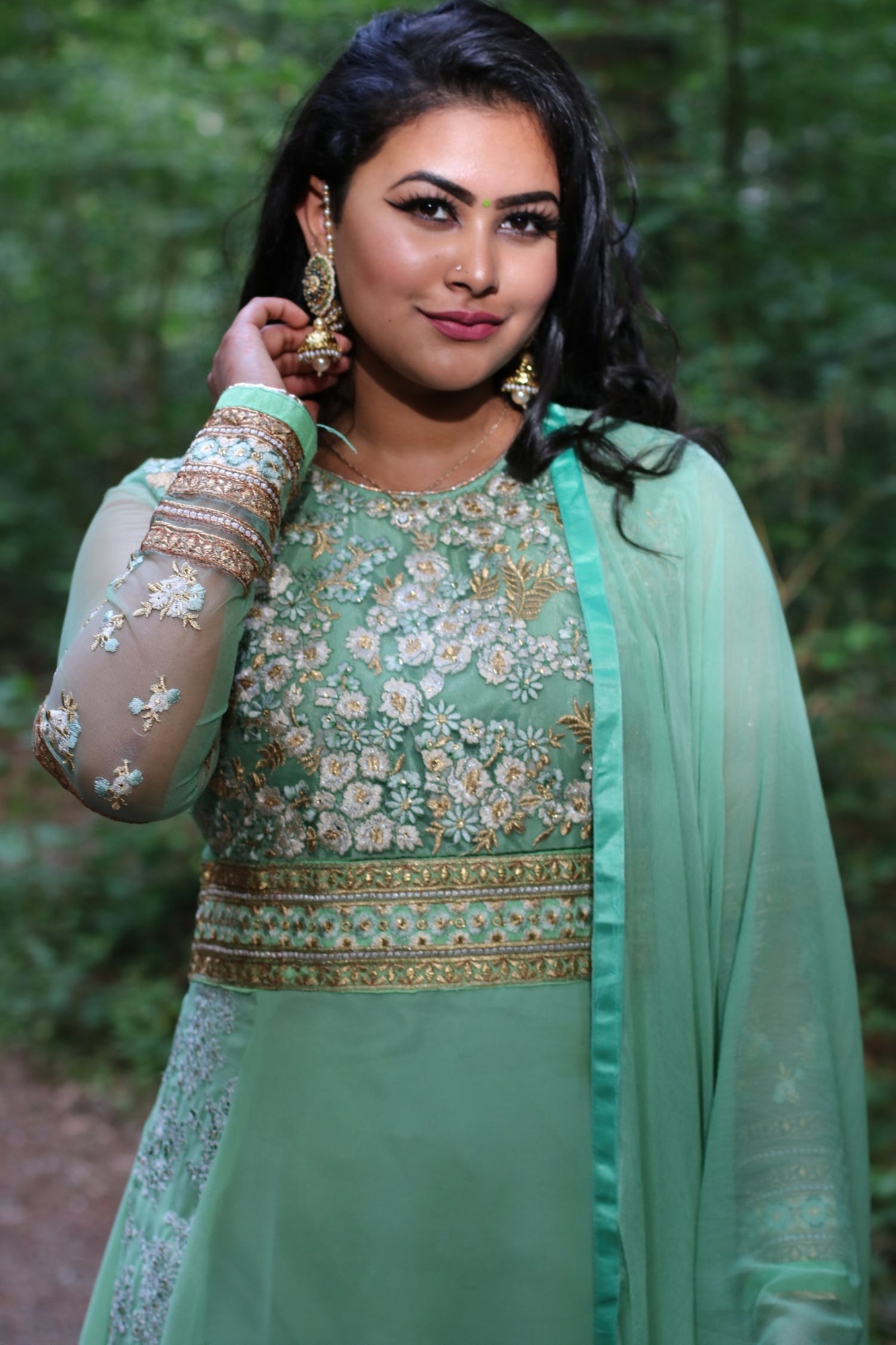} &
    \includegraphics[trim={0 2.55cm 0 0cm},clip,width=0.112\textwidth]{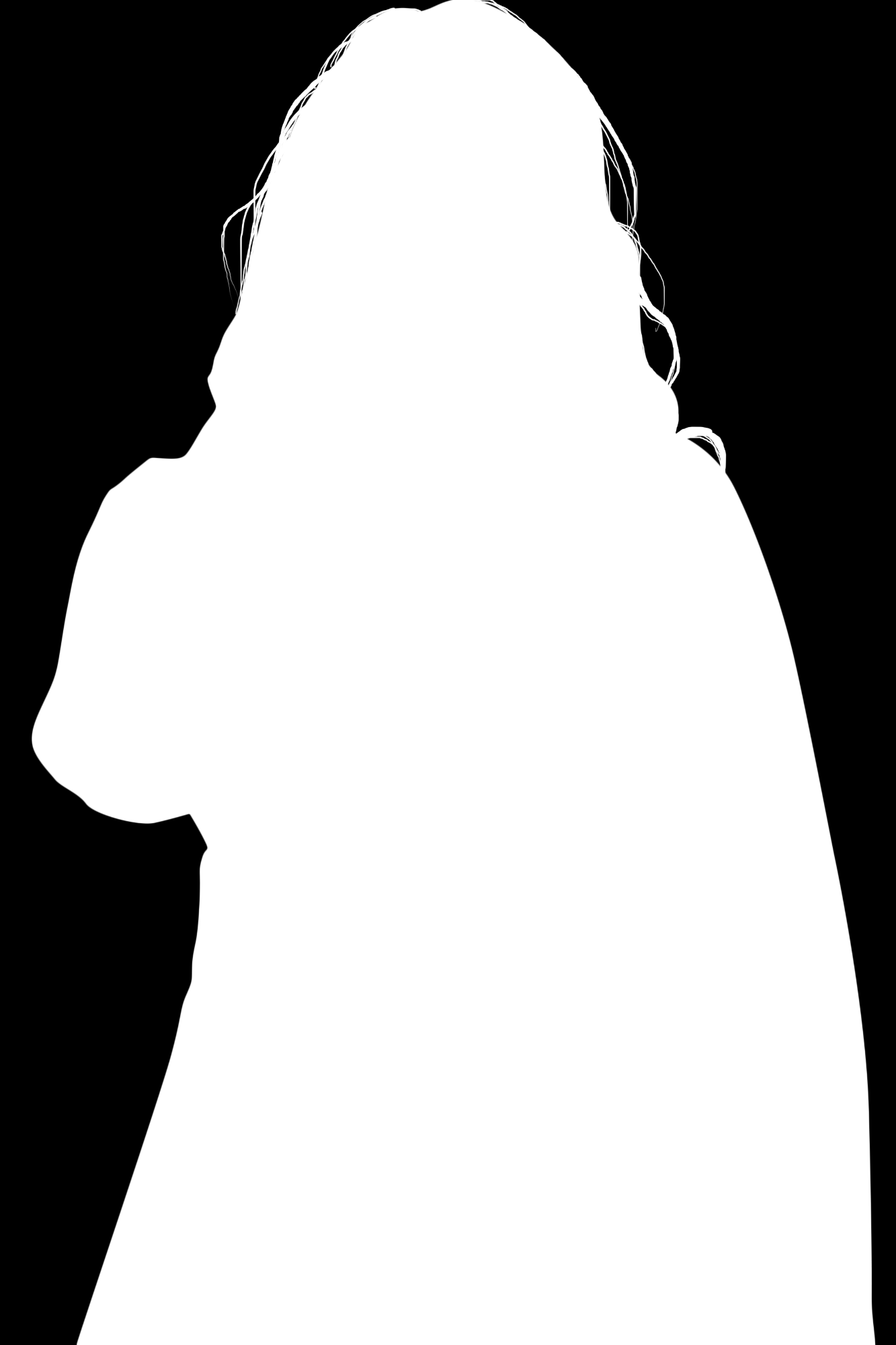} &
    \includegraphics[trim={0 2.55cm 0 0cm},clip,width=0.112\textwidth]{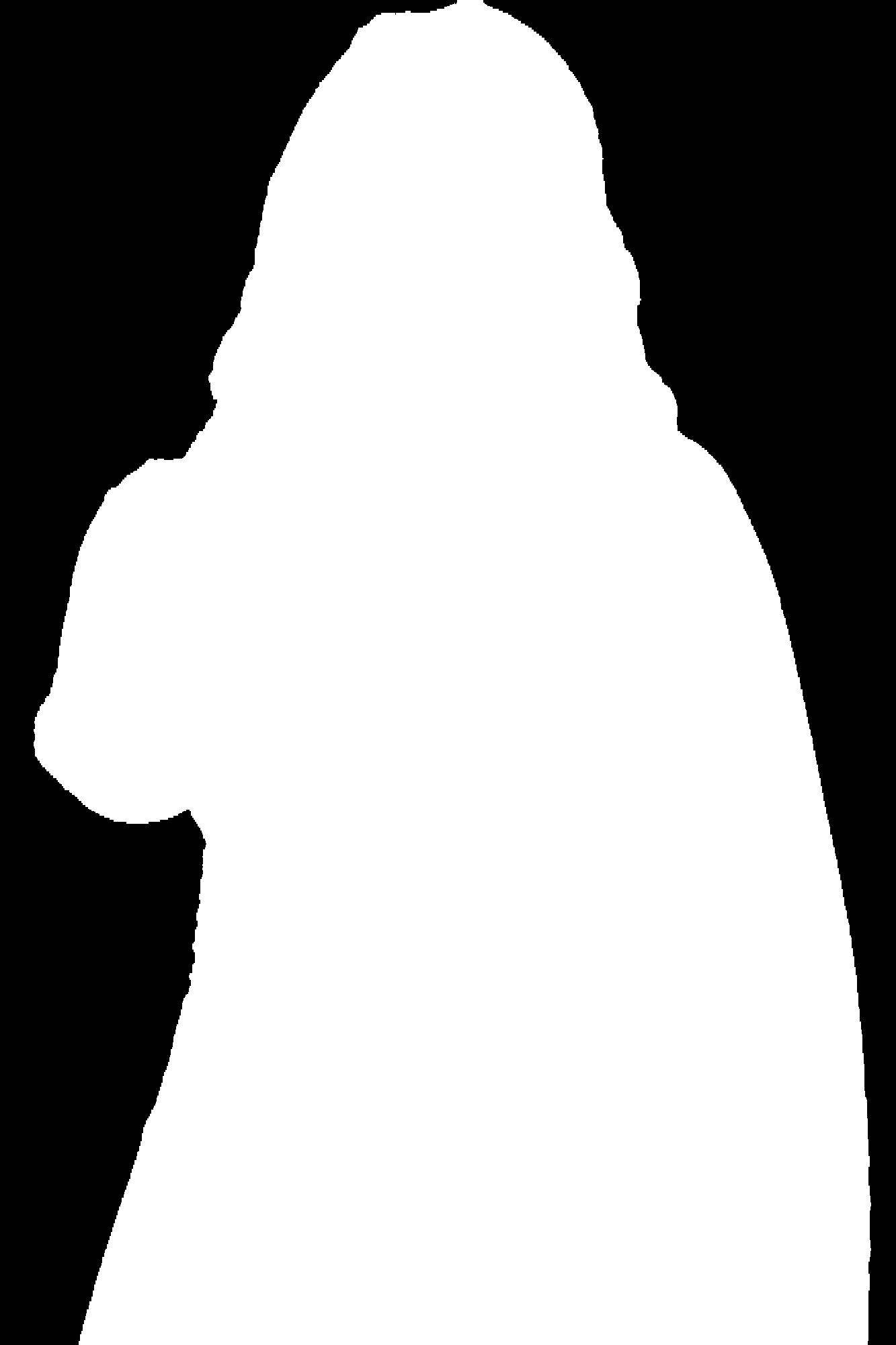} &
    \includegraphics[trim={0 2.55cm 0 0cm},clip,width=0.112\textwidth]{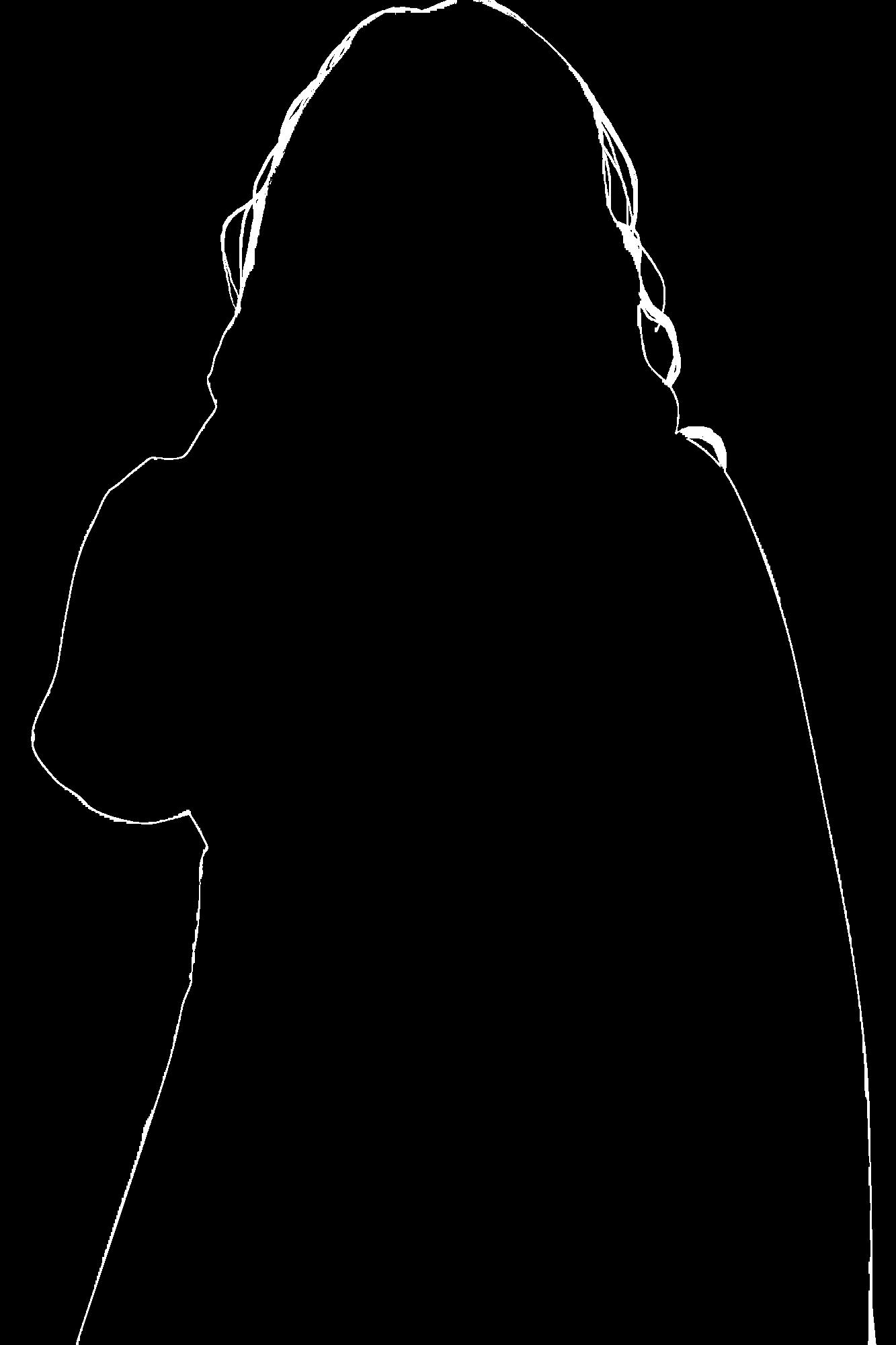} \\

    Image & Matting & Segmentation & Inconsistency
  \end{tabular}
  
  \caption{Visualization of the inconsistency between matting and segmentation masks, which points out important low-level details.}
  \label{fig:ig}
   \vspace{-1.5em}
\end{figure}

Second, we propose an inconsistency-guided detail regularization (IGDR) module to regularize detail refinement and avoid overfitting low-level details.
Generally, matting tasks represent a foreground object with a fine and soft mask, while segmentation tasks represent semantic foregrounds with a binary mask. 
As shown in Fig.~\ref{fig:ig}, it is easy to observe that  inconsistent parts between a matting alpha matte and segmentation mask highlight  objects' fine low-level details which should be refined for matting.  Based on this observation, and inspired by spatial sampling or alignment works~\cite{dcn,DFF_ref,sfnet_ref}, we generate a semantic representation feature map $Se$ from a matting representation feature map $Ma$ by eliminating their spatial inconsistency using a spatial sampling process. Then $Ma$ and $Se$ will be fed to a matting head and a segmentation head, respectively,  for different supervision tasks. Then we obtain  the inconsistency map $IN$ as $IN=Ma-Se$, which indicates low-level details in proper regions, and our IGDR module uses $IN$ to guide and enhance low-level details of objects in proper regions, preventing overfitting in wrong regions like MGMatting~\cite{mgm_ref} in Row 2, Fig.~\ref{fig:illu_vis}.

Third, we propose a novel background line detection task into our auxiliary learning framework, to suppress interference of background lines or textures. Mask-guided matting networks simply learning a matting representation suffer from interference of background details such as lines and textures like MGMatting~\cite{mgm_ref} in Row 3, Fig.~\ref{fig:illu_vis}. Therefore, we incorporate a novel background line detection task into our auxiliary learning framework to learn discriminative representation to better distinguish foreground objects from background lines or textures. For training data, we generate distance maps with LSD~\cite{lsd_ref} for  background images through a homography adaptation~\cite{deeplsd_ref} as pseudo ground truth, then the background images will be composited with  matting foregrounds based on their alpha matte.
For the model, we set a background line detection head on proper high-resolution feature maps in our network decoder,  and supervise it with the distance maps of background lines adapted to corresponding composited samples, to learn a background line aware representation. In this way,  our network learns an effective representation to suppress interference of  background lines or textures.

In addition, we propose a high-quality matting benchmark, Plant-Mat, to evaluate mask-guided matting methods on complex object structures for academic research.

Our contributions can be summarized as follows:
\begin{itemize}
    \item We propose a real-world adaptive semantic representation (RASR) learned through auxiliary semantic segmentation and edge detection tasks on real-world segmentation data, to adapt our network to diverse and complex object structures and real-world scenes.
    \item To overcome the overfitting on low-level details of mask-guided approaches, we propose a novel inconsistency-guided detail regularization
(IGDR) module in our  network to regularize low-level detail refinement.
    \item We propose a novel background line detection task
into our auxiliary learning framework, to suppress interference
of background lines or textures for matting.
    \item Quantitative and qualitative results on the RWP~\cite{mgm_ref}, AIM-500~\cite{aim}, AM-2k~\cite{AM-2k}, PPM-100~\cite{modnet_ref}, and the proposed Plant-Mat benchmarks demonstrate that our approach outperforms SOTA mask-guided  methods.
\end{itemize}
\section{Related Work}
\textbf{Deep trimap-based and trimap-free matting.} Since Adobe~\cite{deepmatting} develops a training method on large-scale synthetic matting datasets that can generate large and diverse composited training matting data with ground-truth alpha strictly following Eq.~\ref{alphaeq}, both trimap-based~\cite{gca,deepmatting,indexnet,rmat_ref} and trimap-free~\cite{lfm_ref,modnet_ref,rethink_p3m,aim,shm,rvm,AM-2k} deep matting methods have been promoted significantly on natural matting. However, trimap-based approaches rely on complex trimaps, while trimap-free approaches lack  user interaction or auxiliary inputs, so their performances rely on the  distribution of training matting data and can not be improved by user guidance. These problems limit the application of deep matting.

\textbf{Mask-guided matting.}  To extend the application of deep matting, MGMatting~\cite{mgm_ref} and IGF~\cite{igf_ref} use accessible coarse guidance masks as auxiliary inputs. To utilize coarse masks, MGMatting designed training perturbation strategies, including dilation and erosion, on guidance masks. For real-world adaptation, similar to Context-aware matting~\cite{context}, MGMatting  applies strong data augmentation including re-JPEGing, gaussian blur, and gaussian noises to input images during training. While MG-Wild~\cite{mgw} adopts the Mean teacher~\cite{mean_teacher} mechanism on the matting task across composited matting data and real-world data to achieve better generalization ability. 

\textbf{Spatial sampling and alignment} are useful learnable techniques to create more flexible neural networks. Deformable convolution~\cite{dcn} combines convolution kernels with learnable offsets to sample or aggregate features with flexible receptive fields. For video processing tasks, ~\cite{DFF_ref} uses learnable spatial alignment as optical flow. SFNet~\cite{sfnet_ref} dynamically aligns feature maps in different resolutions with learnable offsets.

\textbf{Line detection.} Line detection methods can be classified
into handcrafted methods and learning-based methods. Handcrafted methods~\cite{lsd_ref,edlines,mlsd,elsed} are traditionally performed based on the image gradient. Deep line detection was first
introduced through wireframe~\cite{wireframe_ref} parsing tasks. Several approaches estimate the structural lines of a scene by representing the line segments with two endpoints~\cite{lcnn}, attraction fields~\cite{afm}, graphs~\cite{zhang2019ppgnet}, etc. DeepLSD~\cite{deeplsd_ref} combines deep learning methods with classical line extractors, which supervises deep networks with attraction fields generated by LSD~\cite{lsd_ref} through the homography adaptation technique~\cite{superpoint} and uses the prediction of deep networks to improve the results of the LSD detector.

\begin{figure*}[t]
	\centering
	\includegraphics[trim={0 0cm 0 0cm},clip,width = 0.92\textwidth]{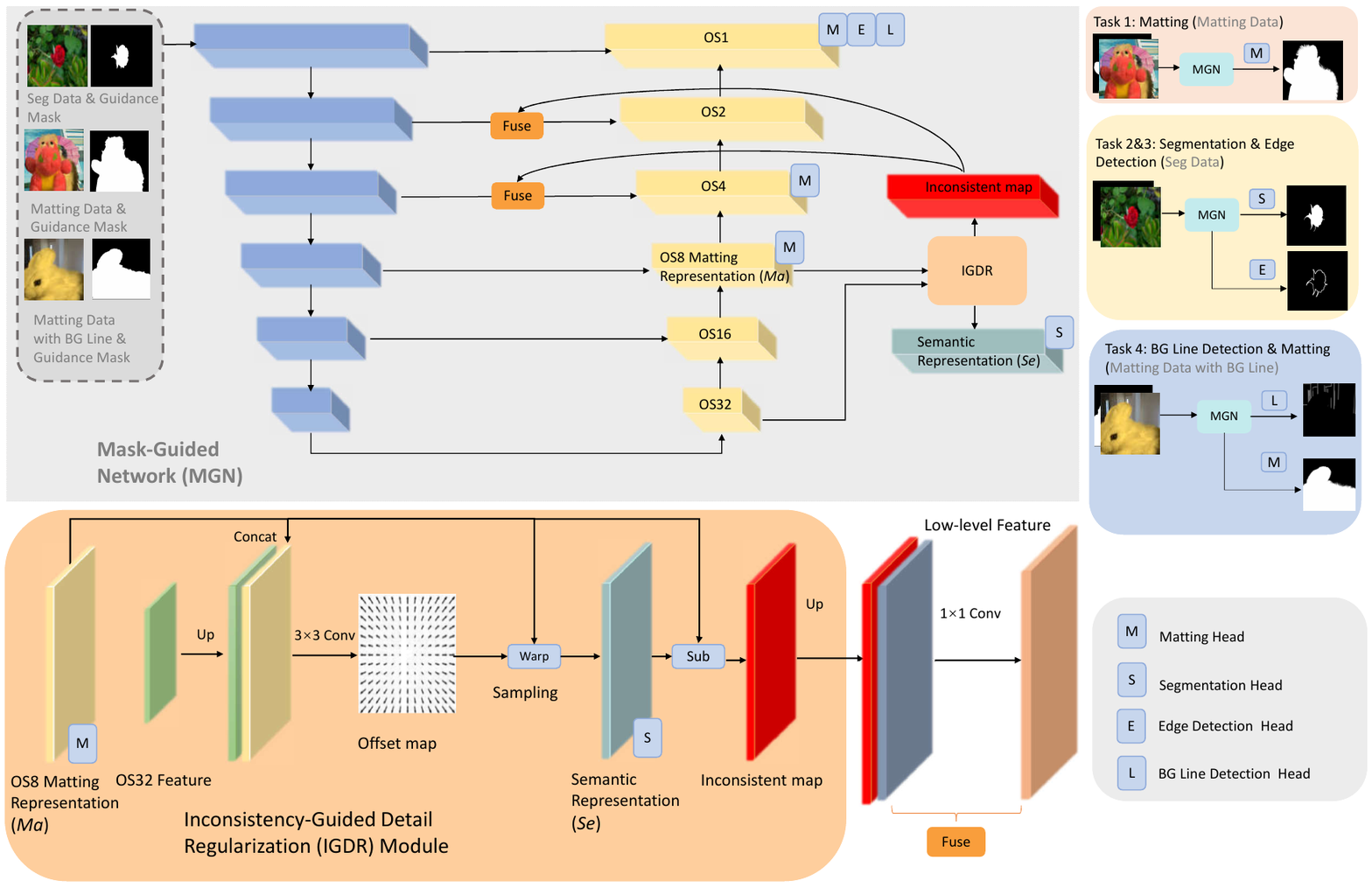}
	\caption{Overview of our proposed auxiliary learning framework and our proposed network. The proposed network leans multiple representations from different types of data and annotations in our auxiliary learning framework. Our IGDR module uses the inconsistency between matting representation and real-world adaptive semantic representation to regularize refinement on low-level details. }
	\label{fig:arch}
	\vspace{-1em}
\end{figure*}

\section{Our Auxiliary Learning Framework}

\textbf{Overall framework}. To generalize to objects with complex structures in  real-world scenes, avoid overfitting on wrong details, and suppress background interference, we proposed a novel auxiliary learning framework with three auxiliary tasks: semantic
segmentation, edge detection, and background line detection, and an inconsistency-guided detail regularization (IGDR) module as shown in Fig.~\ref{fig:arch}. Task 1 uses fine matting data for the matting task to learn detailed matting representations. Tasks 2 and 3 use the real-world segmentation data for both segmentation and edge detection tasks to learn real-world adaptive semantic representations on diverse and complex objects. Since background line detection needs a background image for composition, Task 4 uses only synthetic matting data for background line detection and matting, to learn a discriminative representation for better distinguishing foreground objects from background lines or textures. Additionally, we propose an inconsistency-guided detail regularization
(IGDR) module into our auxiliary learning network, utilizing the inconsistency between matting representation and semantic representation to regularize low-level detail refinement.

\textbf{Network architecture}. We adopt a ResNet34-UNet matting network proposed in~\cite{mgm_ref} with three matting heads as the base network. According to our auxiliary learning framework, we attach our  segmentation head,  edge detection head, and background line detection head to the features at output
stride (OS) 8, 1, and 1 respectively to learn extra representations. And our IGDR module is inserted between OS32 and OS8 features, as shown in Fig.~\ref{fig:arch}.

\subsection{Learning real-world adaptive semantic representation}
Models trained with synthetic or less diverse  matting data tend to fail at complex structures and real-world scenes such as shadows, due to the limitation of composited data and the hardness to acquire precise alpha mattes for diverse objects in diverse real scenes. Additionally, these models simply learn detailed matting representation and neglect the high-level semantics of real objects. Since real-world data with segmentation masks is easier to attain and can provide semantic supervision for a large number of  diverse and complex objects in various real scenes, we introduce an effective auxiliary learning framework to learn a real-world adaptive semantic representation,  enabling the network to adapt to complex structures and various real scenes. 

As shown in Fig.~\ref{fig:arch}, besides the matting task and its supervision $L_{MatData}$ using the $L_1$ regression loss and  Laplacian loss as~\cite{mgm_ref} on matting data, we implement supervisions of semantic segmentation and edge detection on real-world segmentation data. Instead of naively using the segmentation data with binary masks to supervise the original matting heads, we incorporate additional supervisions, introducing an extra segmentation head on high-level feature maps (OS8) in our matting network. This allows us to learn the high-level semantics of target objects. Additionally, we introduce an extra edge detection head on high-resolution feature maps (OS1), fused with low-level features, to capture real-world boundaries and object contours. The segmentation ground-truth mask is provided by real-world segmentation datasets, while we generate binary edge ground-truth masks from segmentation ground-truth masks like~\cite{ce2p_ref} did. Hence, the total loss of Tasks 2 and 3 on real-world segmentation data $L_{SegData}$ can be formulated as:
\begin{equation}
	L_{SegData} = L_{Seg} + L_{Edge} ,
	\label{eq:loss_seg}
\end{equation}
where $L_{Edge}$ denotes the weighted cross entropy loss~\cite{ce2p_ref} function
between the edge map by our edge detection head and the
binary edge GT mask; $L_{Seg}$ denotes the binary cross entropy
loss function between our segmentation output and the binary GT mask.

In addition to learning the matting representation ($Ma$) from matting data, we leverage real-world segmentation data to learn a real-world adaptive semantic representation ($Se$). This auxiliary representation helps the network handle diverse and complex objects across various real-world scenes effectively.

\subsection{Inconsistency-Guided detail regularization}

Matting approaches focus on refining details of objects, but they are also prone to overfit low-level details in the wrong regions (SARI's textures on a woman's body)  as MGMatting in Row 2, Fig.~\ref{fig:illu_vis}.  As shown in Fig.~\ref{fig:ig}, we use labels of matting and semantic segmentation to point out where the network should focus on, for matting and segmentation tasks, respectively. It's easy to observe that the segmentation mask can be treated as a warped alpha matte with fewer low-level details and  their inconsistent regions highlight low-level details of correct regions that should be focused on. Based on this observation, and inspired by  spatial sampling or alignment works~\cite{dcn,DFF_ref,sfnet_ref} for spatial wrapping, we proposed an inconsistency-guided detail regularization module to guide and enhance low-level details of objects in proper regions and avoid overfitting in wrong regions.

As shown in Fig.~\ref{fig:arch}, our IGDR generates semantic representation $Se \in  R^{H\times W\times C}$ from matting representation $Ma\in  R^{H\times W\times C}$ with learnable spatial wrapping and then acquires their inconsistency to guide detail regularization. We firstly introduce the high-level semantic information from the OS32 feature map by concatenating it with the matting representation $Ma$, and use a $3\times3$ convolution to generate an offset map $\Delta\in  R^{H\times W\times 2}$  for spatial sampling. Then, to form $Se$, for every spatial point $p$ in $Se$, the warp process in Fig.~\ref{fig:arch} bilinearly samples a point $p' = p+\Delta (p)$ in $Ma$  as Eq~\ref{warp},
\begin{equation}
	  Se(p) =  \sum_{p_n\in \mathcal{N}(p+\Delta (p))}{w_{p_n}Ma(p_n)},
	\label{warp}
\end{equation}
where $\mathcal{N}(p+\Delta (p))$ denotes neighbors of the warped point $p+\Delta (p)$ in $Ma$, and $w_p$ denotes the bi-linear kernel weights calculated by the distance of warped grid. As shown in Fig.~\ref{fig:arch}, $Ma$ and $Se$ are fed to a matting head and a segmentation head, respectively, to learn corresponding representations. Then, we generate the inconsistent map $IN$ by $IN = Ma - Se$, which points out proper regions of important low-level details in the feature space. Subsequently, we use the inconsistent map $IN$ from our IGDR module as guidance and fuse it with low-level feature maps, to refine low-level details in proper regions and prevent overfitting them in wrong regions.

\subsection{Learning background line detection}

\begin{figure}[t]
  \centering
  \footnotesize
  \setlength{\tabcolsep}{0.5pt} 
  
  \begin{tabular}{cccc}
    \includegraphics[trim={0 0.5cm 0 0cm},clip,width=0.15\textwidth]{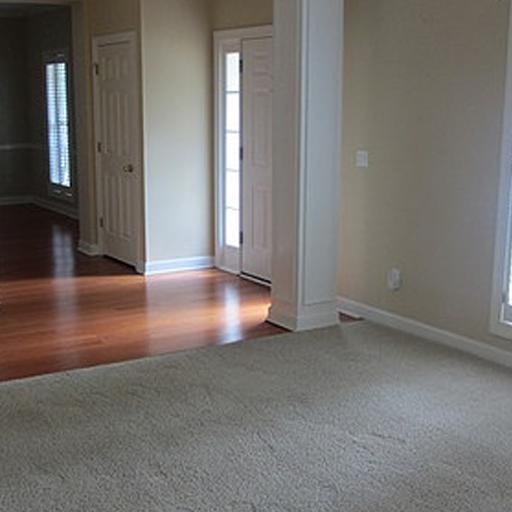} &
     \includegraphics[trim={0 0.5cm 0 0cm},clip,width=0.15\textwidth]{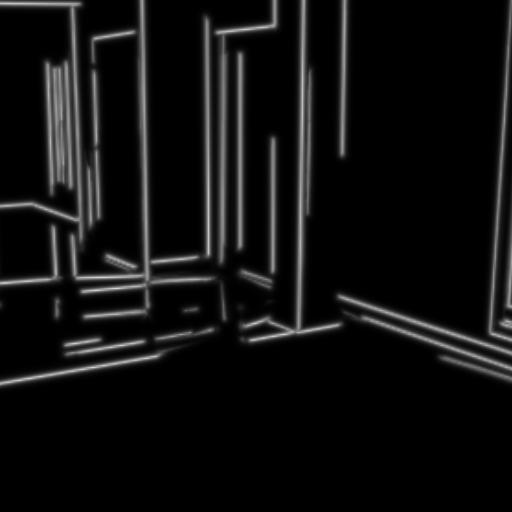} &
     \includegraphics[trim={0 0.5cm 0 0cm},clip,width=0.15\textwidth]{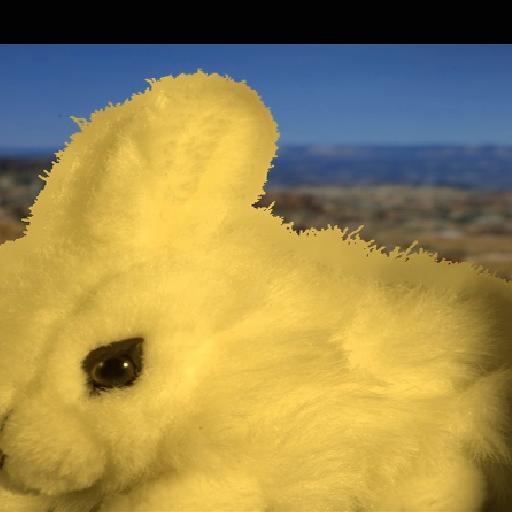}
     \\
     Background &
     Pseudo line &
     Foreground\\
     \includegraphics[trim={0 0.5cm 0 0cm},clip,width=0.15\textwidth]{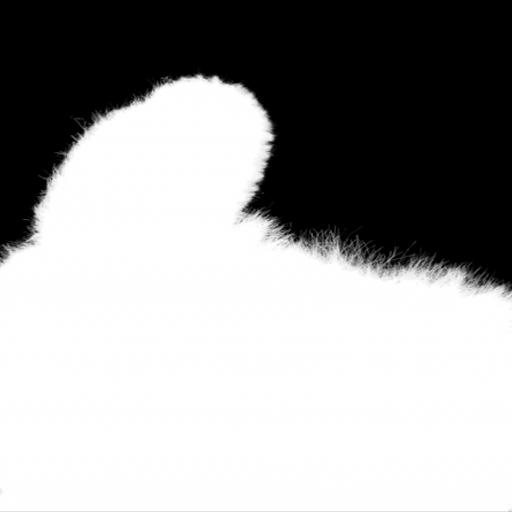} &
     \includegraphics[trim={0 0.5cm 0 0cm},clip,width=0.15\textwidth]{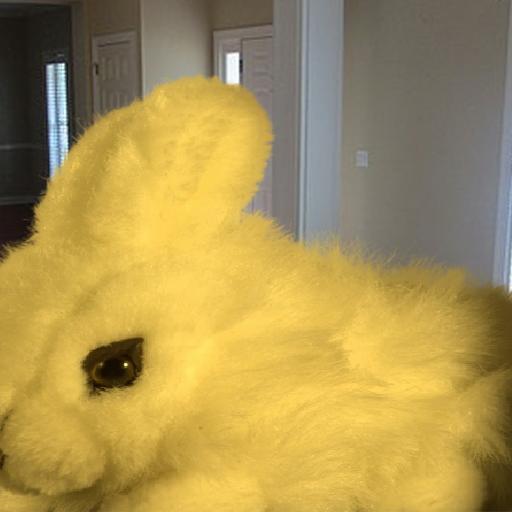} &
     \includegraphics[trim={0 0.5cm 0 0cm},clip,width=0.15\textwidth]{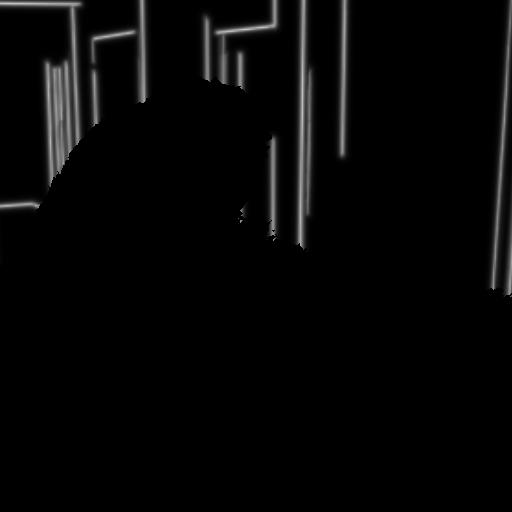}
     \\
     Alpha &
     Composition &
     Background line
  \end{tabular}
  
  \caption{Visualization of a training sample for background line detection.}
  \label{fig:iline}
  \vspace{-1.5em} 
\end{figure}
Distinguishing target objects in detail and suppressing interference of background lines or textures is an important challenge for matting. Previous mask-guided matting networks trained with detailed matting data also suffer from interference of background textures. Therefore, we proposed a novel auxiliary task, the background line detection in our framework, in order to learn discriminative representation to distinguish foreground objects from background lines or textures. 

To generate training samples for this task, for a background image such as ``Background'' in Fig.~\ref{fig:iline}, we first generate a representative pseudo distance field $D\in R^{H\times W}$ denoting the distance from every pixel to the nearest line, through  the homography adaptation~\cite{superpoint,deeplsd_ref}.  In detail, given a single background image $I\in R^{H\times W}$, we warp it with 100 random homographies $H_i$ to generate the warped images $I_i$, detect line segments in all the $I_i$ using the LSD~\cite{lsd_ref} line detector, then warp back the segments into $I$ to get a set $L_i$ of lines, then we generate a pseudo distance field $D_i\in R^{H\times W}$ for every set $L_i$ of lines, and then we calculate the median value of every spatial location of $D_i$ to get a more representative pseudo distance field $D$. Subsequently, we convert $D$ to an activation map $Pl = e^{-\frac{D}{2}}\in (0,1]$,  visually represented as the ``Pseudo Line" in Fig.~\ref{fig:iline}. Subsequently, as shown in Fig.~\ref{fig:iline}, a matting training image (``Composition'') is generated by compositing the background image with a foreground image (``Foreground'') using its alpha matte $A$ (``Alpha''). To get background line GT for supervision, we zero out $Pl$ in the unseen part of the background based on the corresponding  alpha matte $A$, and  we get background lines GT $Bl$ visualized as ``Background line'' in Fig.~\ref{fig:iline} by
\begin{equation}
Bl = \begin{cases}
Pl,  A<0.8,\\
ignore, 0.8\le A<1,\\
0, A=1,
\end{cases}
\label{eq:bl},
\end{equation}
we assign  the value as ignore, when $0.8\le A<1$ , to prevent step of $Bl$, and if the value of a pixel $i$ in $Bl$ is $ignore$, the loss function will not be calculated at $i$. 

As shown in Fig.~\ref{fig:arch}, we place the background line detection head on the high-resolution feature map (OS1). For Task 4, we use the pseudo GT $Bl$ to supervise the output of background line detection head $\hat{Bl}$, and use GT alpha matte $A$ to supervise the output of matting head $\hat{A}$, using $L_1$ regression loss in the neighborhood of lines in a background image based on the distance field $D$. The distance threshold for background line detection and matting in Task 4 are 13 and 3, respectively. The total loss $L_{BG}$ for Task 4 can be formulated as:
\begin{equation}
	L_{BG} = L_1^{Line}(D\le 13) + L_1^{Mat}(D\le 3),
	\label{eq:loss_line}
\end{equation}
where $L_1^{Line}$ is the $L_1$ regression loss between  $\hat{Bl}$ and $Bl$, and  $L_1^{Mat}$ is the $L_1$ regression loss between  $\hat{A}$ and $A$. With this novel auxiliary task and supervision of pseudo background line, our network learns a discriminative representation to suppress background interference of lines or textures.

Finally, we establish the total matting framework with all our auxiliary tasks, and the total loss is formulated as:
\begin{equation}
	L_{total} =L_{MatData}+L_{SegData} + L_{BG} .
	\label{eq:loss_seg}
\end{equation}

\section{Our Plant-Mat Benchmark}
To evaluate mask-guided matting for complex objects under proper and  clear backgrounds or scenes, we propose a plant matting test dataset, the Plant-Mat, containing 130 plant images and ground-truth alpha mattes with complex object structures, high-quality annotations, and look-natural composition. Unlike portraits and animals, plant objects usually have a large number of holes and elongated branches, as well as complex shadows and reflections on leaves, which is hard to annotate manually in high quality. Therefore, we capture the image of diverse plants in a blue screen studio and utilize a blue screen matting technique~\cite{smith1996blue} to generate preliminary foregrounds and alpha mattes. Then we denoise the alpha mattes with filters and manually refine the remaining defects. In this way, we can generate high-quality matting annotations for plants. Then we carefully composite the foregrounds of plants on proper clear background images by considering whether scenes or relationships between plants and background objects are proper. Finally, we get our high-quality Plant-Mat benchmark containing various plants with diverse and complex structures under clear backgrounds.

\section{Experiments}
\begin{figure*}[!t]
  \centering
  \scriptsize
  \setlength{\tabcolsep}{0.5pt} 
  
  \begin{tabular}{ccccc}
     \includegraphics[trim={0 0cm 0 2.5cm},clip,width=0.19\textwidth]{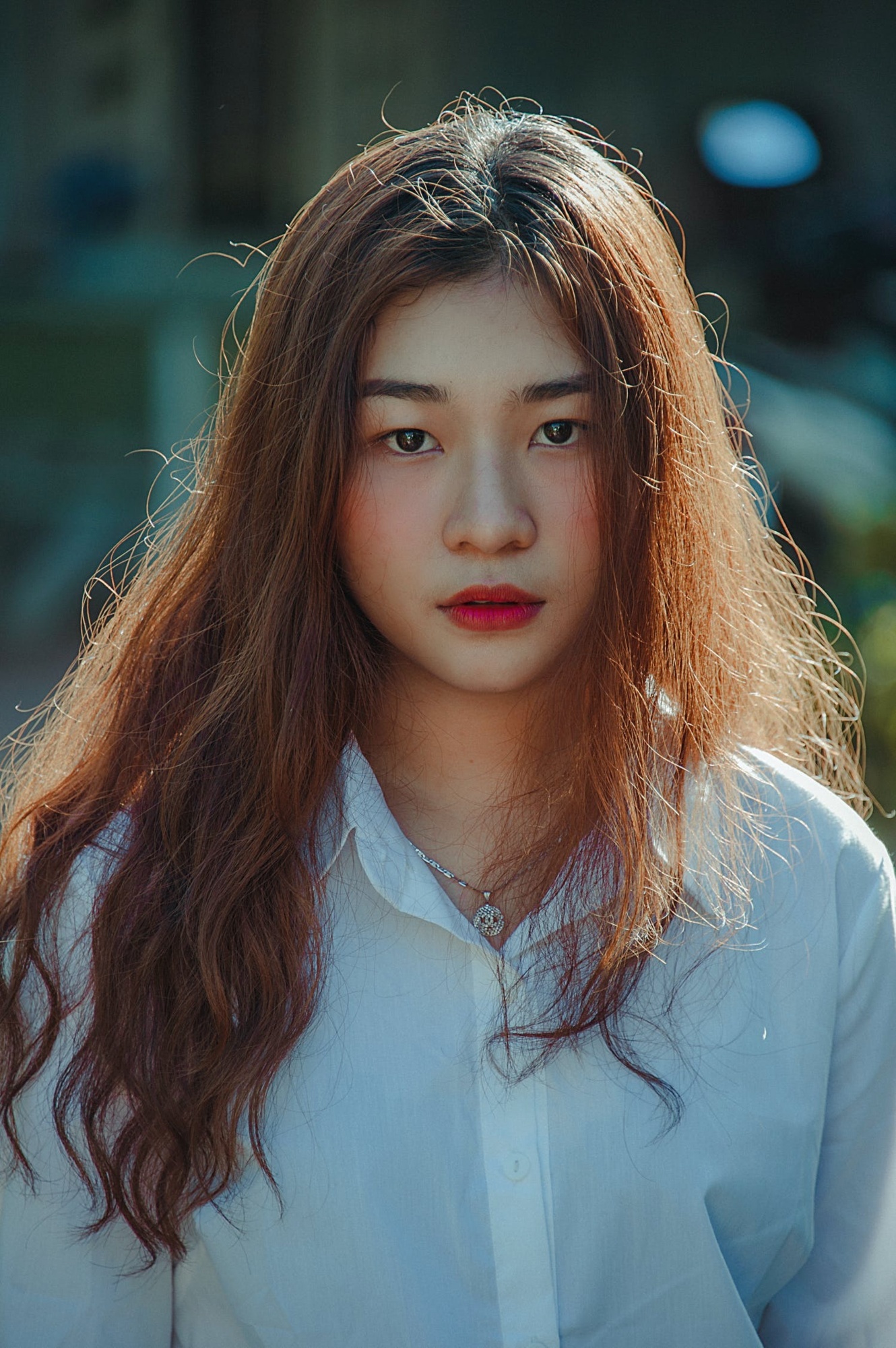} &
     \includegraphics[trim={0 0cm 0 2.5cm},clip,width=0.19\textwidth]{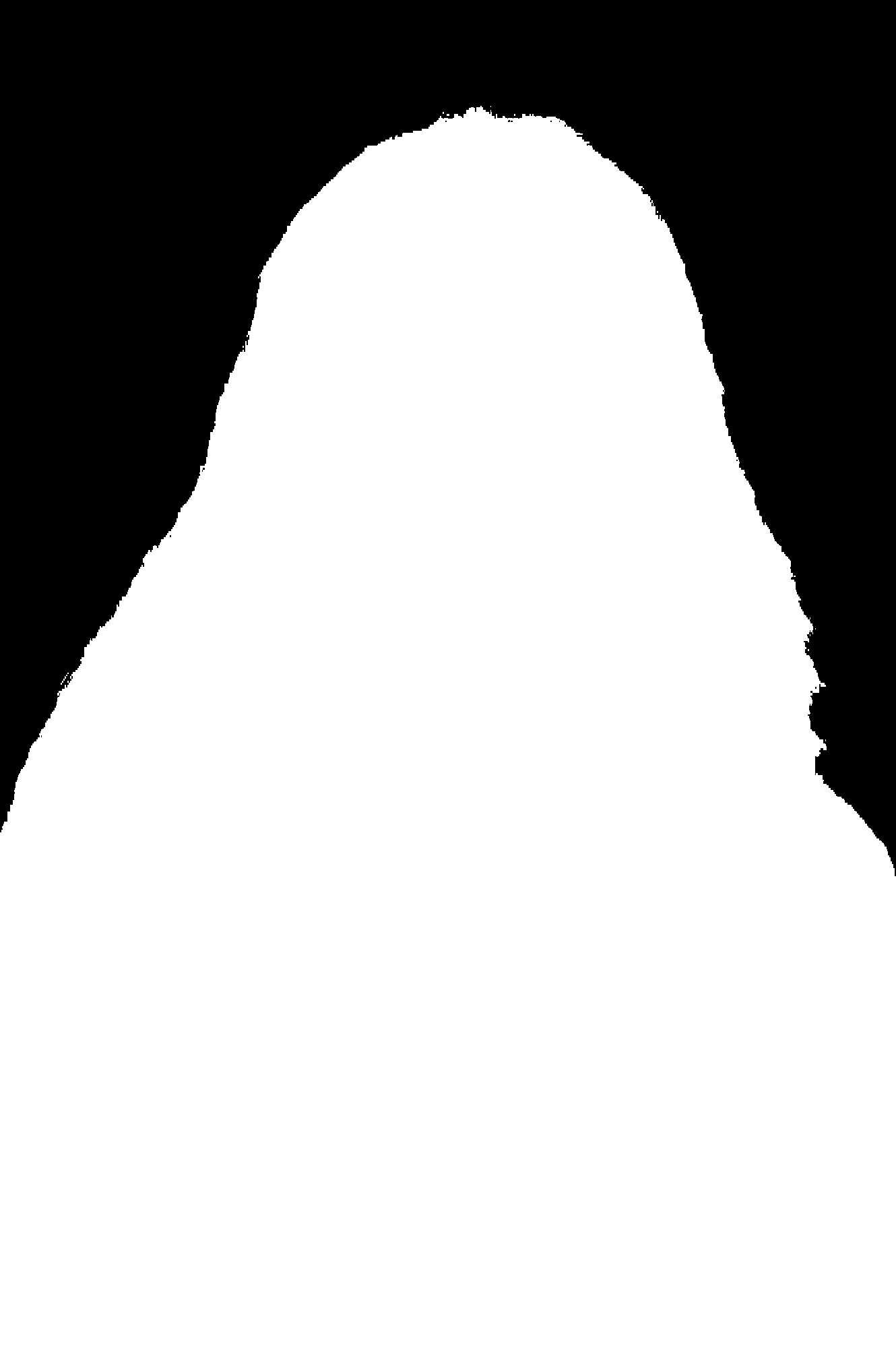} &
     \includegraphics[trim={0 0cm 0 2.5cm},clip,width=0.19\textwidth]{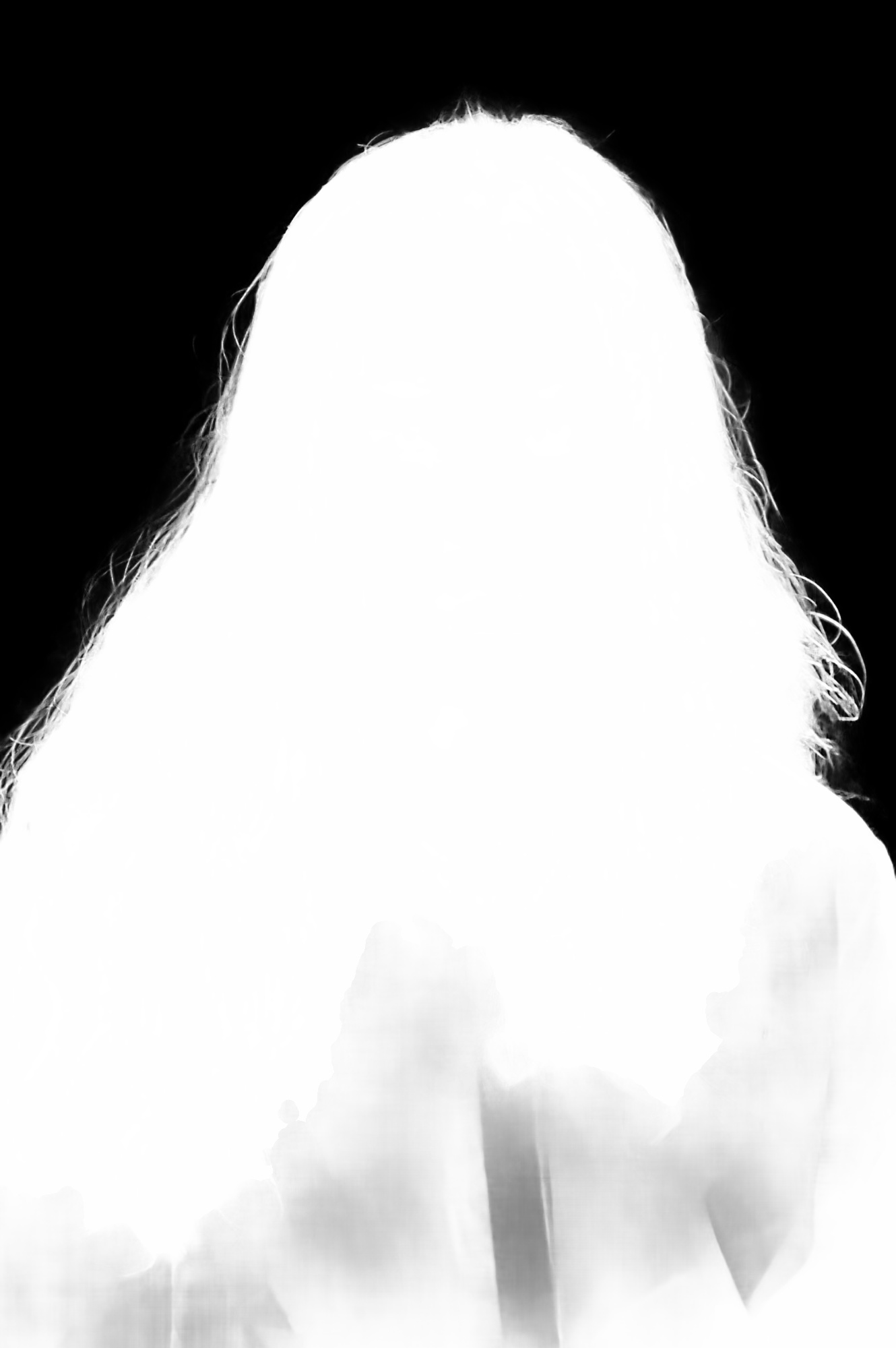} &
     \includegraphics[trim={0 0cm 0 2.5cm},clip,width=0.19\textwidth]{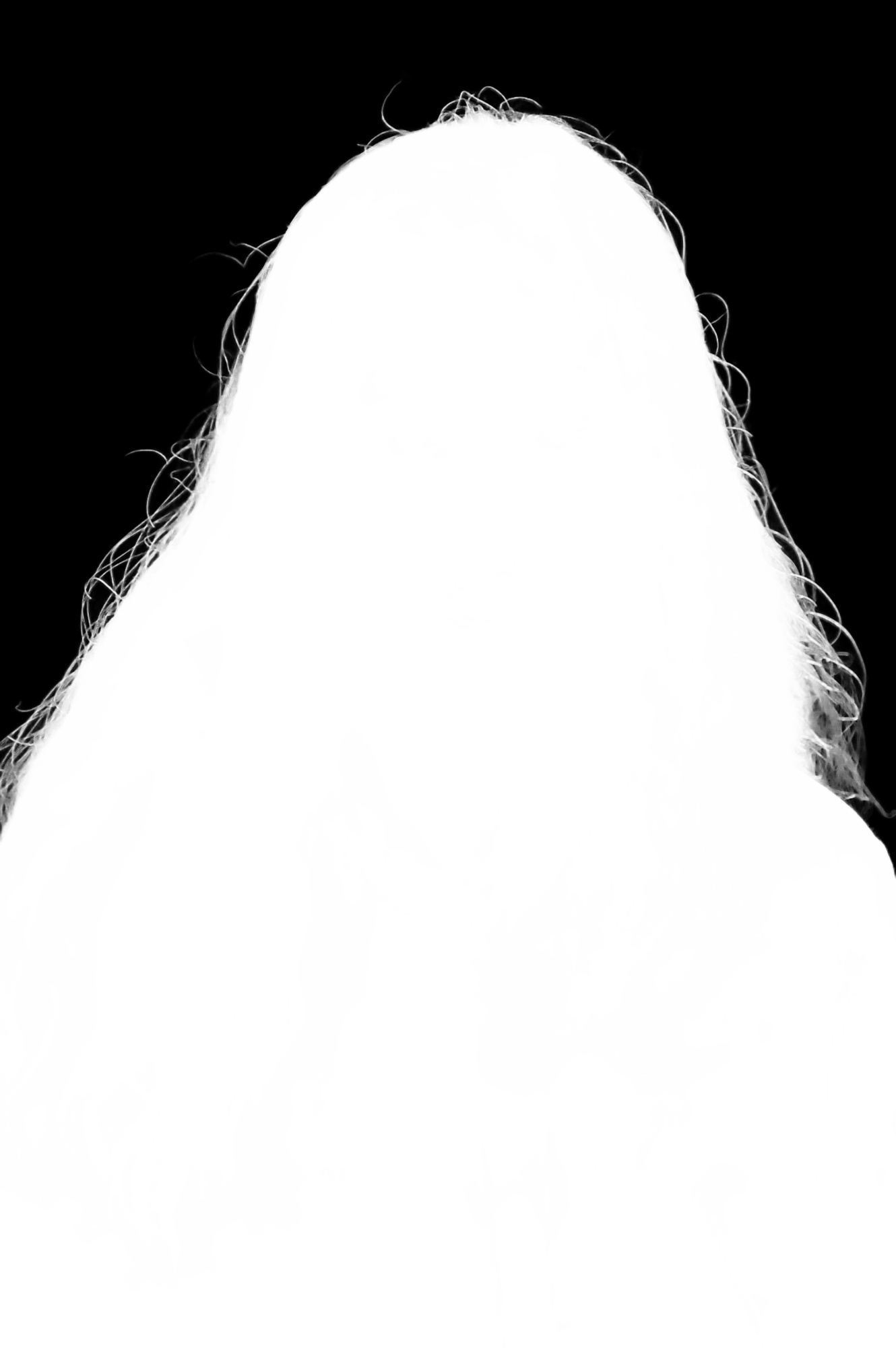} &
     \includegraphics[trim={0 0cm 0 2.5cm},clip,width=0.19\textwidth]{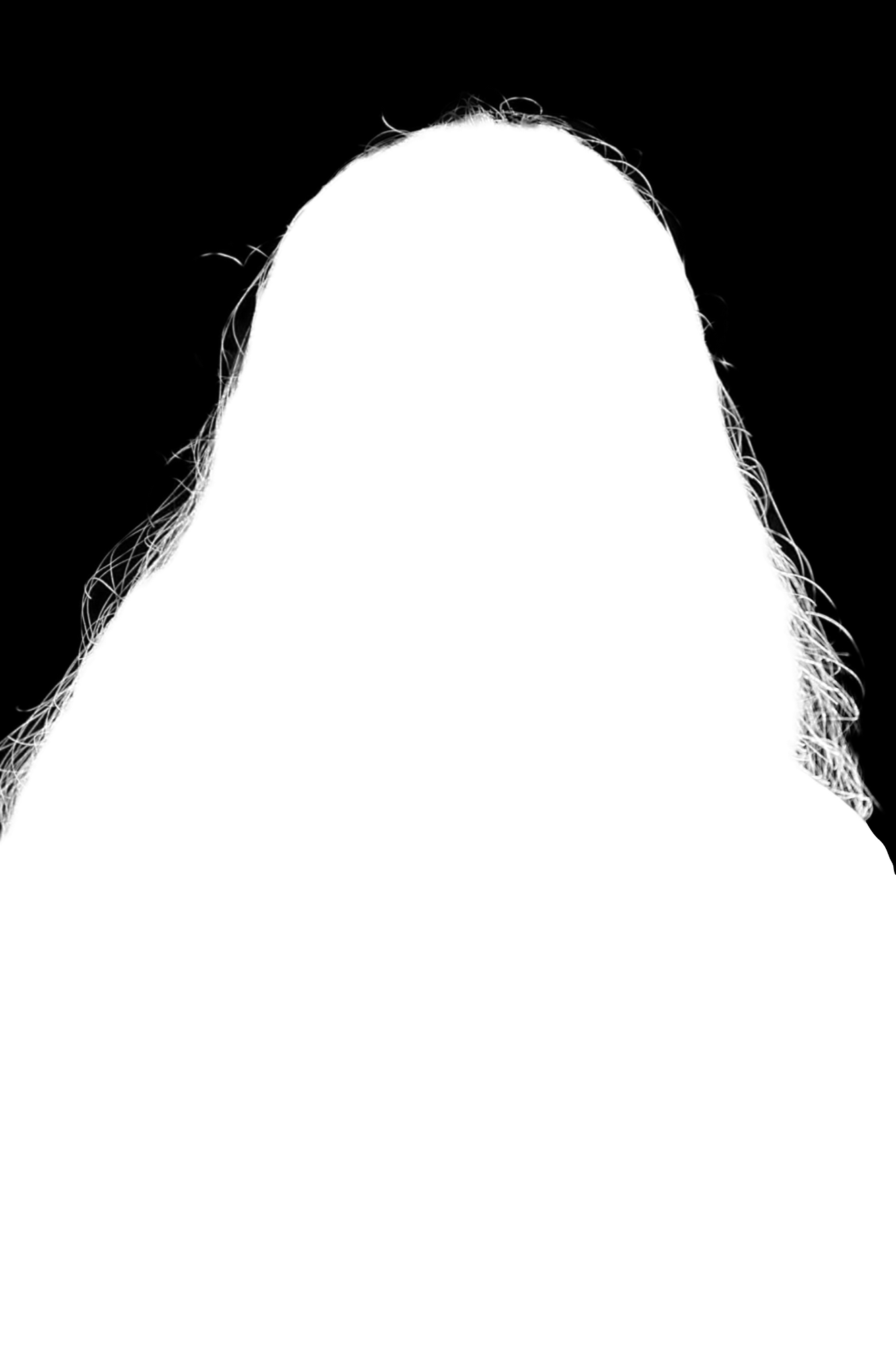}
     \\
     \includegraphics[trim={0 0cm 0 6cm},clip,width=0.19\textwidth]{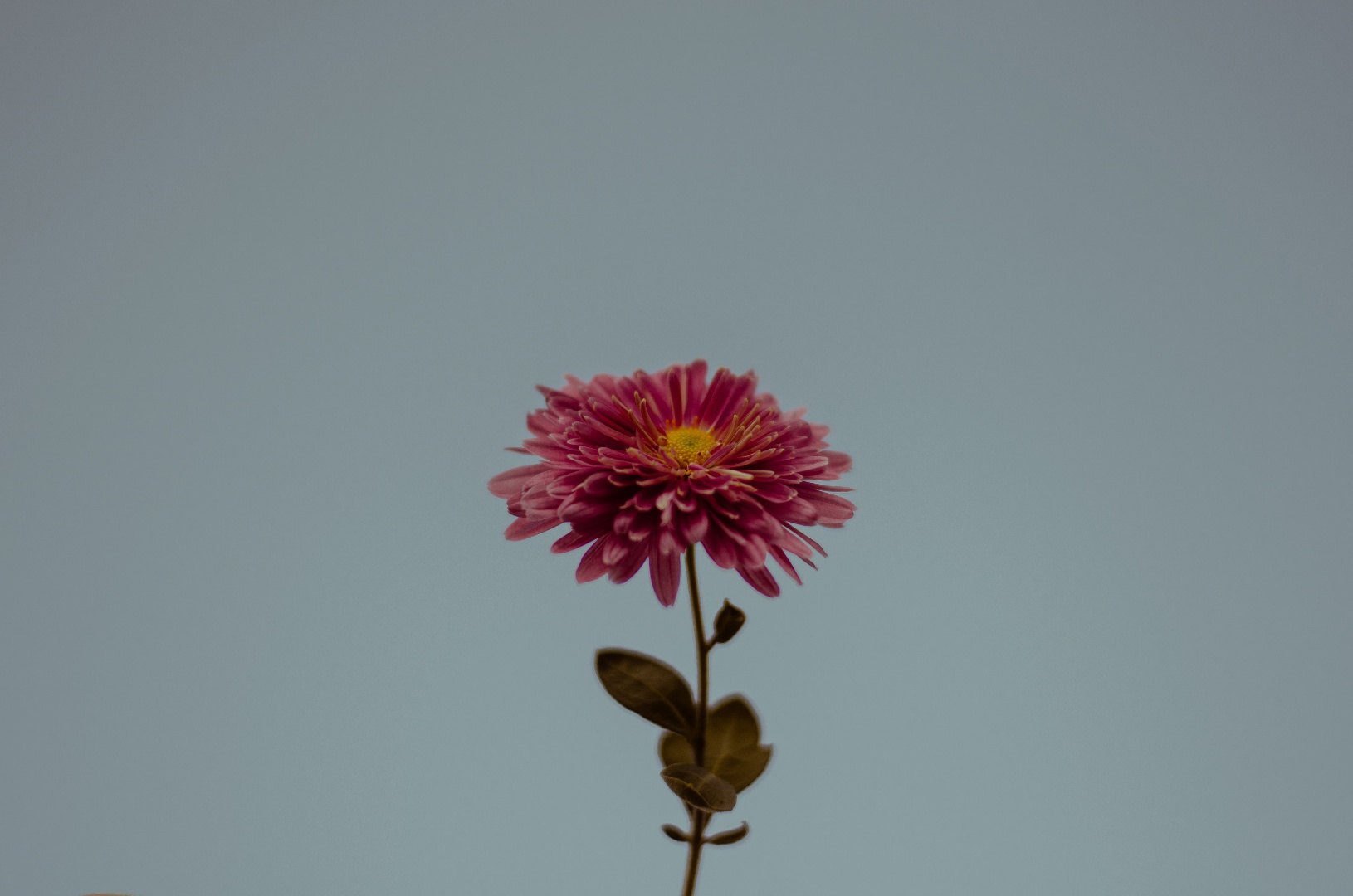} &
     \includegraphics[trim={0 0cm 0 6cm},clip,width=0.19\textwidth]{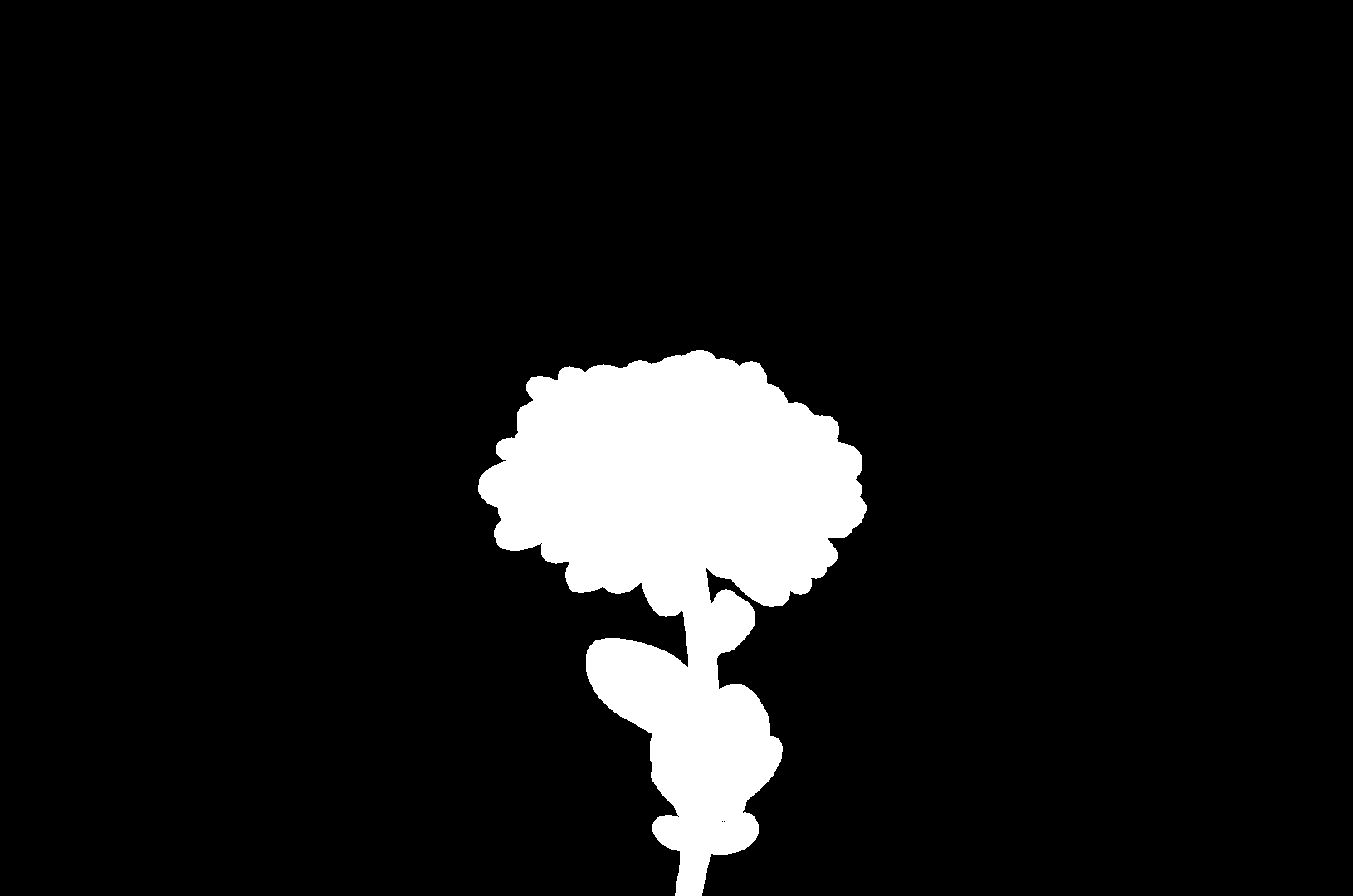} &
     \includegraphics[trim={0 0cm 0 6cm},clip,width=0.19\textwidth]{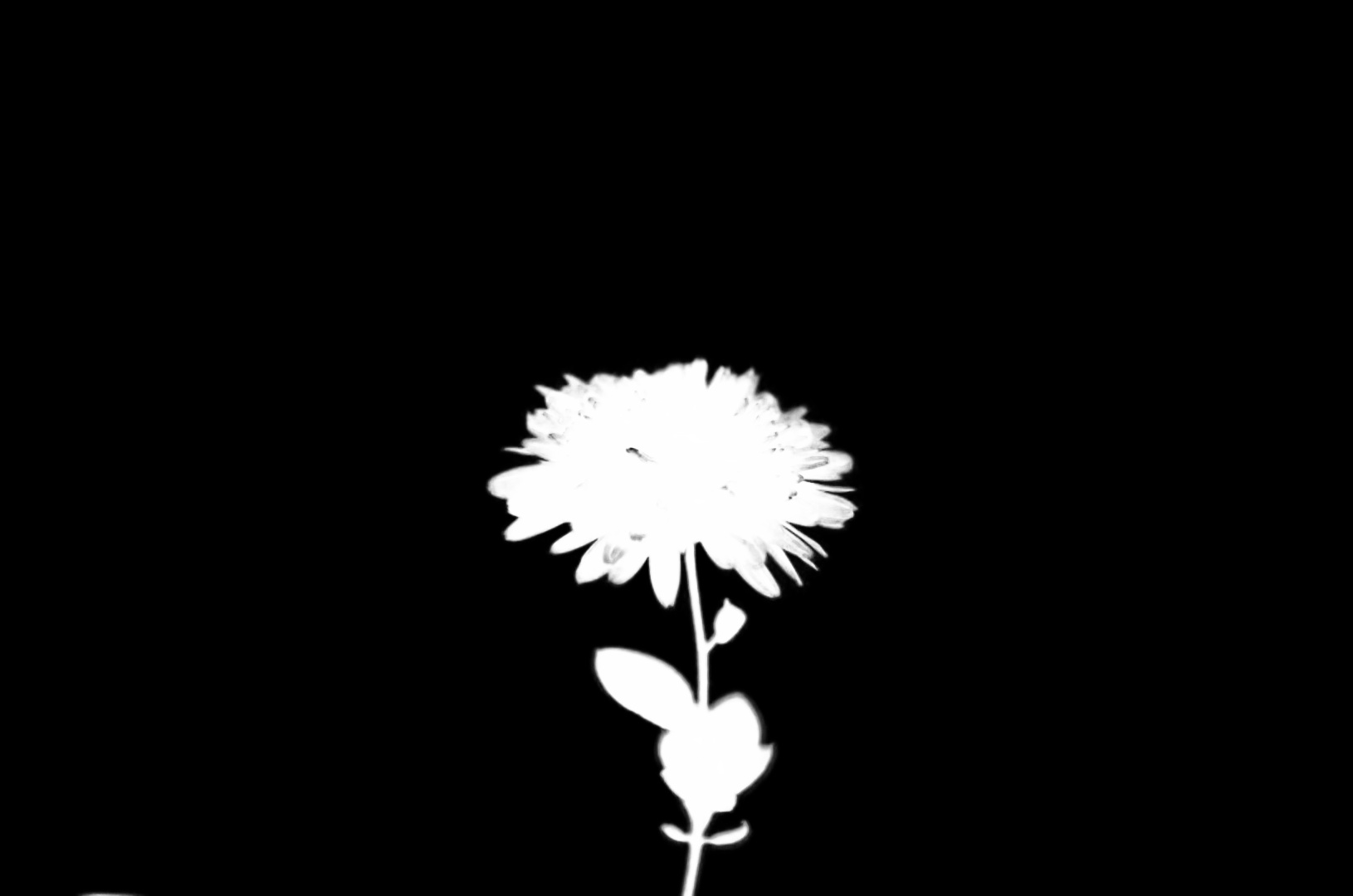} &
     \includegraphics[trim={0 0cm 0 6cm},clip,width=0.19\textwidth]{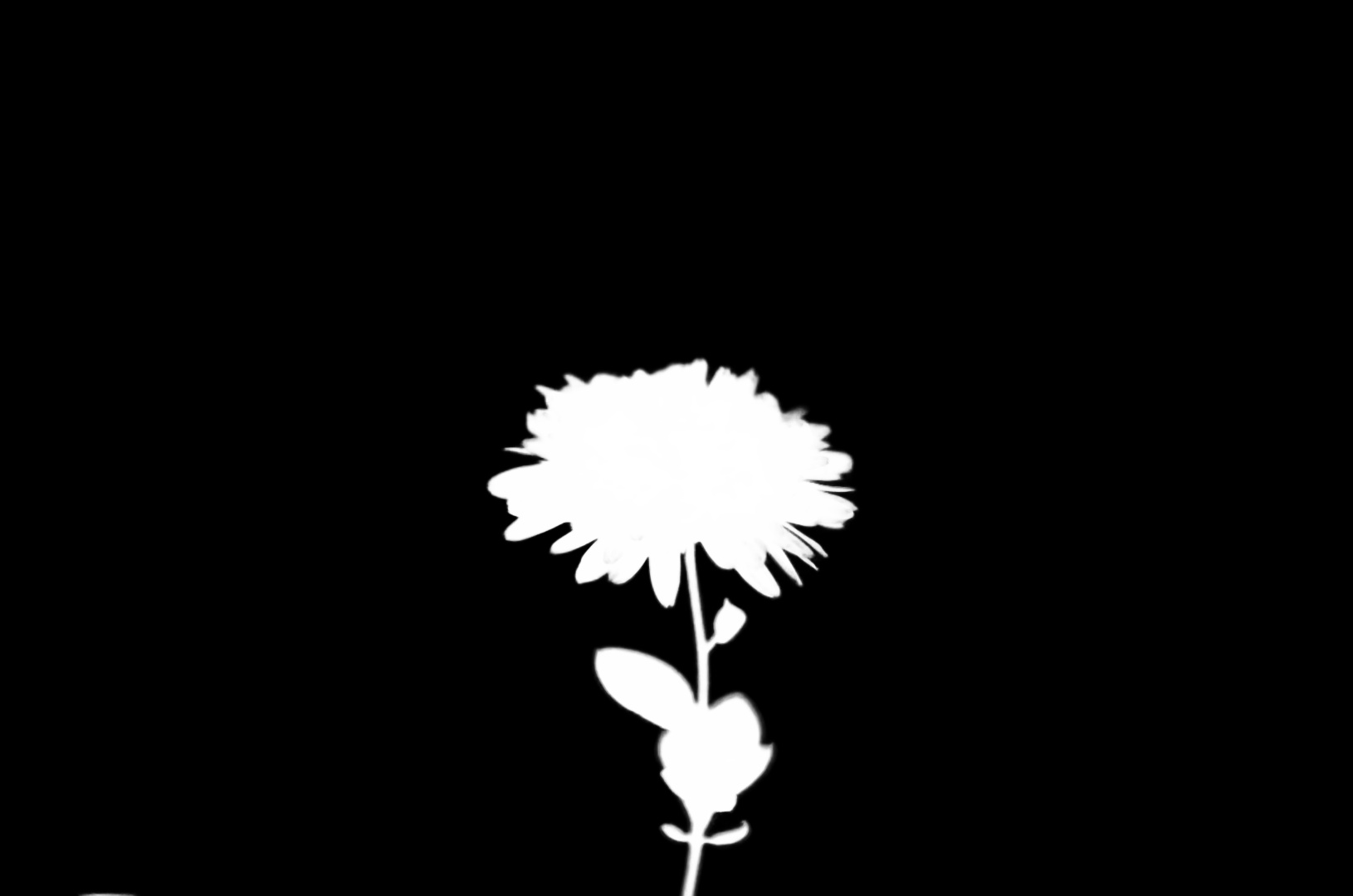} &
     \includegraphics[trim={0 0cm 0 6cm},clip,width=0.19\textwidth]{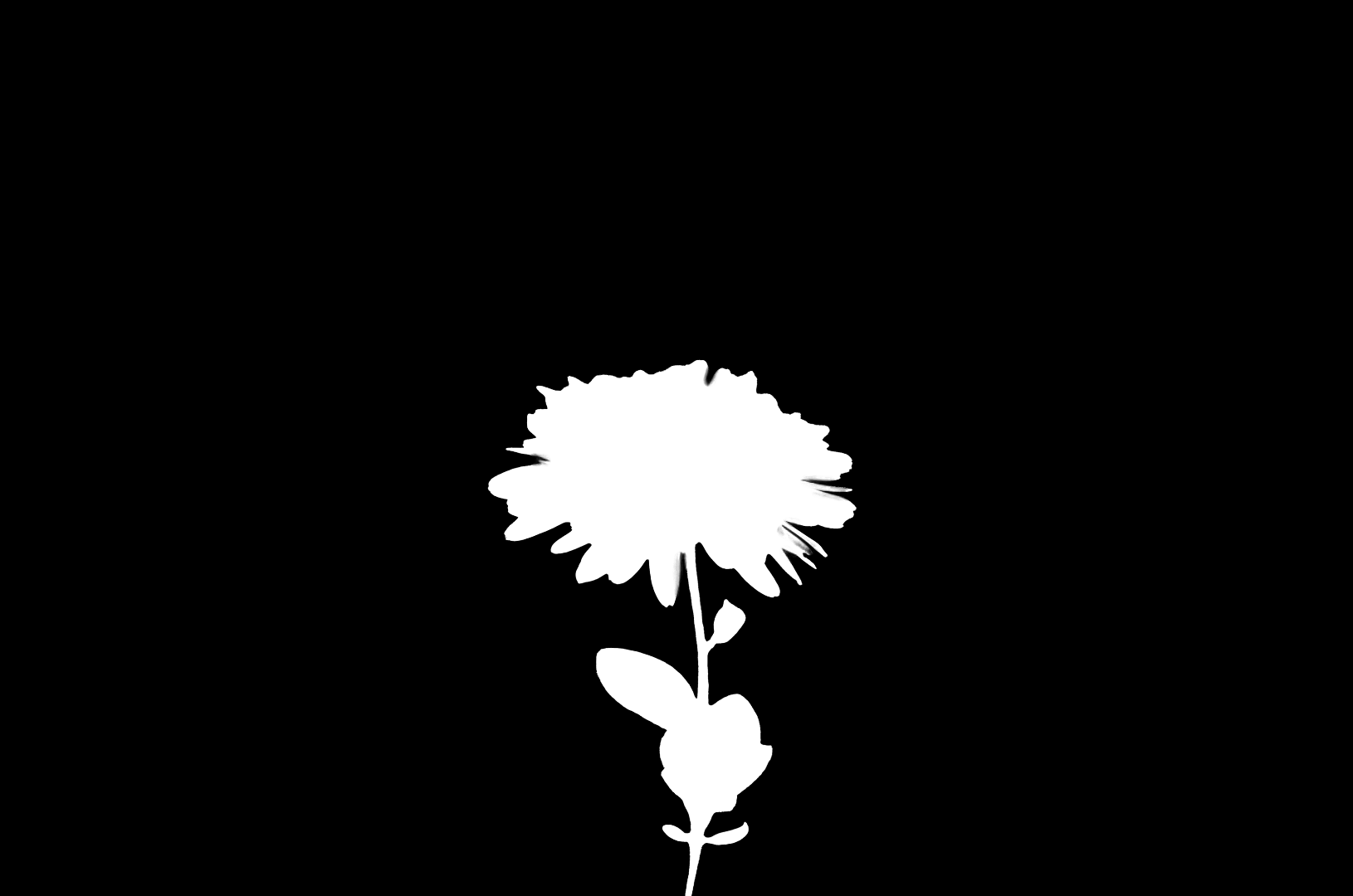}
     \\
     \includegraphics[width=0.19\textwidth]{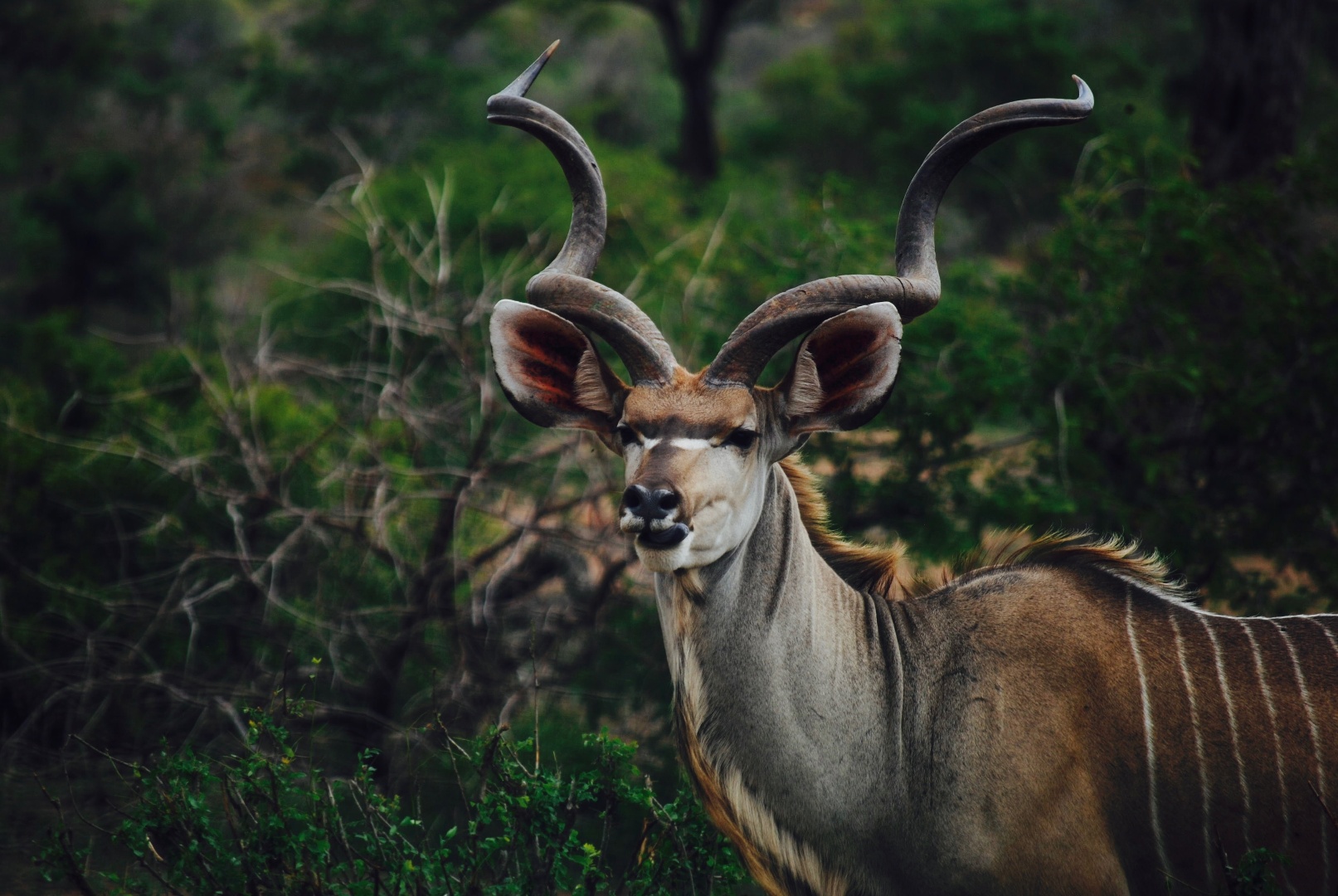} &
     \includegraphics[width=0.19\textwidth]{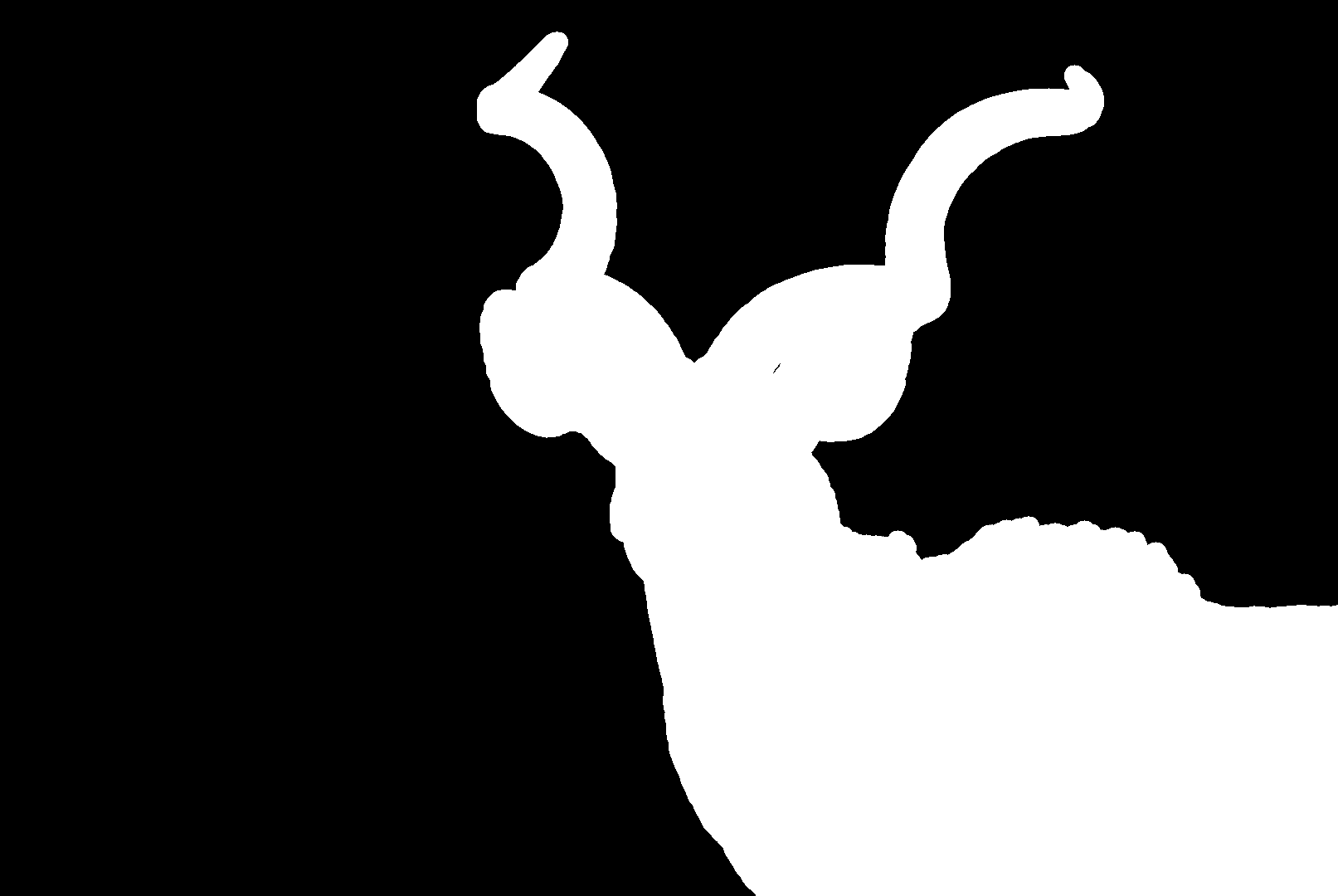} &
     \includegraphics[width=0.19\textwidth]{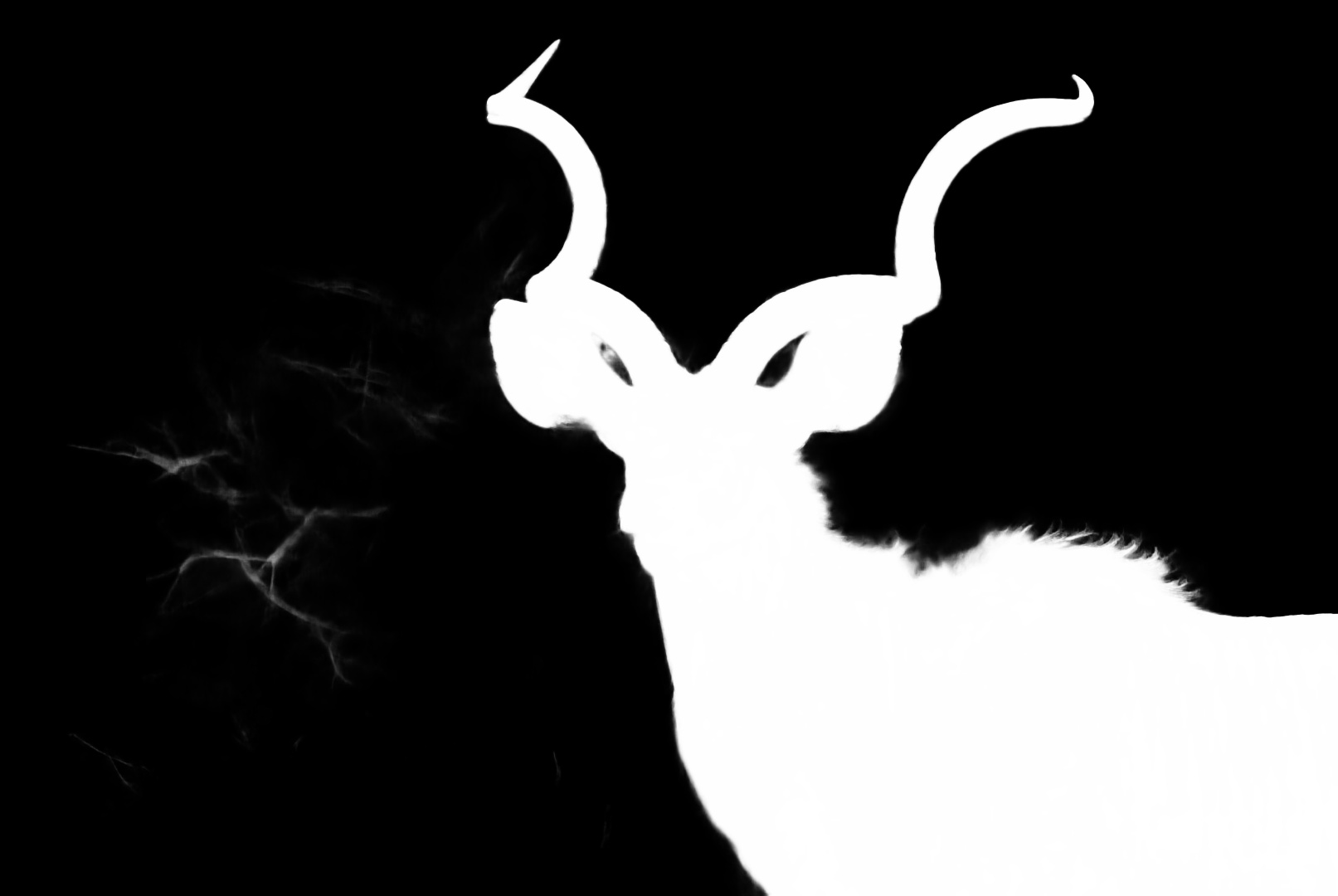} &
     \includegraphics[width=0.19\textwidth]{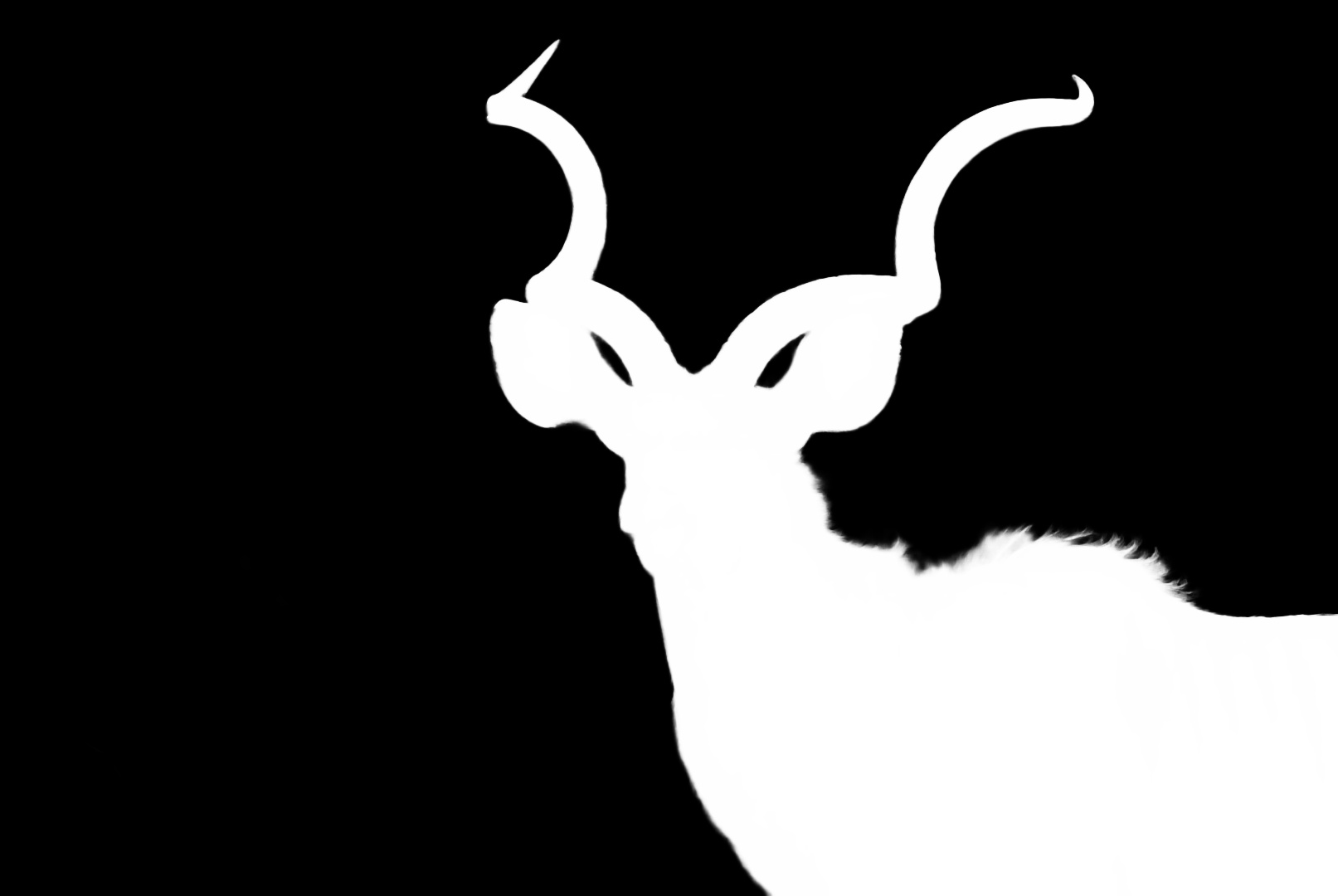} &
     \includegraphics[width=0.19\textwidth]{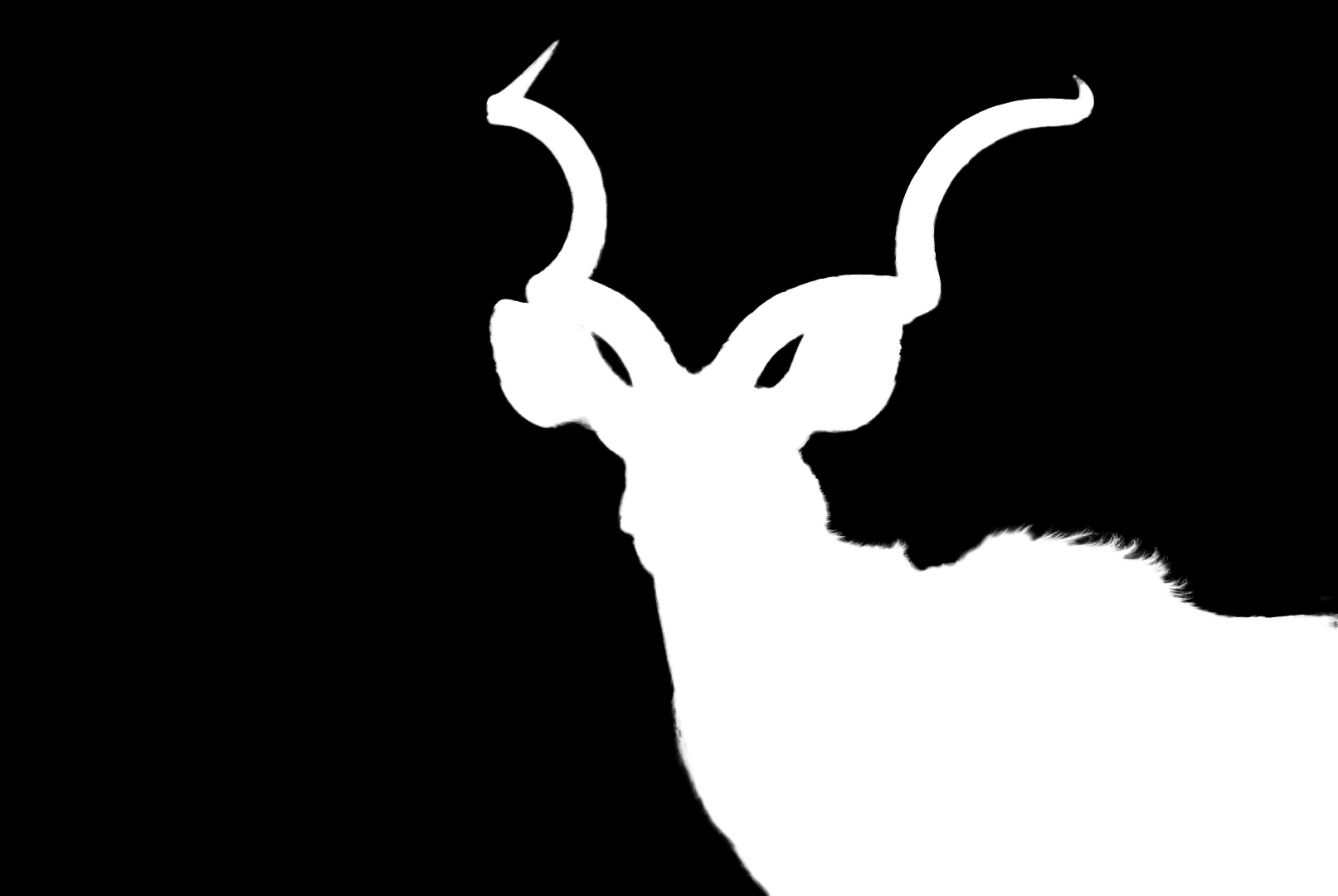}
     \\
     \includegraphics[width=0.19\textwidth]{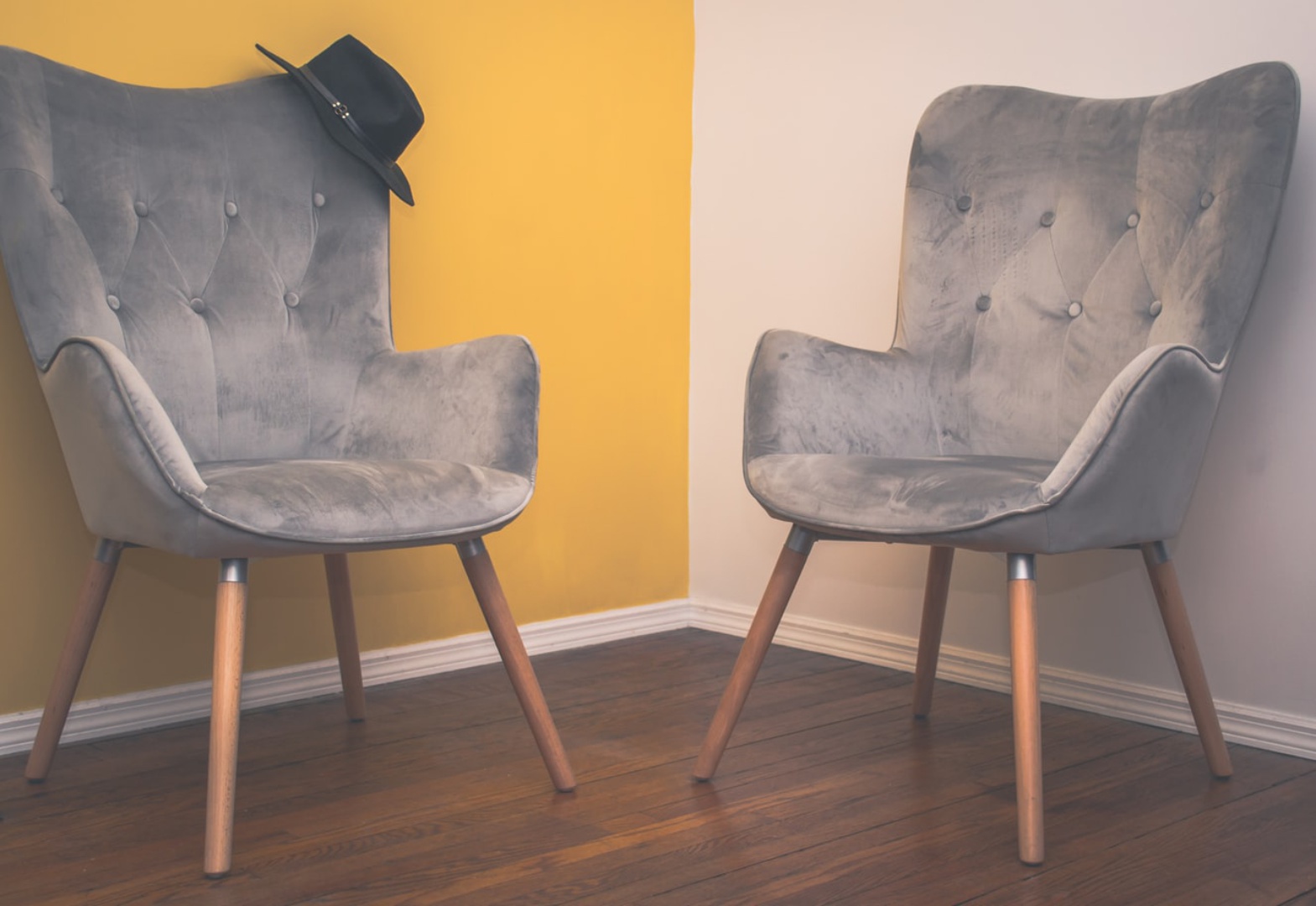} &
     \includegraphics[width=0.19\textwidth]{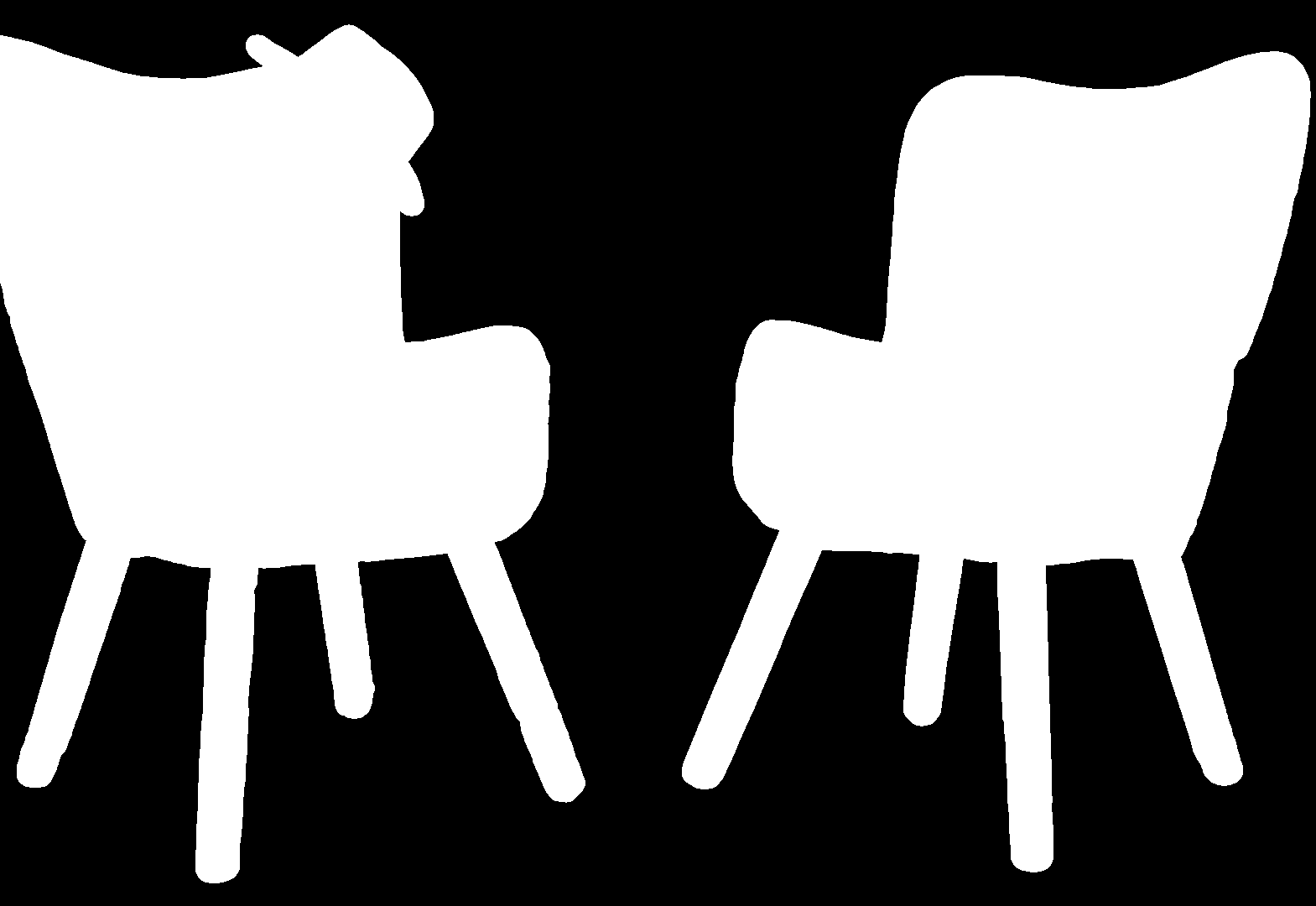} &
     \includegraphics[width=0.19\textwidth]{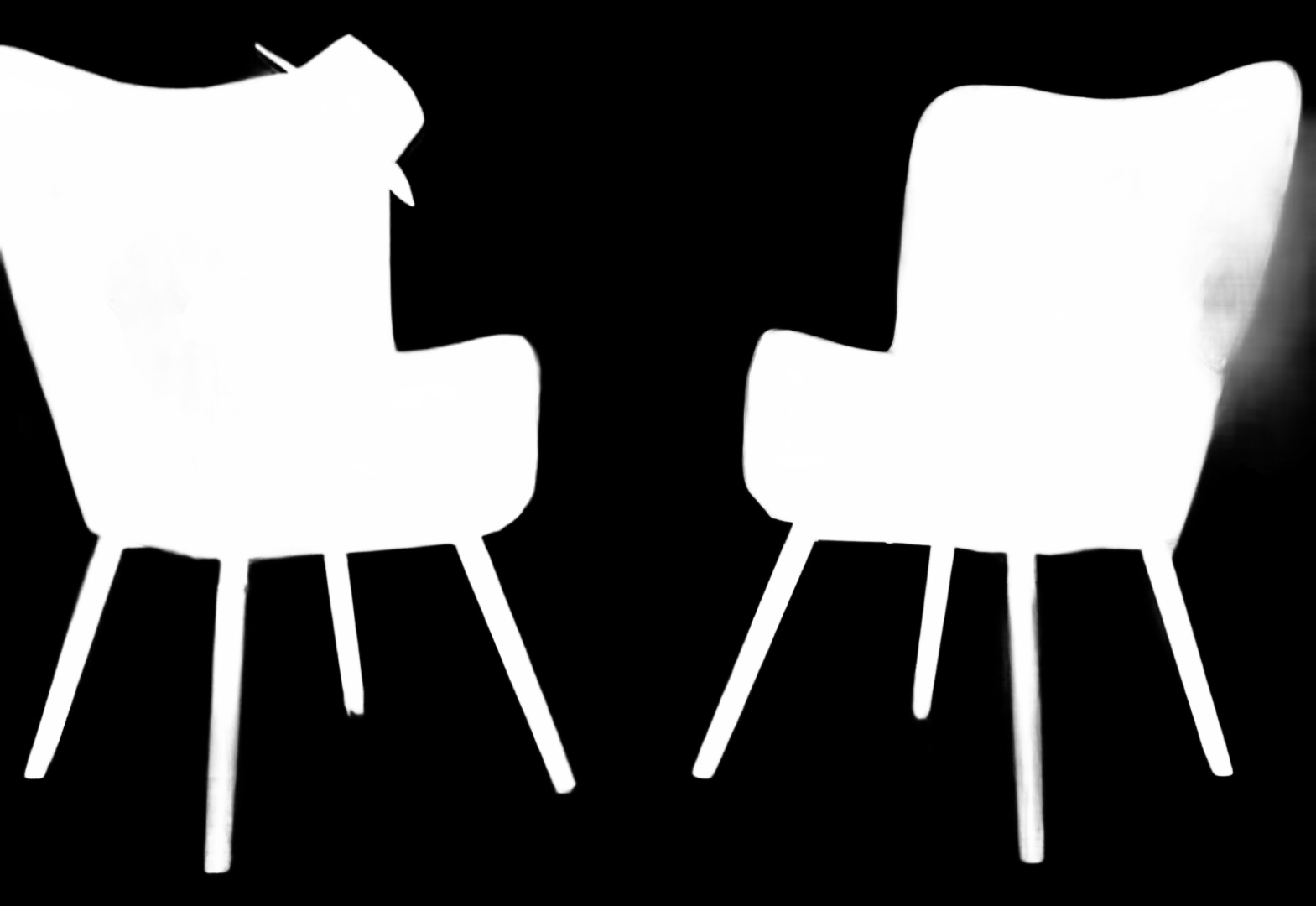} &
     \includegraphics[width=0.19\textwidth]{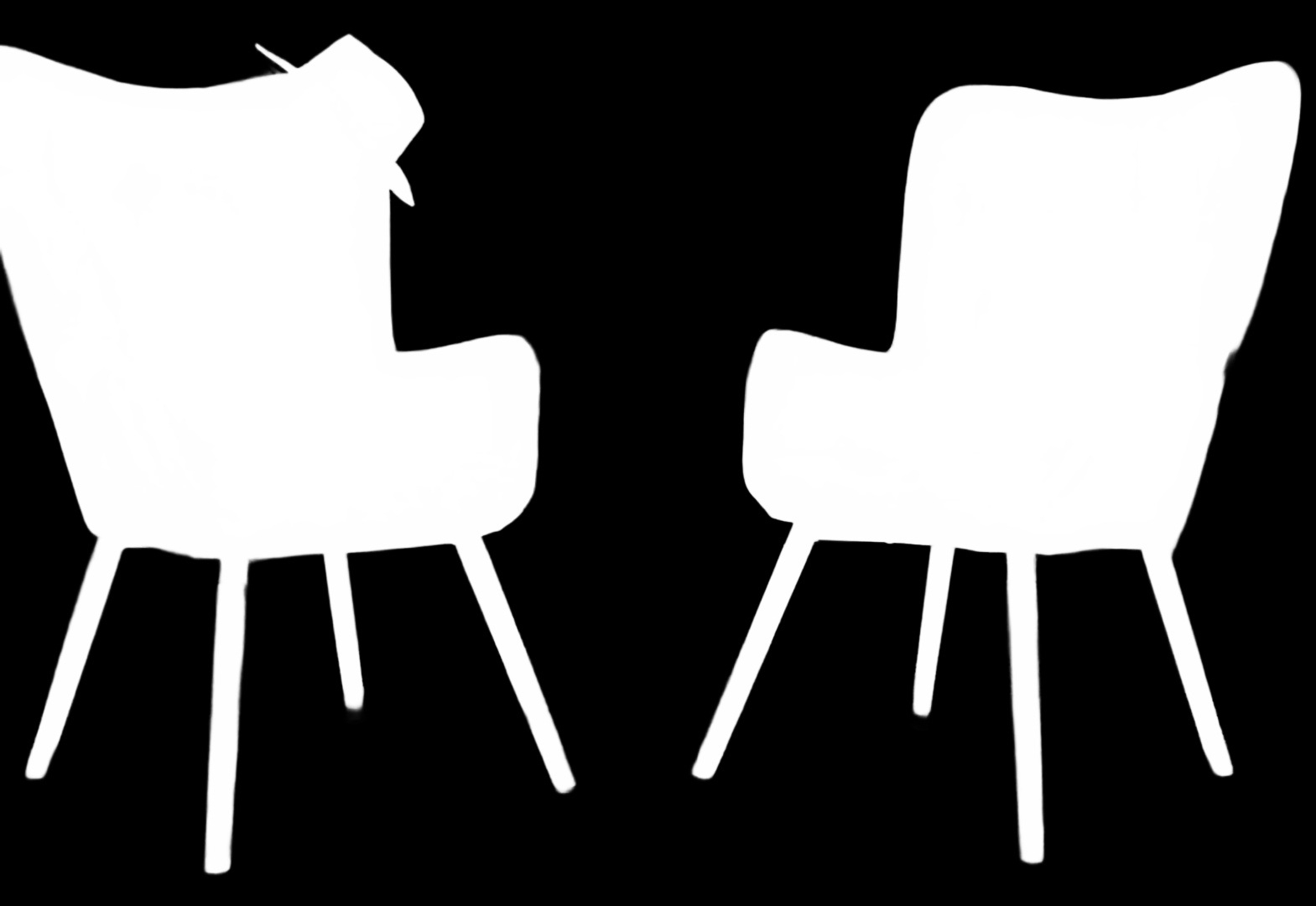} &
     \includegraphics[width=0.19\textwidth]{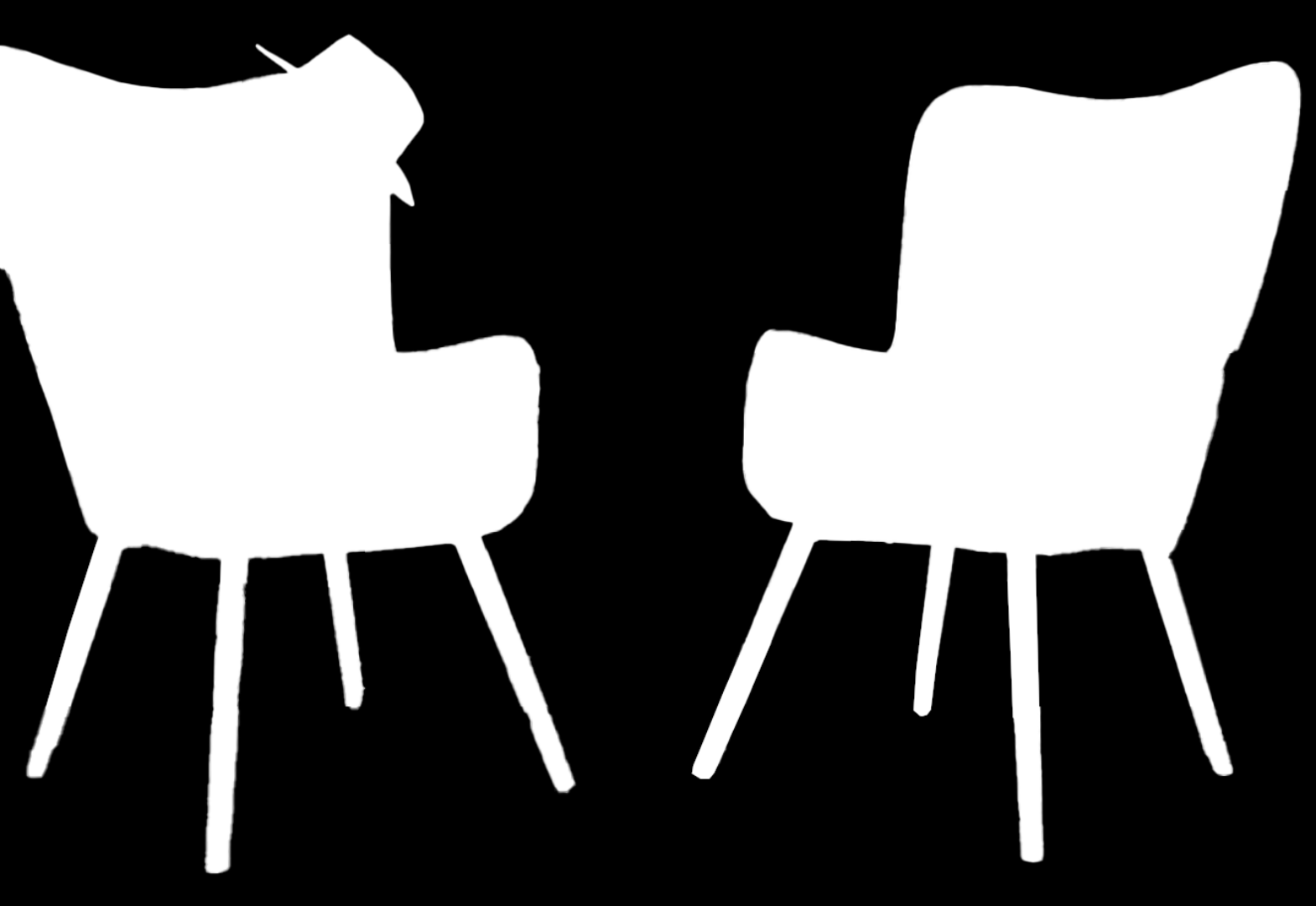}
     \\
     Image &
     Guidance &
     MGMatting &
     Ours &
     GT
     
 \end{tabular}
  
  \caption{The qualitative comparisons between MGMatting~\cite{mgm_ref} and ours on various real-world images from test sets~\cite{aim,mgm_ref,AM-2k}.}
  \label{fig:CompVis}
\end{figure*}
\subsection{Implementation details} 
\textbf{Training data and annotations}. We implement our auxiliary learning framework on different types of data and different types of annotation. For fine matting data, we adopt synthetic matting datasets including a subset of Adobe~\cite{deepmatting} with 269 foreground images like~\cite{mgm_ref} did, Human-2k~\cite{timinet} and Animal-2k (AM-2k)~\cite{AM-2k}, and a real-world portrait matting dataset P3M-10k~\cite{p3m} with about 10k images. To learn real-world semantic representation from diverse and complex real-world data, we introduce 2 high-resolution segmentation datasets, including UHRSD~\cite{uhrsd_ref} and HRSOD~\cite{hrsod_ref}. As for unlabeled background images for composited training data, we adopt COCO~\cite{coco} and Wireframe~\cite{wireframe_ref}. We generate a distance field of lines through a homography adaptation~\cite{deeplsd_ref} using~\cite{lsd_ref} for a background image, and then generate pseudo GT for background line detection based on the corresponding composited foreground.

\textbf{Training data augmentation}. We follow the strong data augmentation on synthetic training images and guidance perturbation in the real-world setting of MGMatting~\cite{mgm_ref} for all experiments. Especially, for Task 4 in our framework, we add  binary lines on binary guidance masks with random widths (from 2 to 8) based on the distance field $D$, with a probability of 0.2, to perturb the guidance.

\subsection{Benchmarks}

\begin{table}[tb]
\begin{center}
\caption{Results on RWP~\cite{mgm_ref} benchmark.}
\label{tab:rwp}
\begin{tabular}{l|ll|ll}
\hline
\multirow{2}{*}{Model}                                                & \multicolumn{2}{c|}{Whole Image}                     & \multicolumn{2}{c}{Detail}                        \\ \cline{2-5} 
                                                                      & \multicolumn{1}{l|}{SAD}           & \begin{tabular}[c]{@{}l@{}}MSE\\ ($10^{-3}$)\end{tabular}           & \multicolumn{1}{l|}{SAD}           & \begin{tabular}[c]{@{}l@{}}MSE\\ ($10^{-3}$)\end{tabular}           \\ \hline
DIM~\cite{deepmatting}                                                                   & \multicolumn{1}{l|}{28.5}          & 11.7          & \multicolumn{1}{l|}{19.1}          & 74.6          \\ 
GCA~\cite{gca}                                                                   & \multicolumn{1}{l|}{29.2}          & 12.7          & \multicolumn{1}{l|}{19.7}          & 82.3          \\ 
Index~\cite{indexnet}                                                                 & \multicolumn{1}{l|}{28.5}          & 11.5          & \multicolumn{1}{l|}{18.8}          & 72.7          \\ 
LFM~\cite{lfm_ref}                                                                   & \multicolumn{1}{l|}{78.6}          & 39.8          & \multicolumn{1}{l|}{24.3}          & 88.3          \\ 
MODNet~\cite{modnet_ref}                                                               & \multicolumn{1}{l|}{35.9}          & 14.6          & \multicolumn{1}{l|}{67.7}          & 145.7         \\ 
P3MNet~\cite{rethink_p3m}                                                                & \multicolumn{1}{l|}{36.5}          & 18.6          & \multicolumn{1}{l|}{19.3}          & 80.7          \\ 
\begin{tabular}[c]{@{}l@{}}MGMatting(official weight)\end{tabular} & \multicolumn{1}{l|}{28.6}          & 9.39          & \multicolumn{1}{l|}{17.0}          & 55.6          \\ 
\begin{tabular}[c]{@{}l@{}}MGMatting(matting only)\end{tabular}    & \multicolumn{1}{l|}{27.9}          & 9.55          & \multicolumn{1}{l|}{16.8}          & 58.0          \\ \hline
Ours                                                                  & \multicolumn{1}{l|}{\textbf{24.6}} & \textbf{9.26} & \multicolumn{1}{l|}{\textbf{16.1}} & \textbf{55.3} \\ \hline
\end{tabular}
\end{center}
 \vspace{-1em}
\end{table}

\begin{table}[tb]
\begin{center}
\caption{Results on AIM-500~\cite{aim} benchmark.}
\label{tab:aim}
\begin{tabular}{l|l|l|l|l}
\hline
Model     & SAD           & MSE($10^{-3}$)           & GRAD          & CONN          \\ \hline
Context-Aware~\cite{context}  & 32.2          & 38.8          & 30.3          & 31.0         \\ 
GFM~\cite{AM-2k}       & 52.7          & 21.3          & 46.1          & 52.7          \\ 
AIMNet~\cite{aim}    & 43.9          & 16.1          & 33.1          & 43.2          \\
MGMatting~\cite{mgm_ref} & 26.2          & 5.60          & 15.8          & 14.5          \\ 
MG-Wild~\cite{mgw}   & 16.7          & 3.00          & 14.7          & 12.0          \\ \hline
Ours      & \textbf{14.3} & \textbf{2.67} & \textbf{12.4} & \textbf{11.2} \\ \hline
\end{tabular}
\end{center}
\vspace{-1em}
\end{table}

\begin{table}[tb]
\begin{center}
\caption{Results on AM-2k~\cite{AM-2k} benchmark.}
\label{tab:am}
\begin{tabular}{l|l|l|l|l}
\hline
Model                                                              & SAD  & MSE($10^{-4}$)  & GRAD & CONN \\ \hline
AIMNet~\cite{aim}    & 27.5          & 10.0         & 17.9          & 12.2          \\
SHM~\cite{shm}                                                                & 17.8 & 68.0 & 12.5 & 17.0 \\ 
GFM~\cite{AM-2k}                                                                & 10.3 & 29.0 & 8.82 & 9.57 \\ 
MGMatting~\cite{mgm_ref}                                                          & 10.1 & 10.4 & 5.58 & 7.11 \\ 
\begin{tabular}[c]{@{}l@{}}MGMatting (matting only)\end{tabular} & 8.35 & 8.07 & 4.92 & 6.58 \\ \hline
Ours                                                               & \textbf{5.86} &\textbf{5.95} & \textbf{3.67} & \textbf{4.55} \\ \hline
\end{tabular}
\end{center}
 \vspace{-1em}
\end{table}

\begin{table}[tb]
\begin{center}
\caption{Results on PPM-100~\cite{modnet_ref} benchmark.}
\label{tab:ppm}
\begin{tabular}{l|l|l|l|l}
\hline
Model                                                              & SAD  & MSE($10^{-4}$)  & GRAD & CONN \\ \hline
AIMNet~\cite{aim}    & 201.9         & 185.9        & 79.63       &68.21         \\
P3MNet~\cite{rethink_p3m} &130.8 &128.6 &56.37 &130.4          \\
MODNet~\cite{modnet_ref} &95.1 &44.7 &64.26 &80.82\\
RVM~\cite{rvm} &108.2 &65.3 &63.13 &105.2 \\
MGMatting & 67.6 & 18.9 &37.46 &29.48\\
MGMatting(matting only) &40.0 & 8.80 &36.17 &28.50\\ \hline
Ours & \textbf{30.9} & \textbf{7.79} & \textbf{32.76} & \textbf{21.74} \\ \hline
\end{tabular}
\end{center}
 \vspace{-1em}
\end{table}

\begin{table}[tb]
\begin{center}
\caption{Ablation study on PPM-100~\cite{modnet_ref} benchmark.}
\label{tab:ppm_ablation}
\begin{tabular}{l|l|l|l|l}
\hline
Model                                                              & SAD  & MSE($10^{-4}$)  & GRAD & CONN \\ \hline
Matting only &40.0 & 8.80 &36.17 &28.50\\

Ours(RASR)    & 35.4          & 8.64          & 33.45          & 23.88          \\ 
Ours(RASR+IG)   & 32.3          & 8.00          & 32.95          & 21.97          \\ 
Ours(RASR/IG/LD) & \textbf{30.9} & \textbf{7.79} & \textbf{32.76} & \textbf{21.74} \\ \hline
\end{tabular}
\end{center}
 \vspace{-1em}
\end{table}

\begin{table}[tb]
\vspace{-1em}
\begin{center}
\caption{Results of mask-guided matting methods on our Plant-Mat benchmark.}
\label{tab:plant}
\begin{tabular}{l|l|l|l|l}
\hline
Methods                                                               & SAD  & MSE($10^{-3}$)&GRAD&CONN \\ \hline
MGMatting~\cite{mgm_ref}& 97.4 & 3.36 &  49.8&   19.5       \\ \hline
Matting only                                                          & 60.0 & 2.17  &48.0&18.4         \\ 
Ours(RASR)                                                              &  45.7    &     1.65  &45.6 & 18.1         \\ 
Ours(RASR/IG)                                                           &  31.2    &     1.42   &42.5 & 17.5        \\ 
Ours(RASR/IG/LD)                                                        &   \textbf{24.0 }  &       \textbf{1.34} & \textbf{39.3} & \textbf{17.1}      \\ \hline
\end{tabular}
\end{center}
\vspace{-1em}
\end{table}

\begin{figure*}[!t]
  \centering
  \scriptsize
  \setlength{\tabcolsep}{0.0pt} 
  
  \begin{tabular}{ccccccc}
     \includegraphics[width=0.142\textwidth]{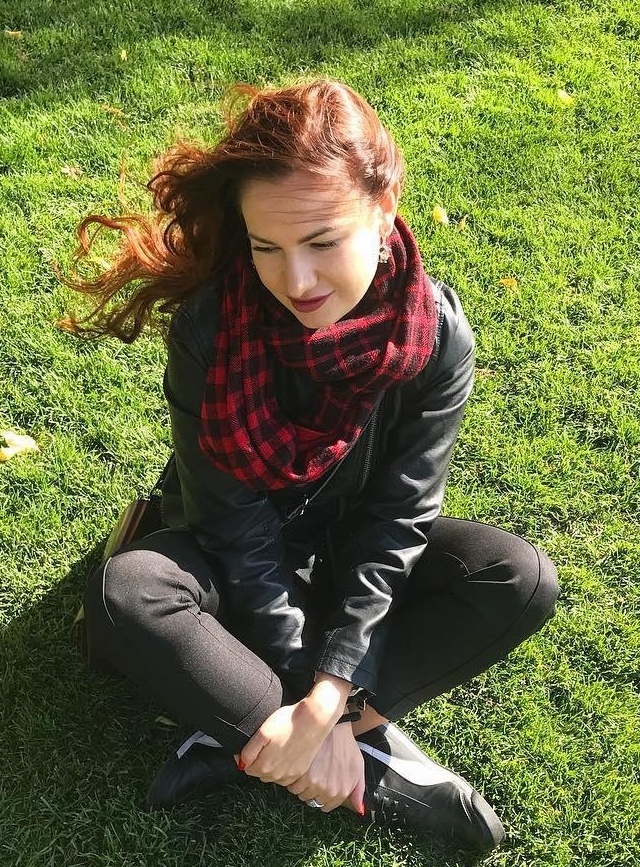} &
     \includegraphics[width=0.142\textwidth]{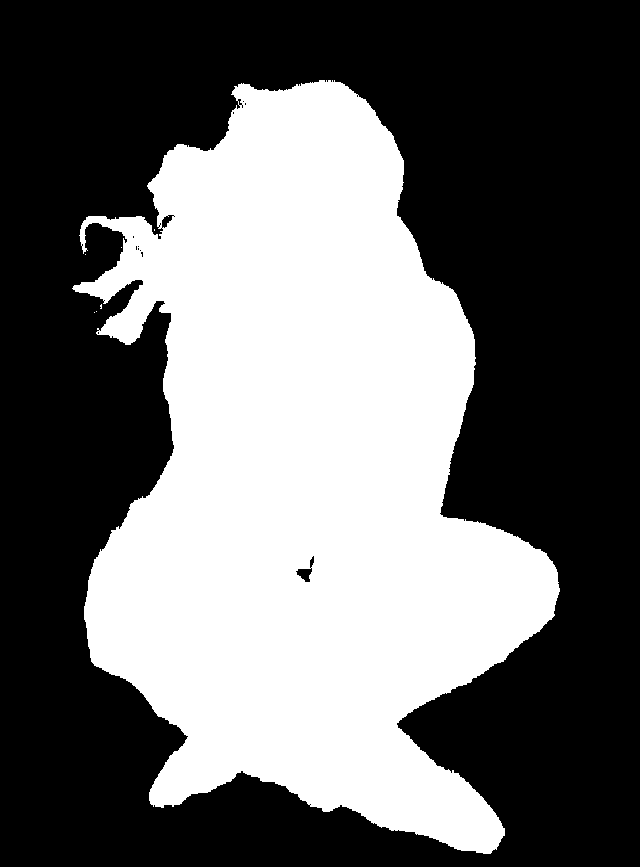} &
    \includegraphics[width=0.142\textwidth]{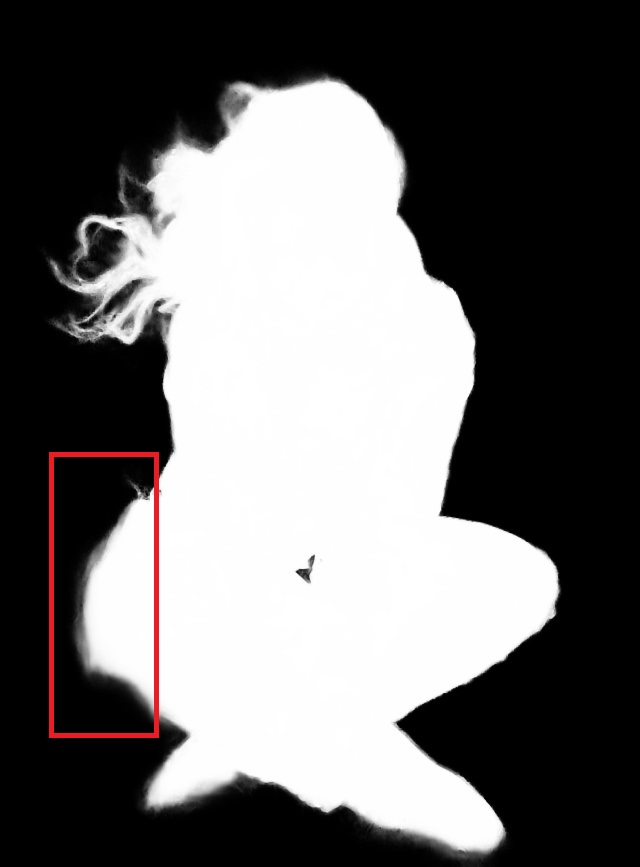} &
     \includegraphics[width=0.142\textwidth]{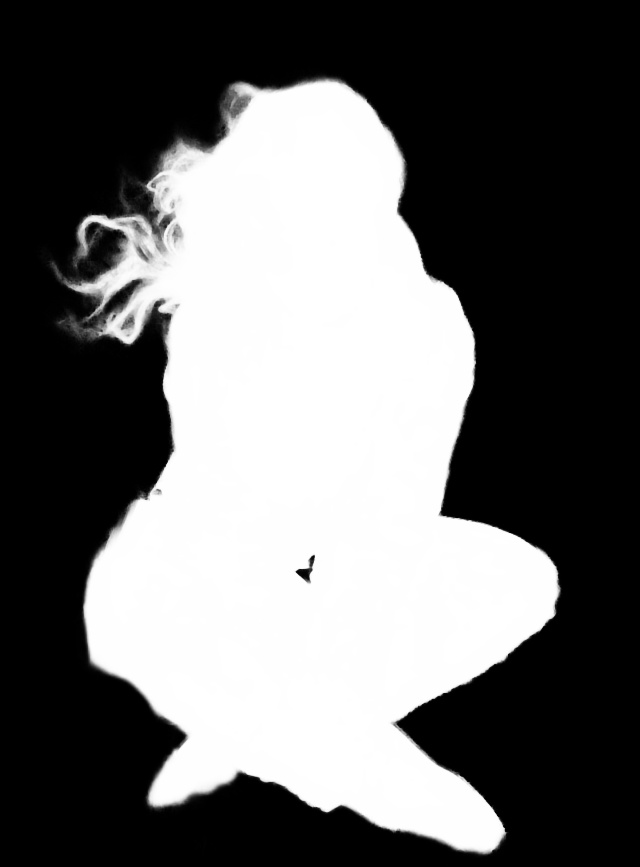} &
     
     \includegraphics[width=0.142\textwidth]{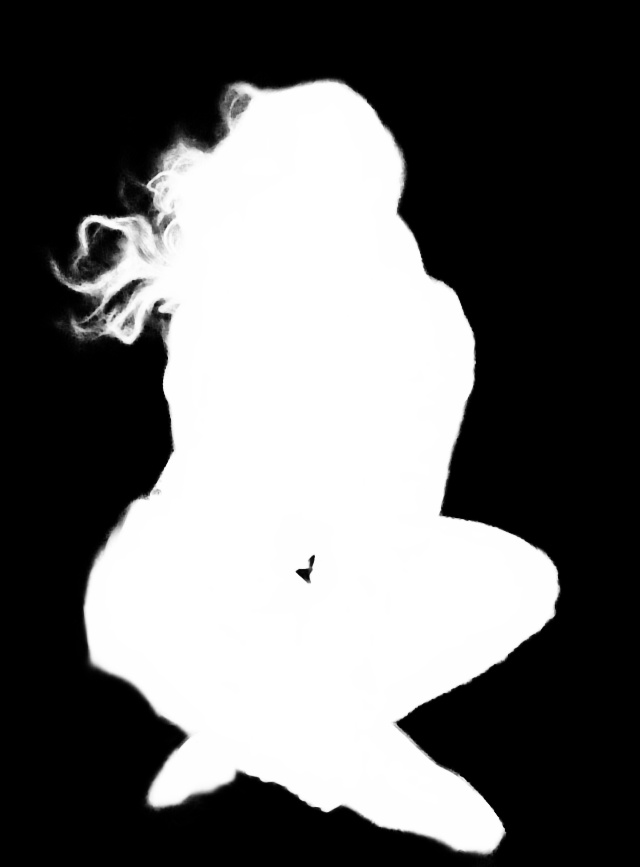} &
     \includegraphics[width=0.142\textwidth]{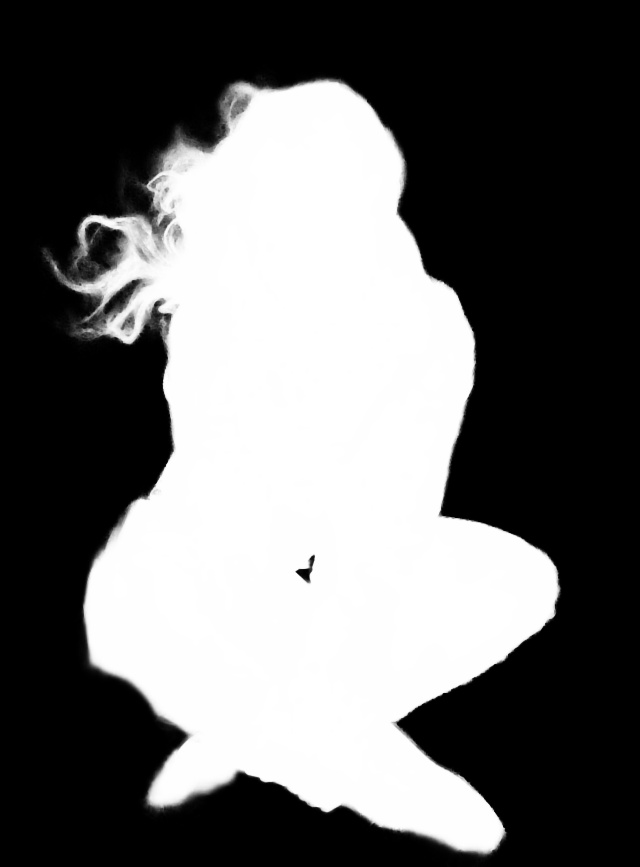} &
     \includegraphics[width=0.142\textwidth]{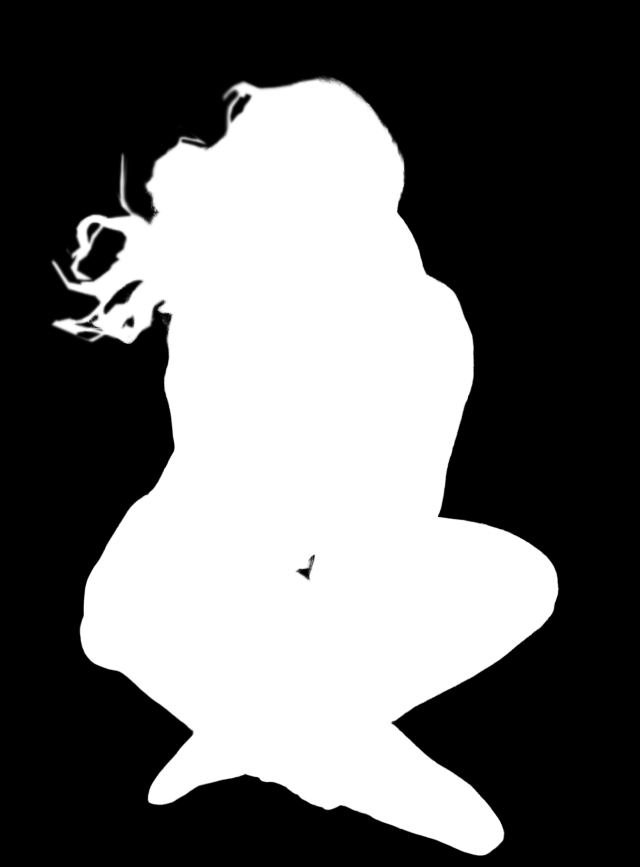}
     \\
     \includegraphics[trim={0 6cm 0 1cm},clip,width=0.142\textwidth]{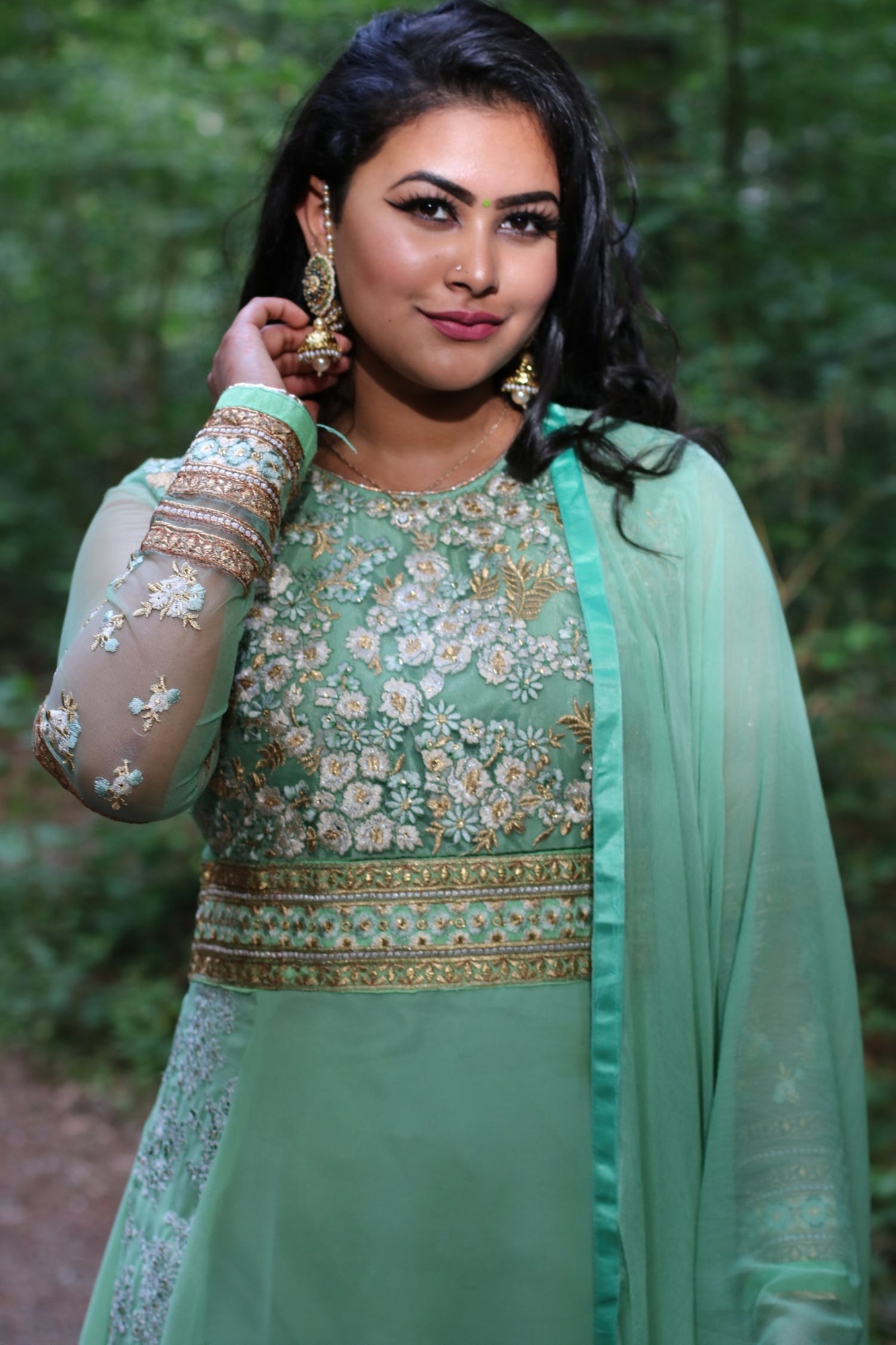} &
     \includegraphics[trim={0 6cm 0 1cm},clip,width=0.142\textwidth]{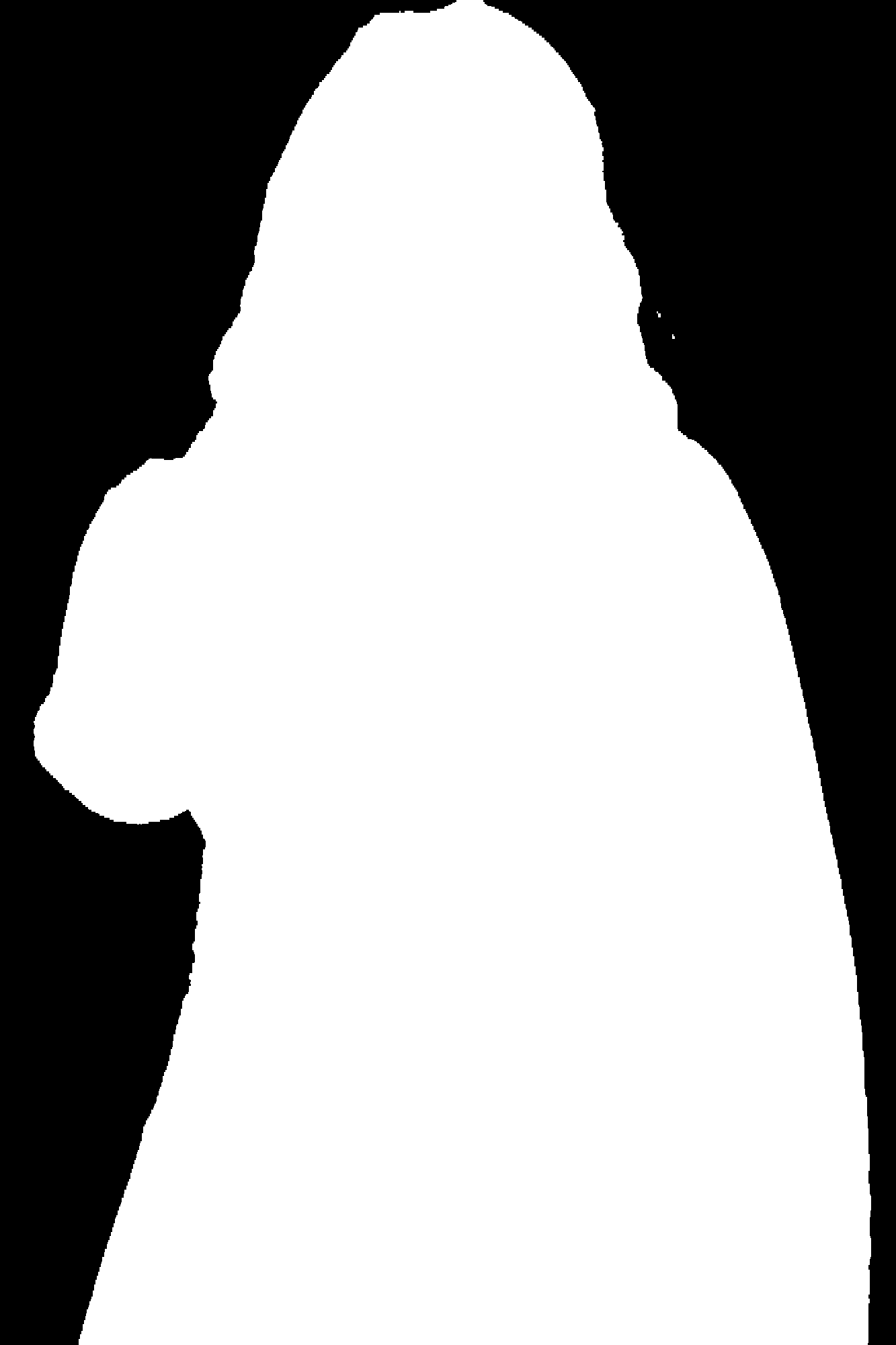} &
     \includegraphics[trim={0 6cm 0 1cm},clip,width=0.142\textwidth]{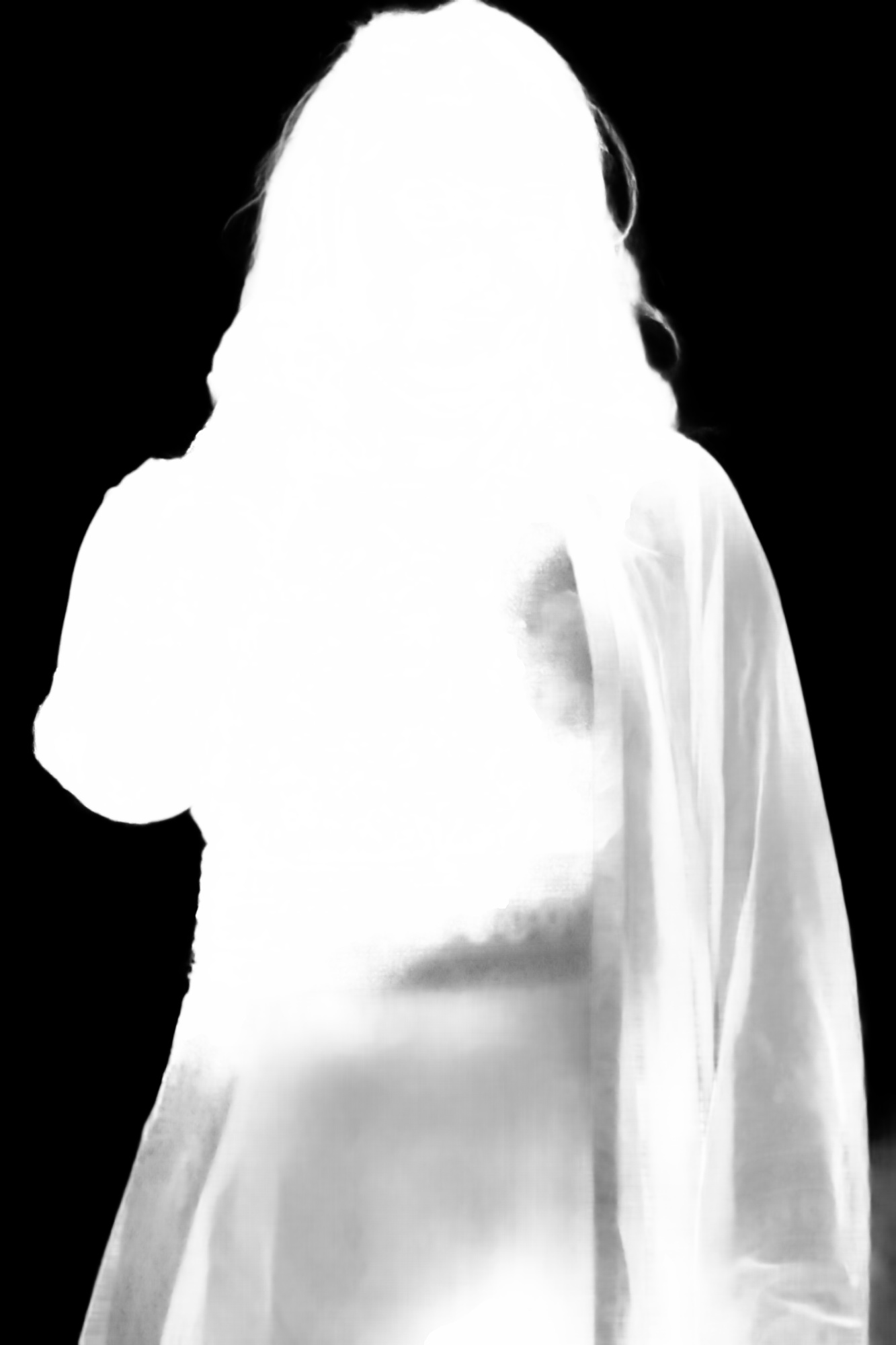} &
     \includegraphics[trim={0 6cm 0 1cm},clip,width=0.142\textwidth]{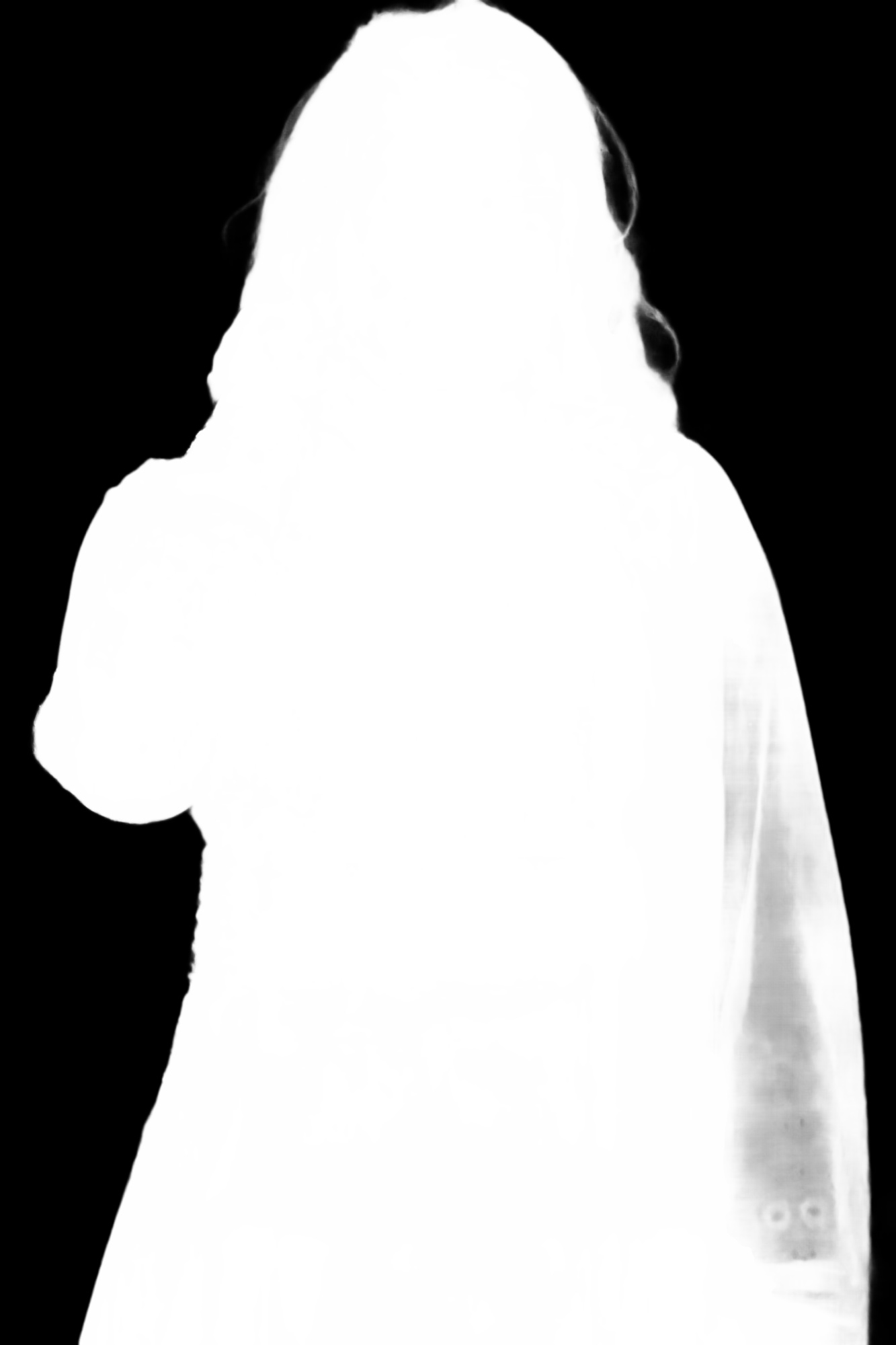} &
     \includegraphics[trim={0 6cm 0 1cm},clip,width=0.142\textwidth]{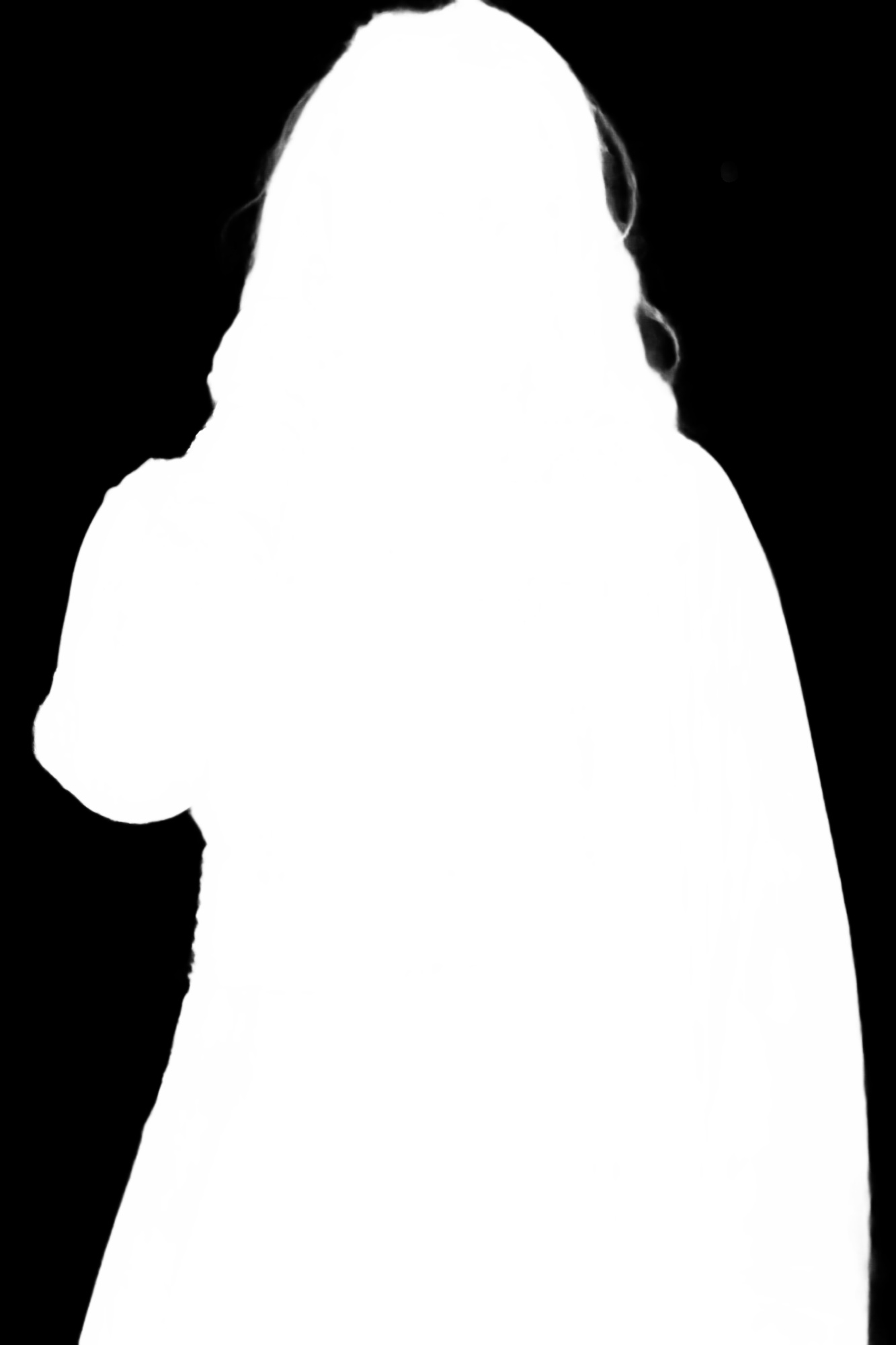} &
     \includegraphics[trim={0 6cm 0 1cm},clip,width=0.142\textwidth]{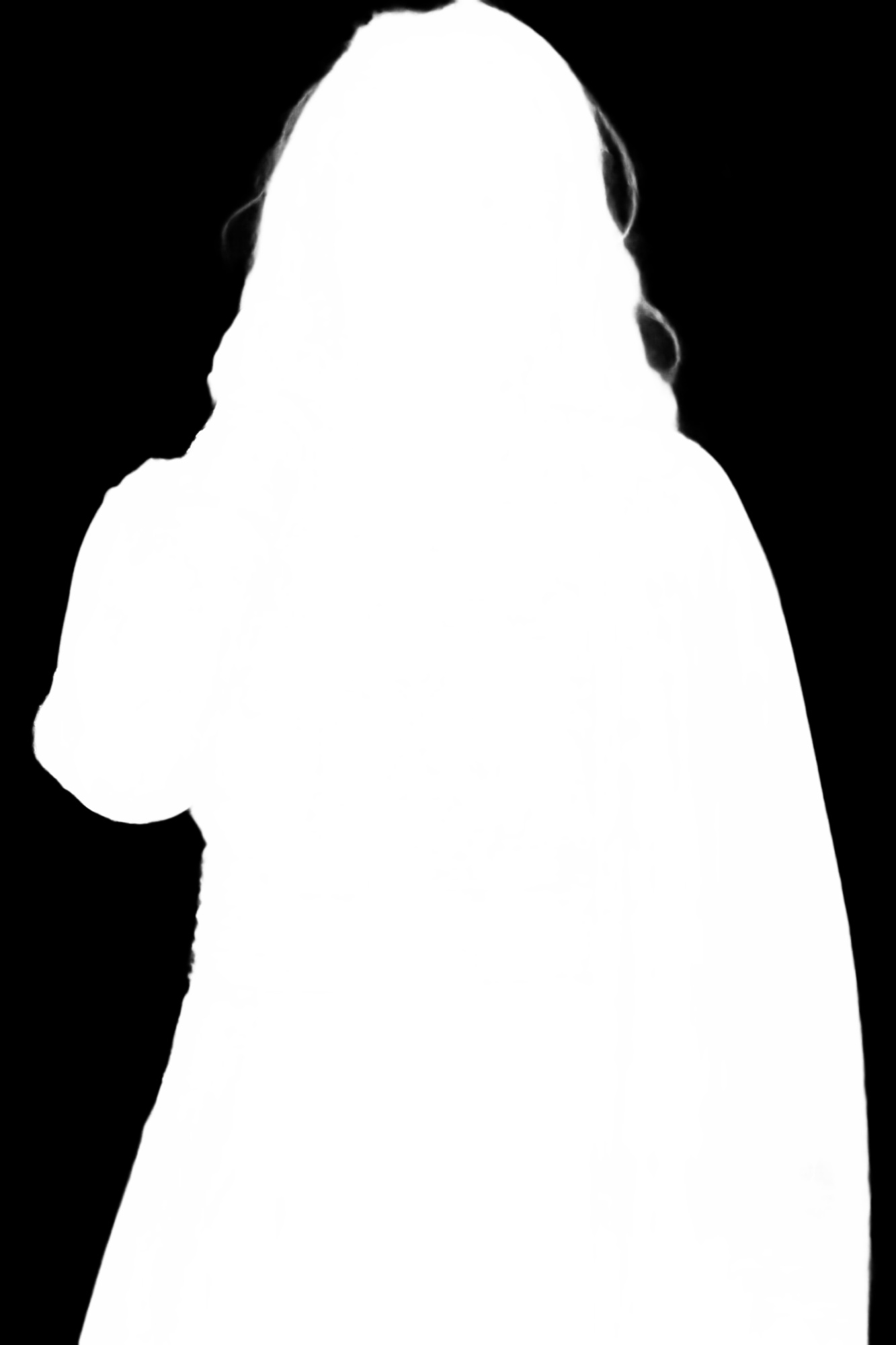} &
     \includegraphics[trim={0 6cm 0 1cm},clip,width=0.142\textwidth]{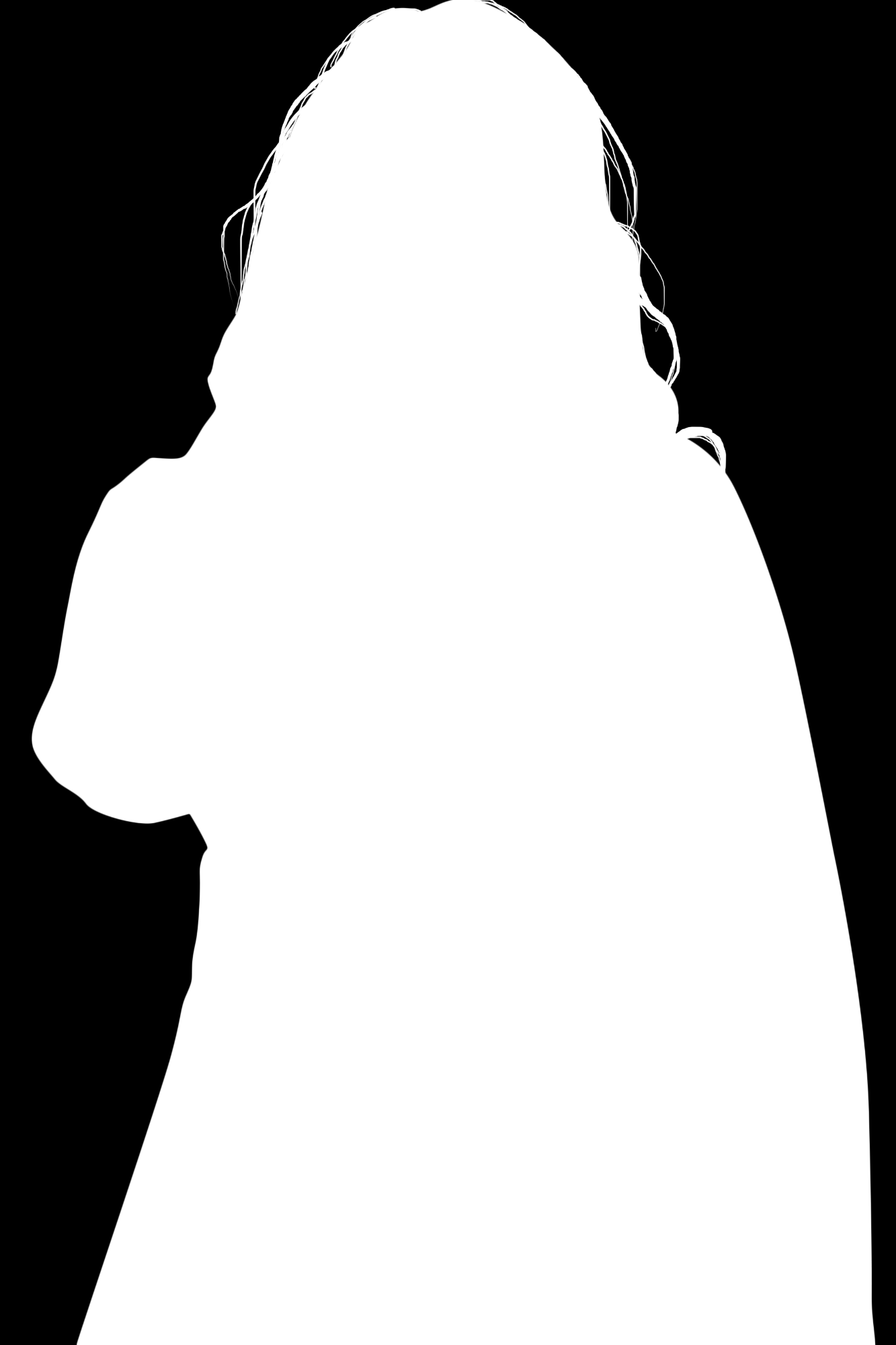}
     \\
     \includegraphics[width=0.142\textwidth]{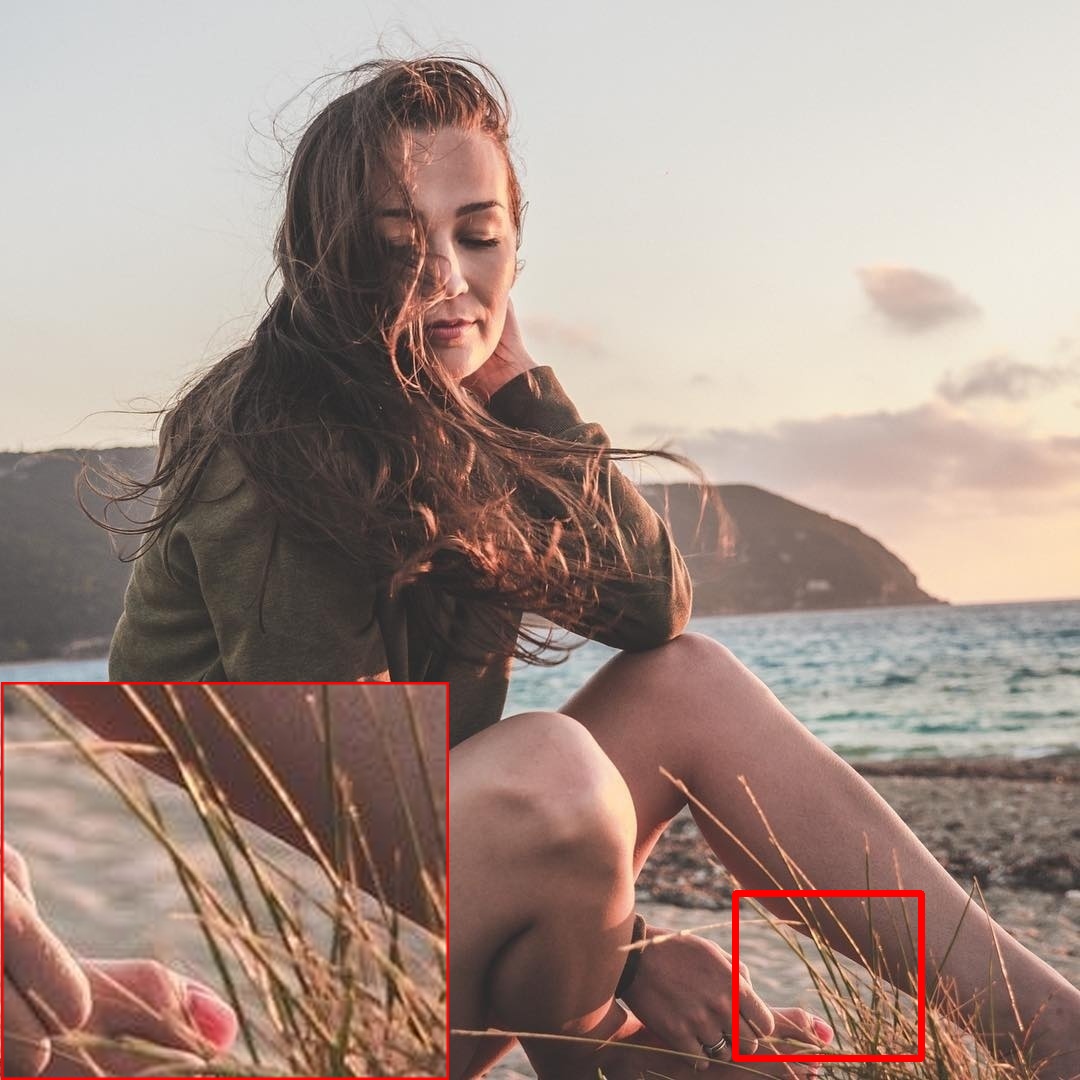} &
     \includegraphics[width=0.142\textwidth]{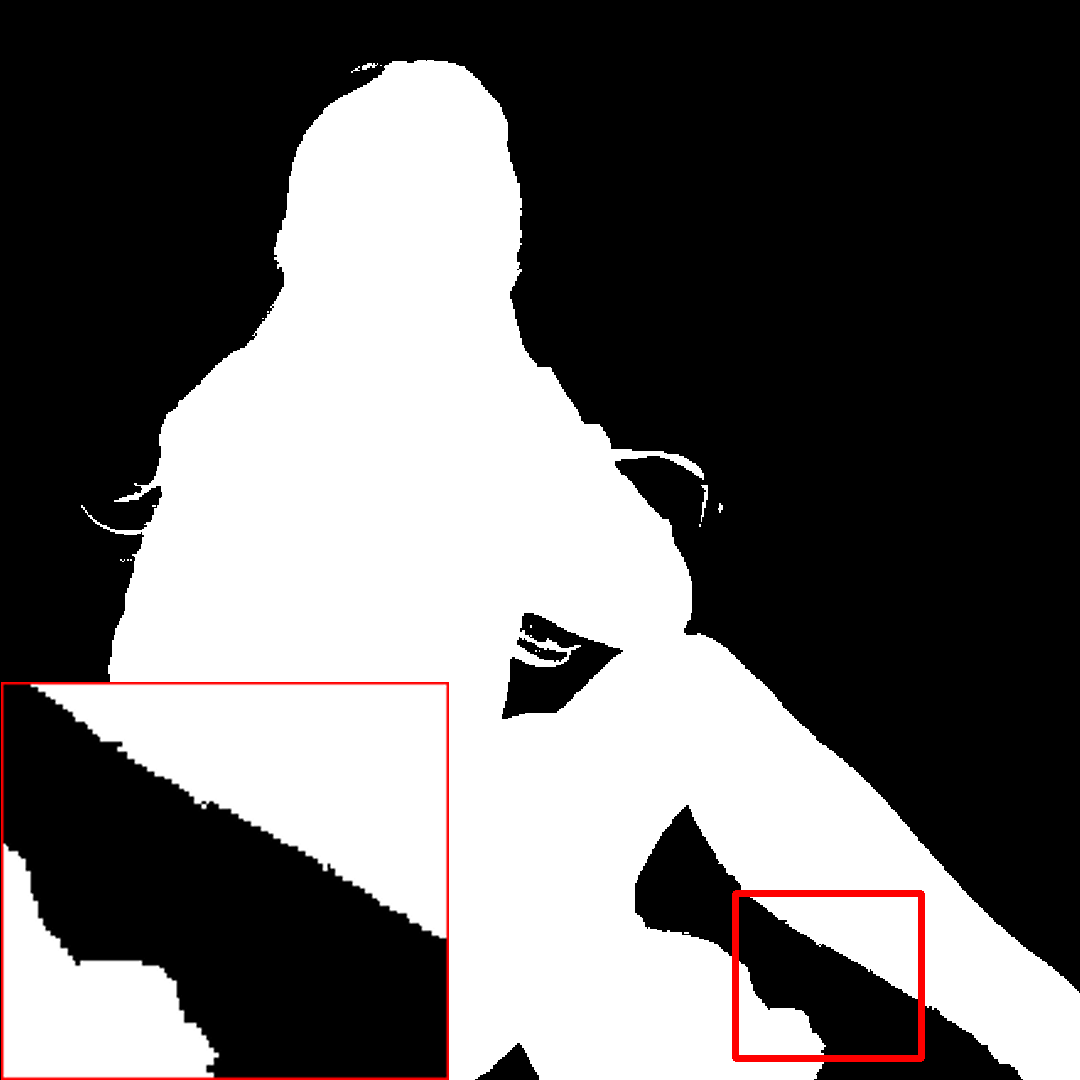} &
     \includegraphics[width=0.142\textwidth]{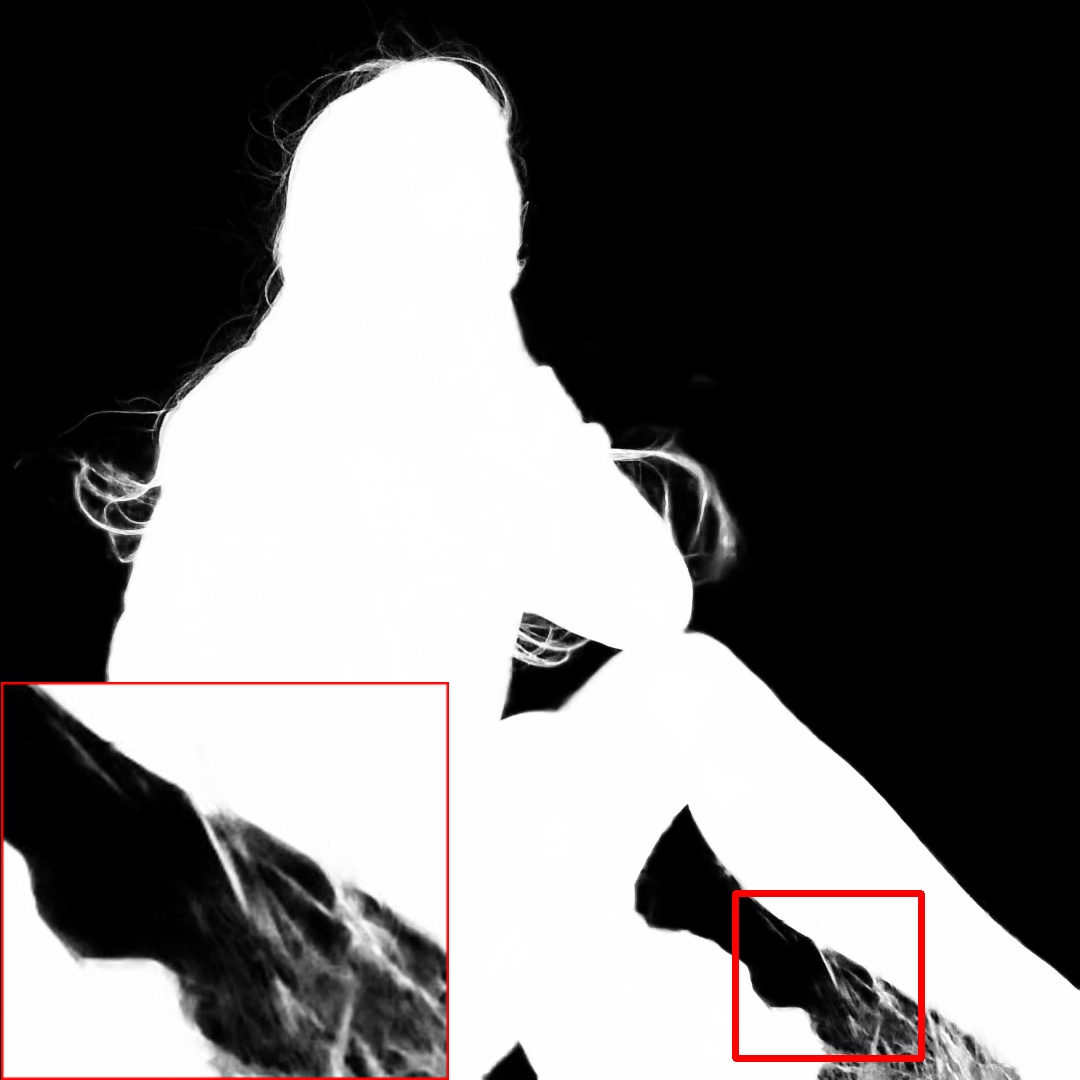} &
     \includegraphics[width=0.142\textwidth]{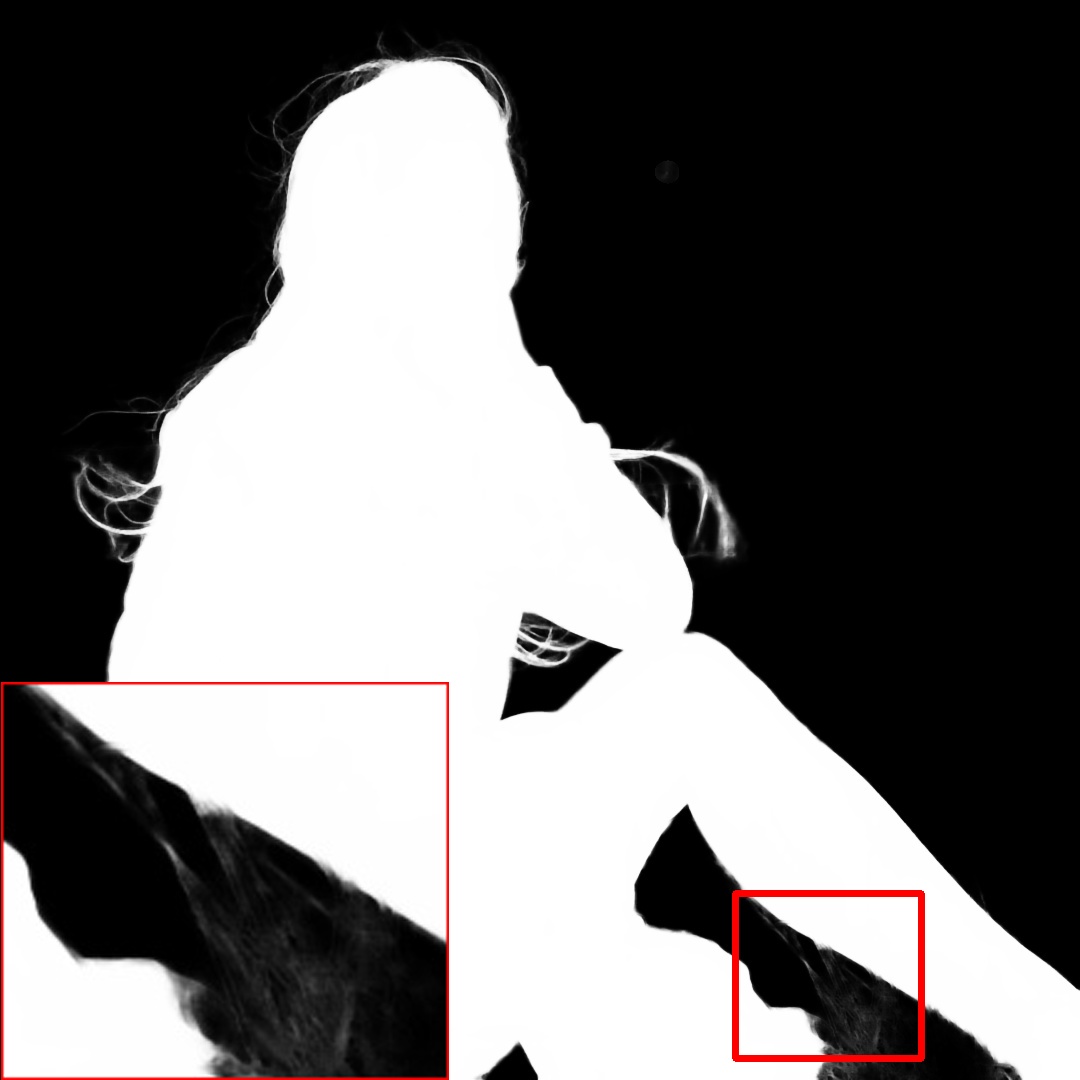} &
     \includegraphics[width=0.142\textwidth]{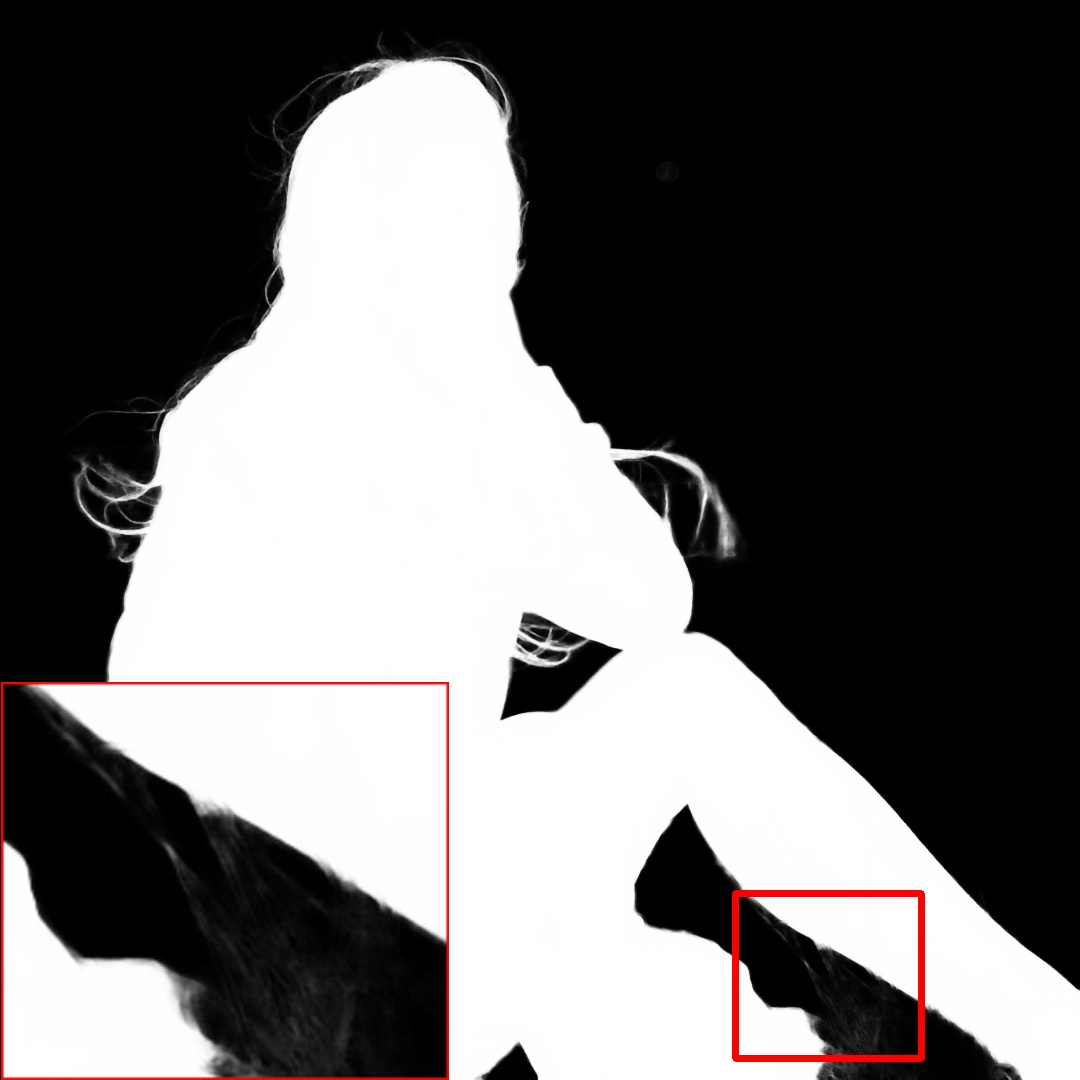} &
     \includegraphics[width=0.142\textwidth]{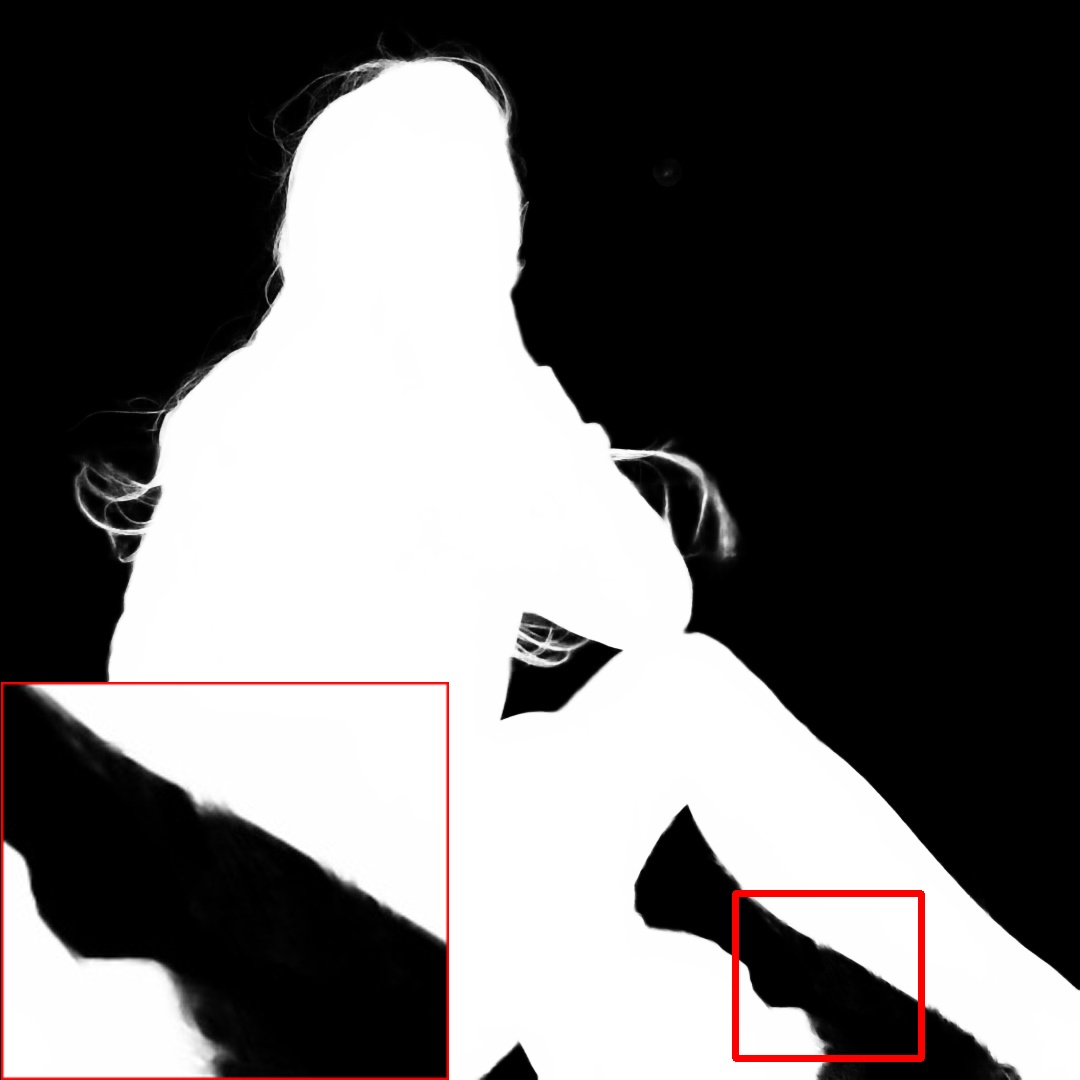} &
     \includegraphics[width=0.142\textwidth]{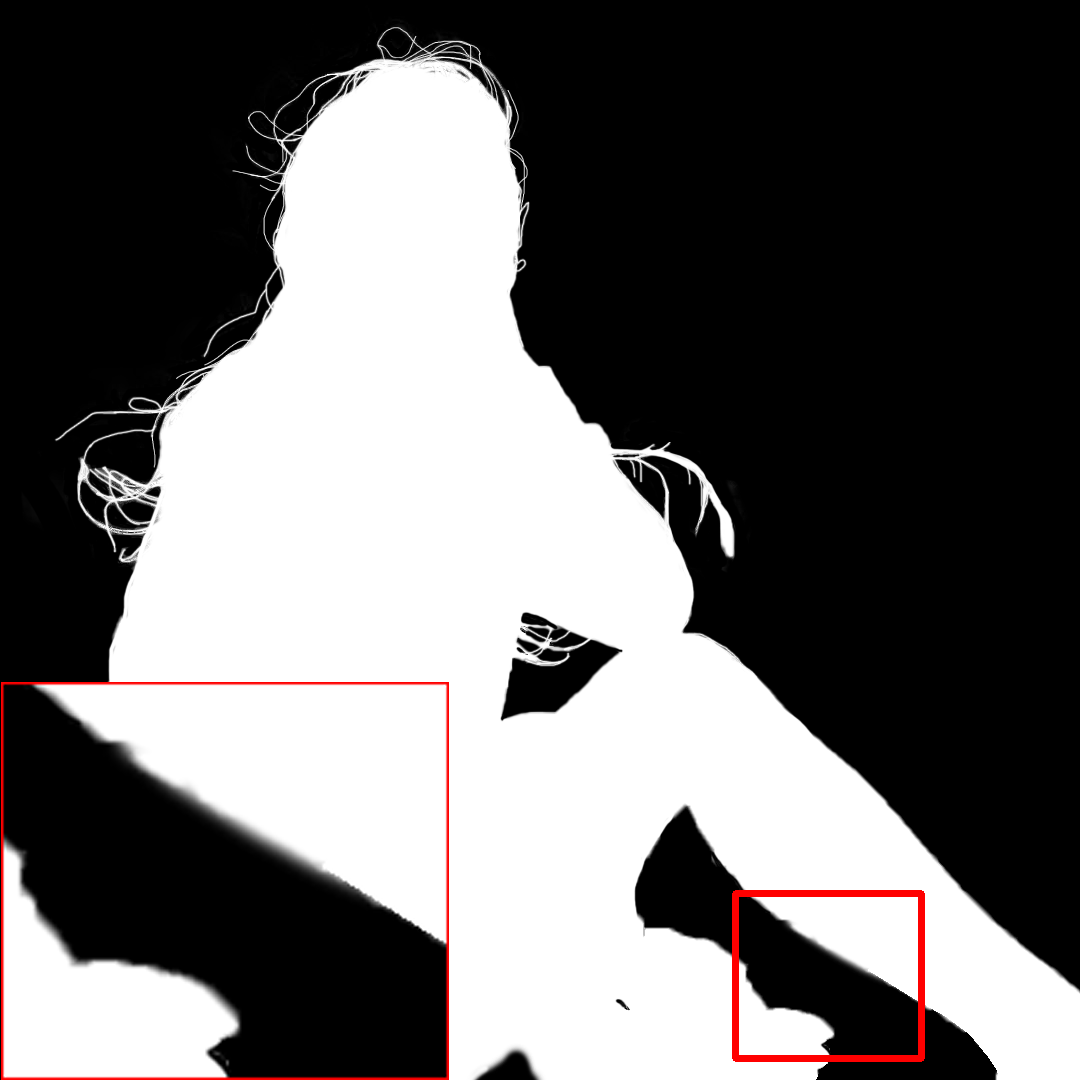}
     \\
     Image &
     Guidance &
     MGMatting &
     Ours(RASR) &
     Ours(RASR/IG) &
     Ours(RASR/IG/LD) &
     GT
  \end{tabular}
  
  \caption{The visual comparison results among different methods on real-world images. RASR: real-world adaptive semantic representation. IG: with our IGDR module. LD: auxiliary learning with background line detection as Task 4.}
  \label{fig:FinalVis}
\end{figure*}
\begin{figure*}[!t]
  \centering
  \scriptsize
  \setlength{\tabcolsep}{0.0pt} 
  
  \begin{tabular}{ccccccc}
    \includegraphics[trim={0 2.55cm 0 11cm},clip,width=0.142\textwidth]{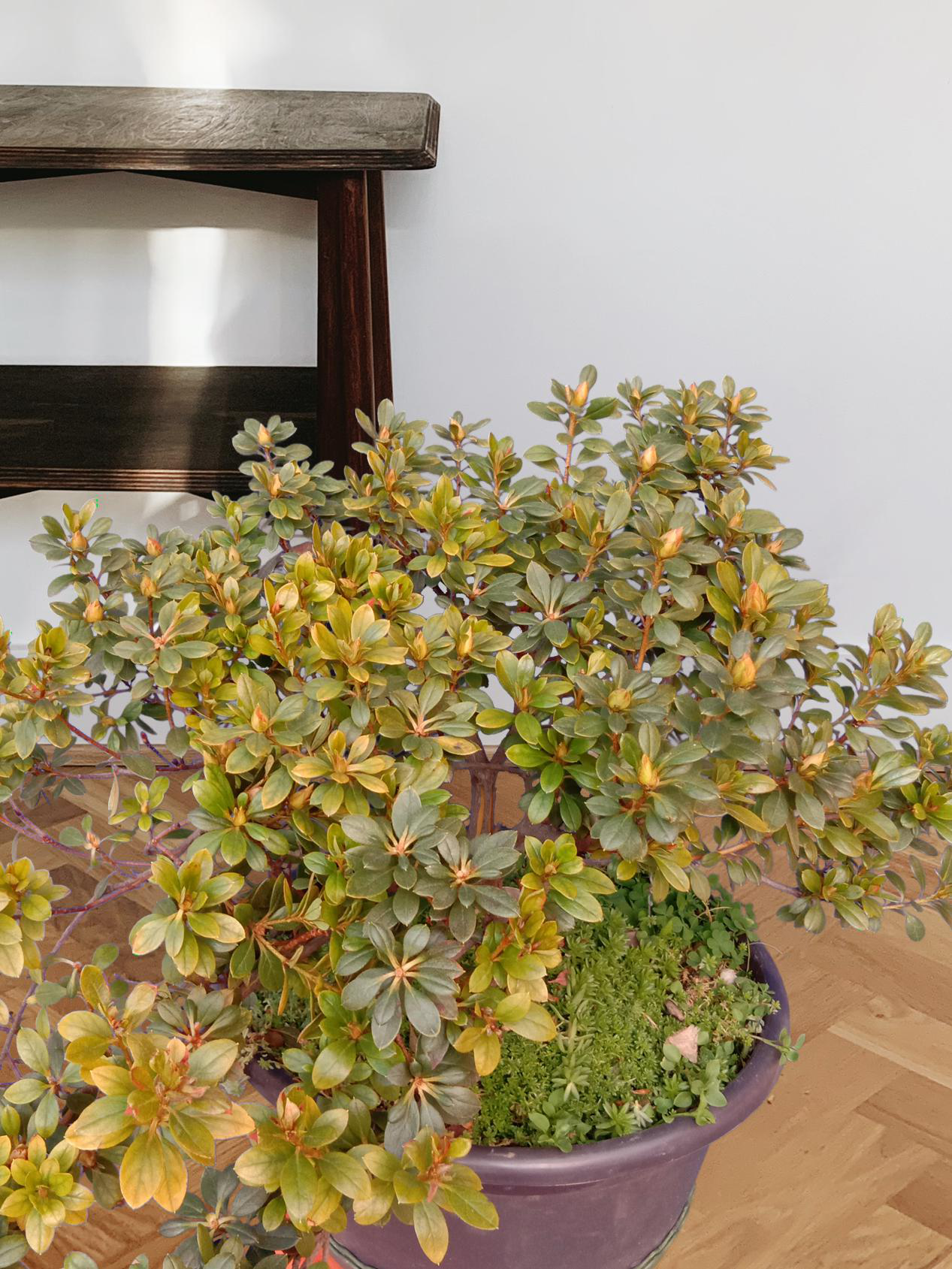} &
     \includegraphics[trim={0 2.55cm 0 11cm},clip,width=0.142\textwidth]{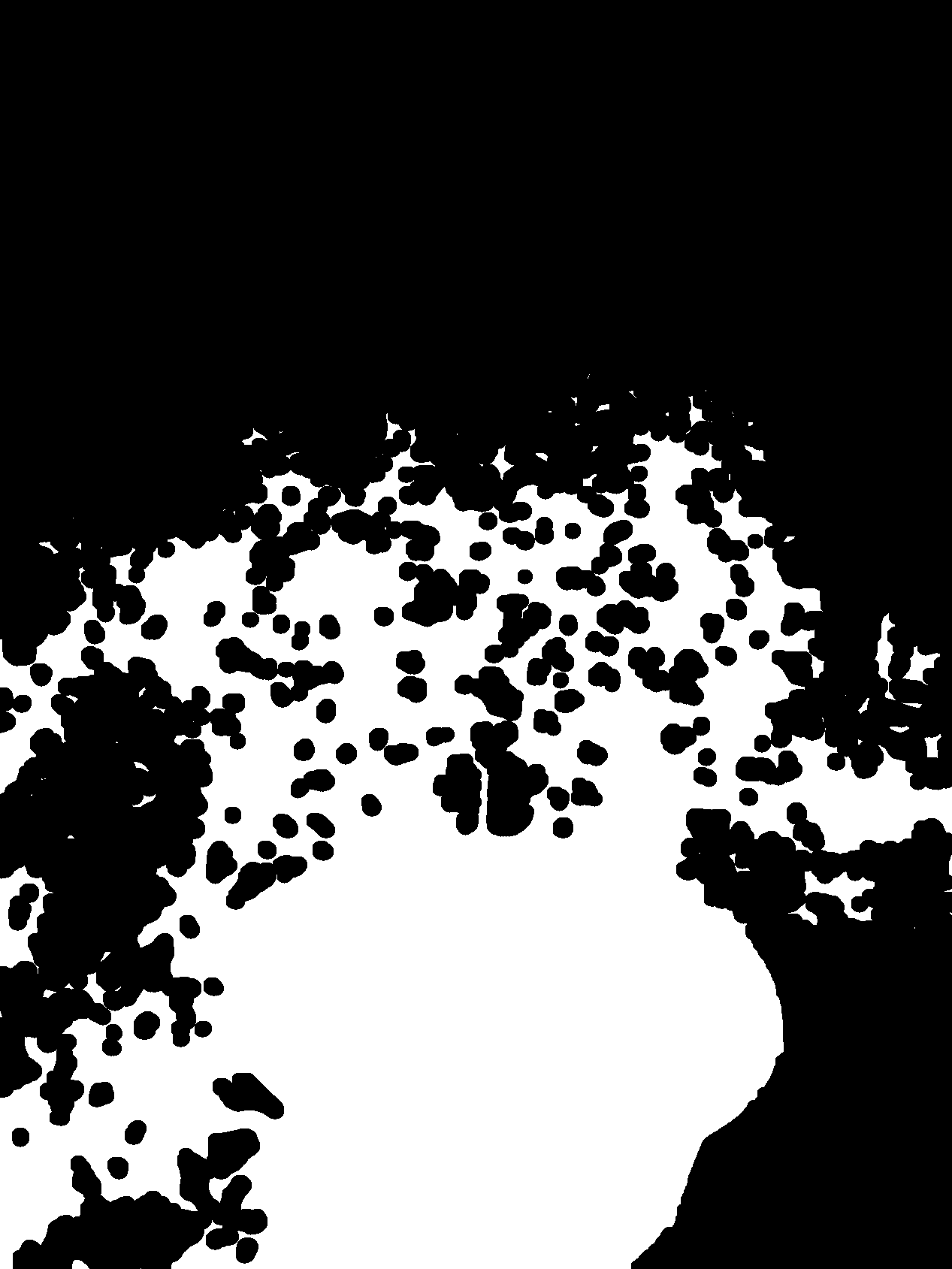} &
     \includegraphics[trim={0 2.55cm 0 11cm},clip,width=0.142\textwidth]{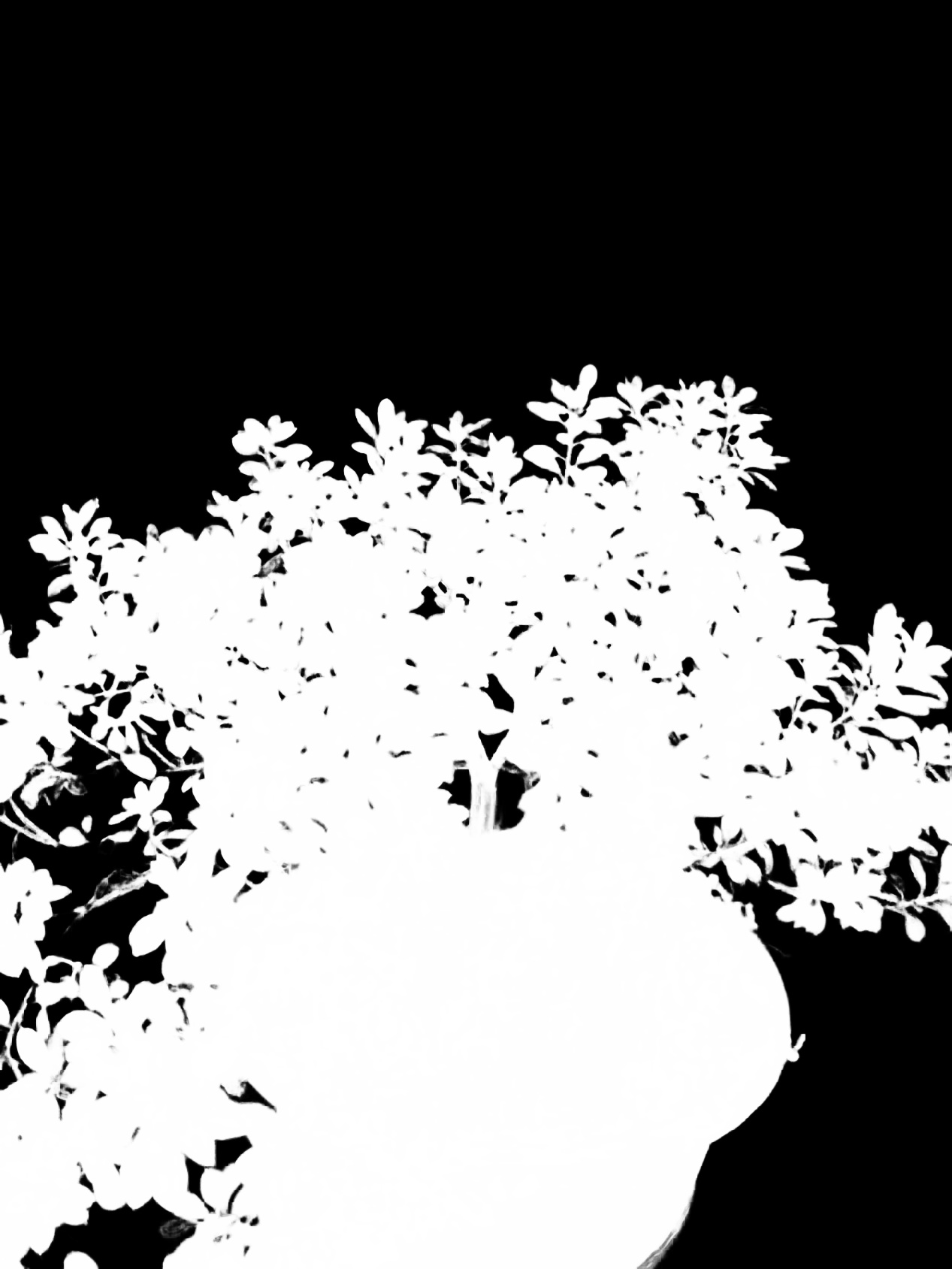} &
     \includegraphics[trim={0 2.55cm 0 11cm},clip,width=0.142\textwidth]{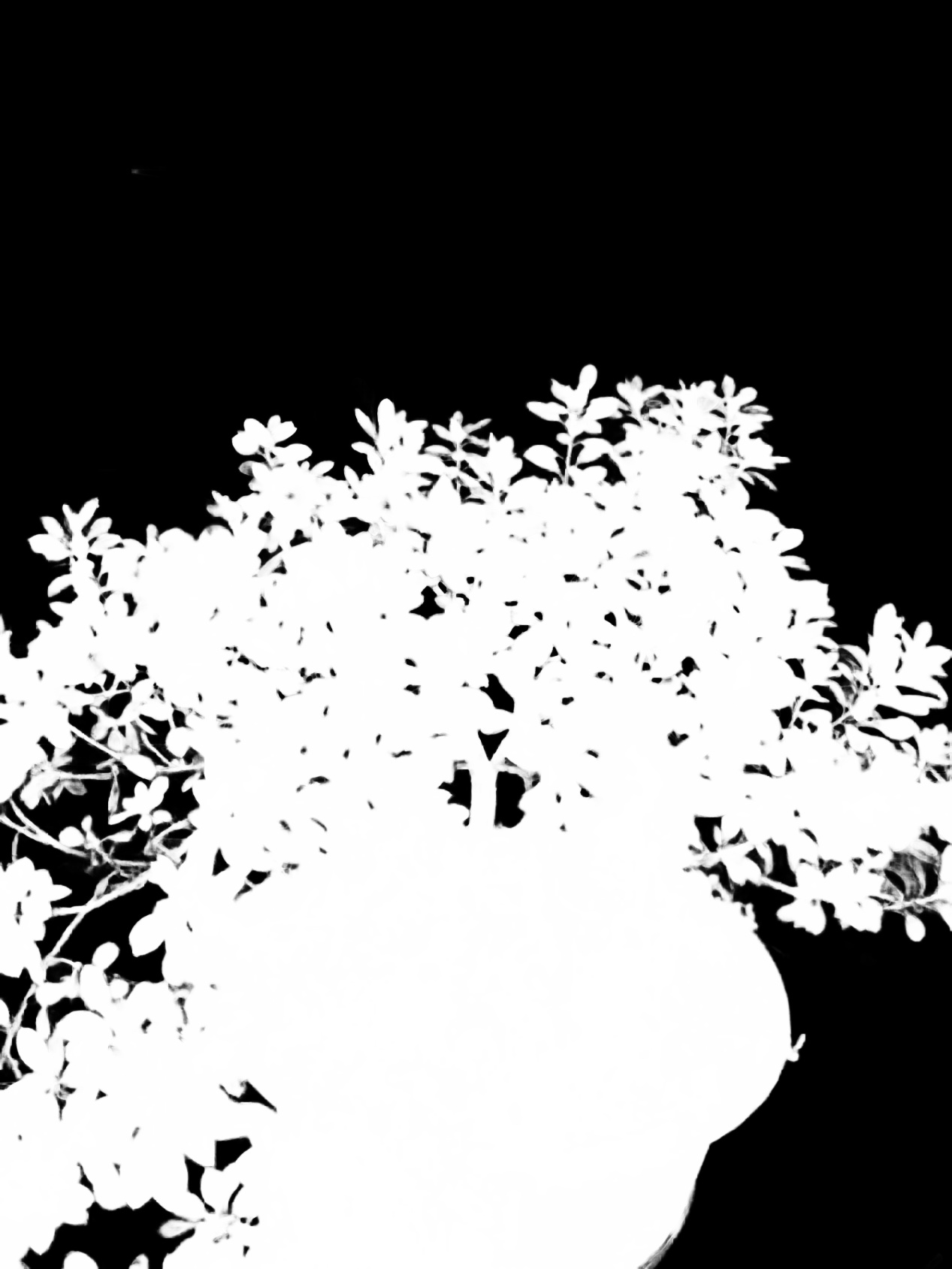} &
     \includegraphics[trim={0 2.55cm 0 11cm},clip,width=0.142\textwidth]{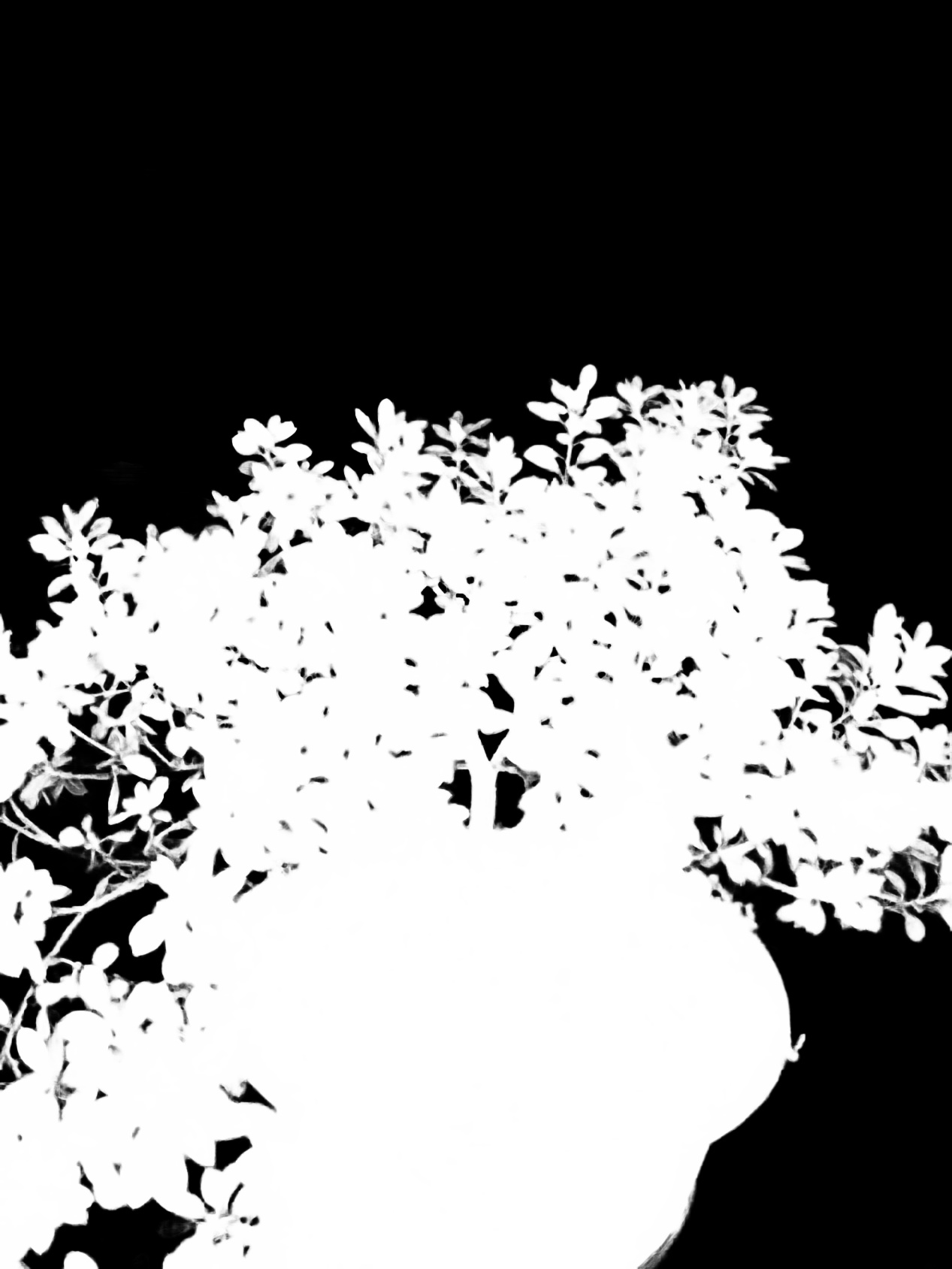} &
     \includegraphics[trim={0 2.55cm 0 11cm},clip,width=0.142\textwidth]{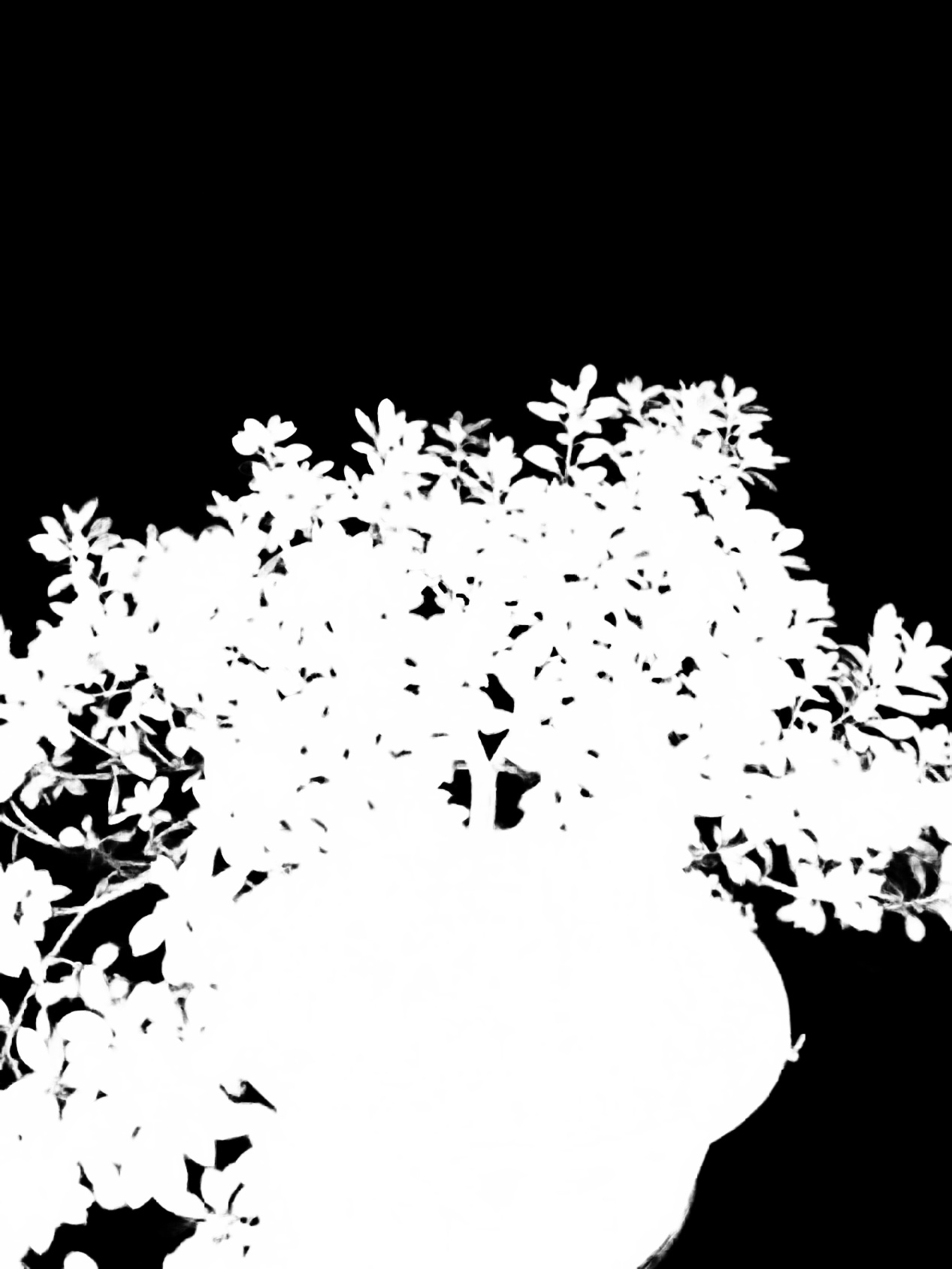} &
     \includegraphics[trim={0 2.55cm 0 11cm},clip,width=0.142\textwidth]{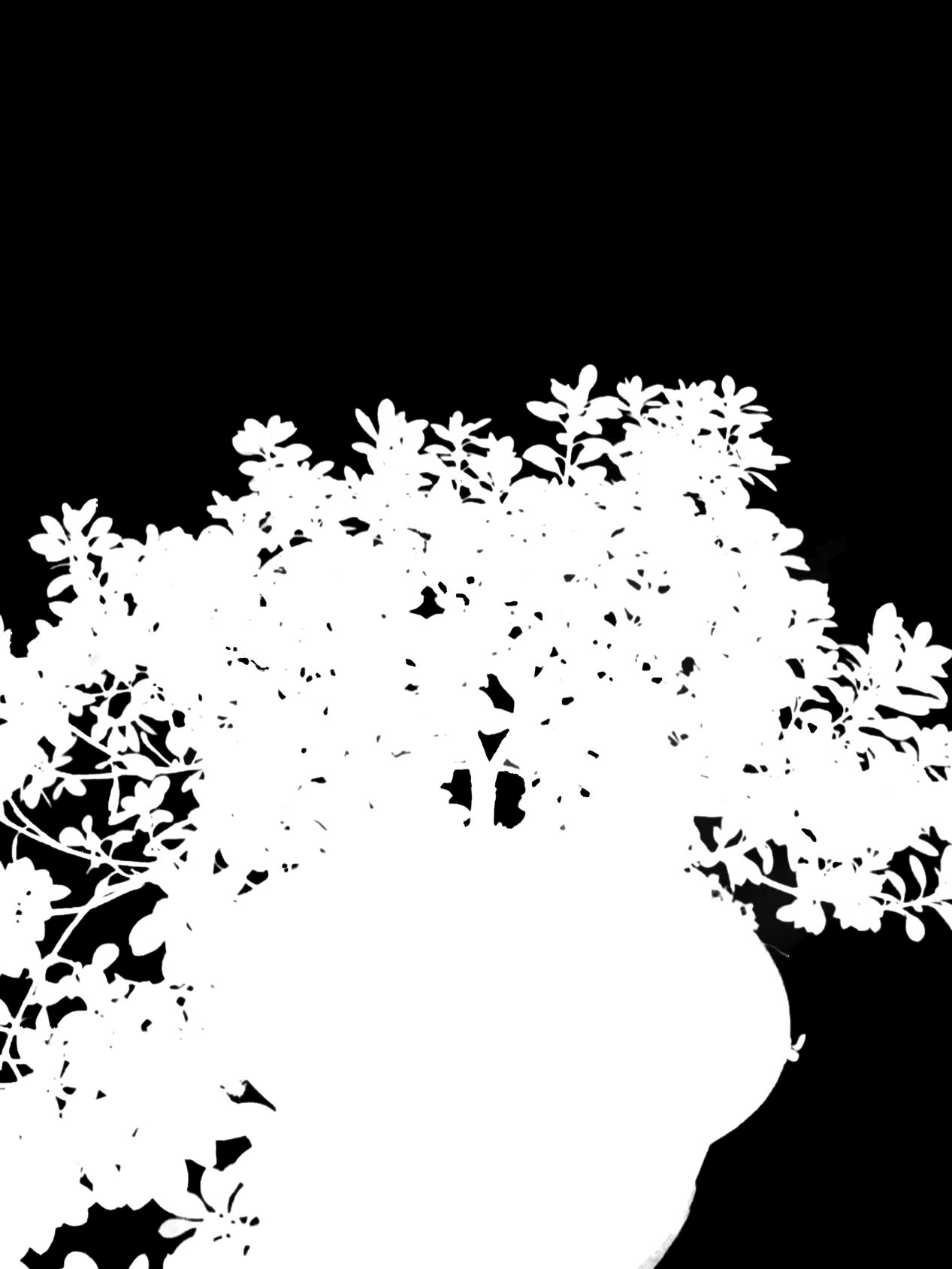}
     \\
     \includegraphics[height=0.106\textwidth,width=0.142\textwidth]{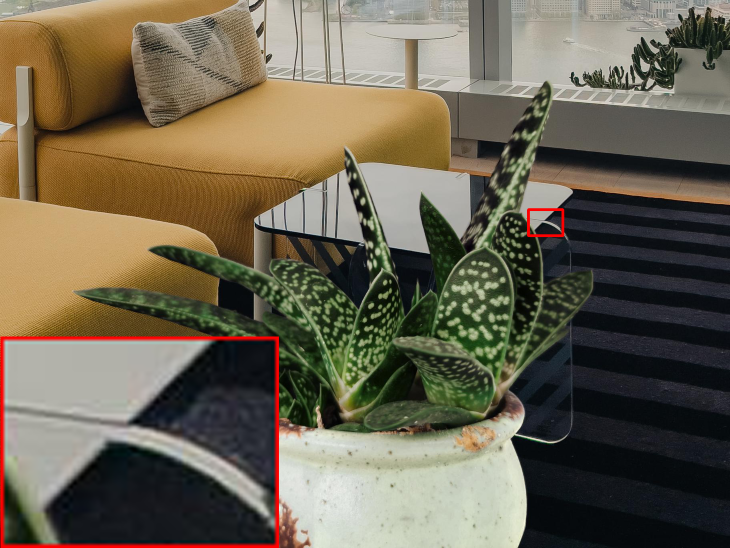} &
     \includegraphics[height=0.106\textwidth,width=0.142\textwidth]{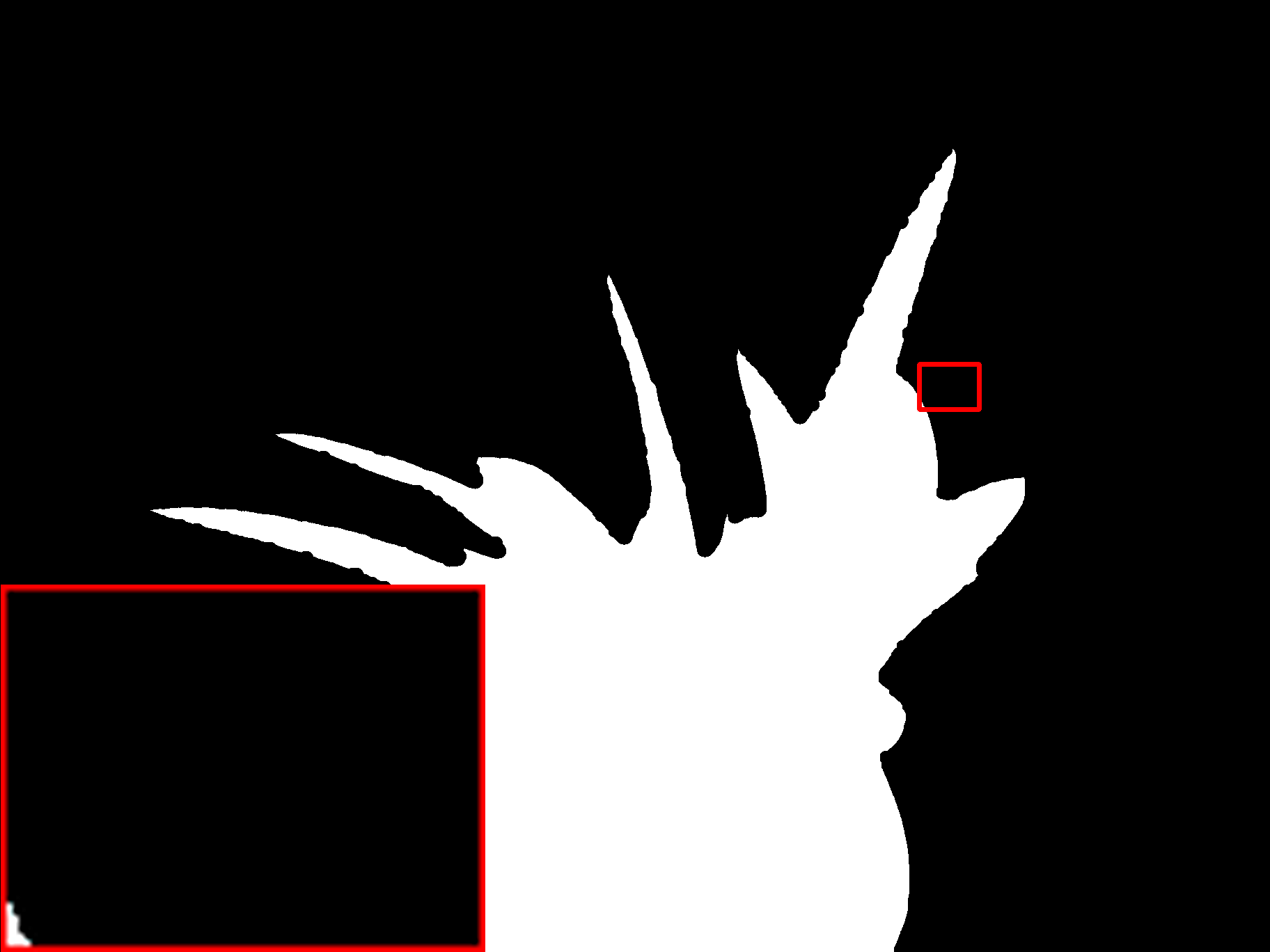} &
     \includegraphics[height=0.106\textwidth,width=0.142\textwidth]{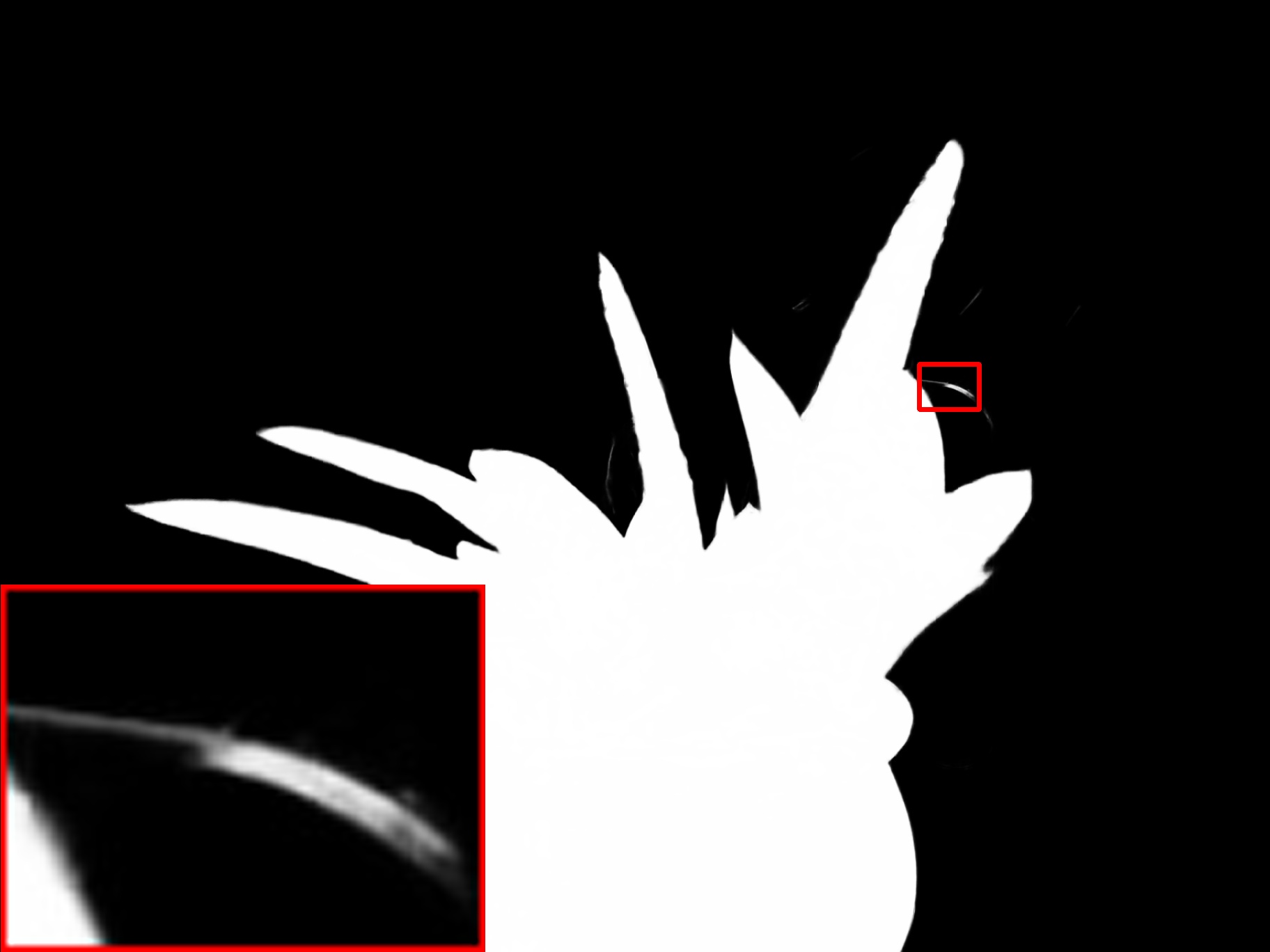} &
     \includegraphics[height=0.106\textwidth,width=0.142\textwidth]{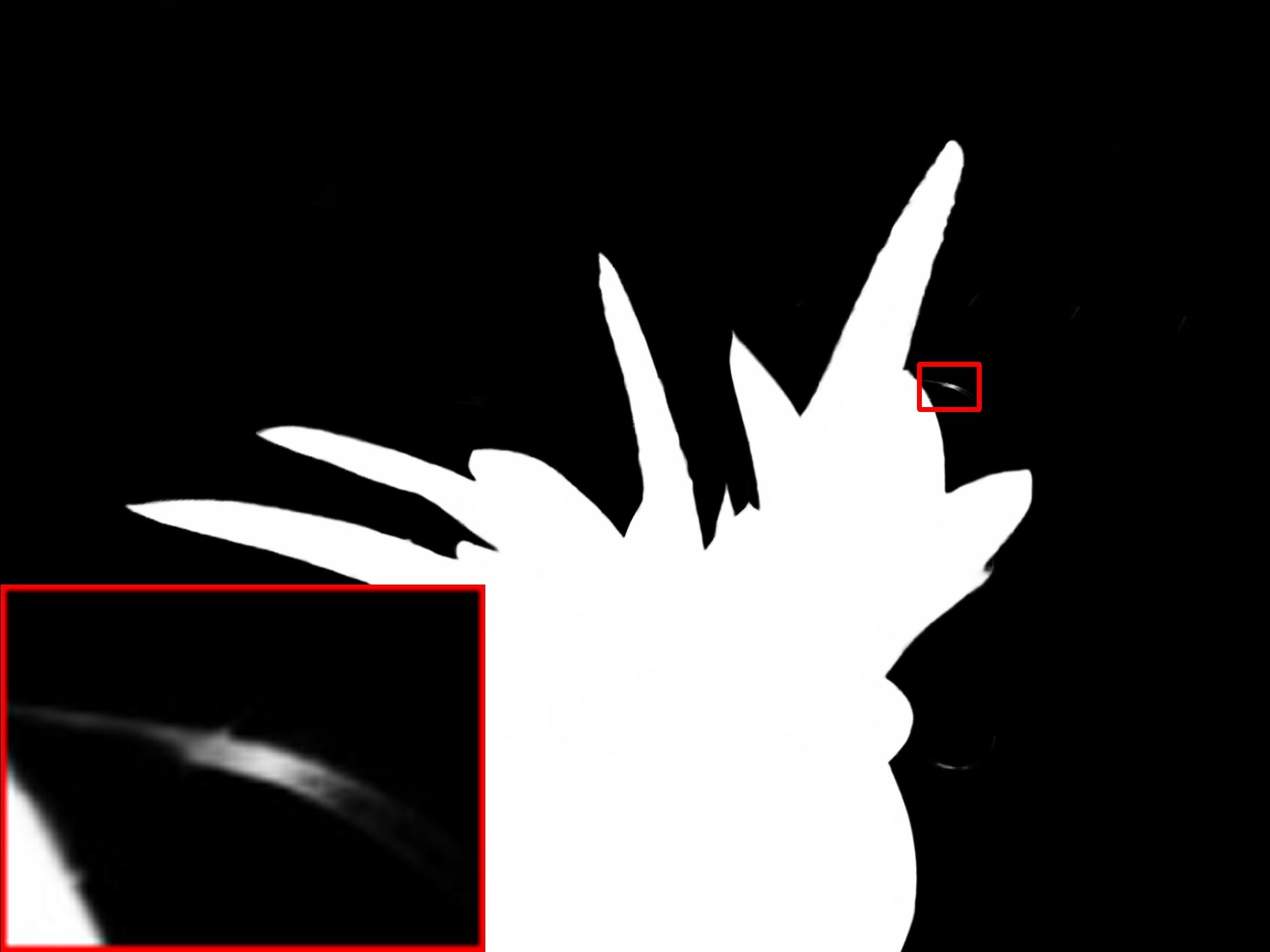} &
     \includegraphics[height=0.106\textwidth,width=0.142\textwidth]{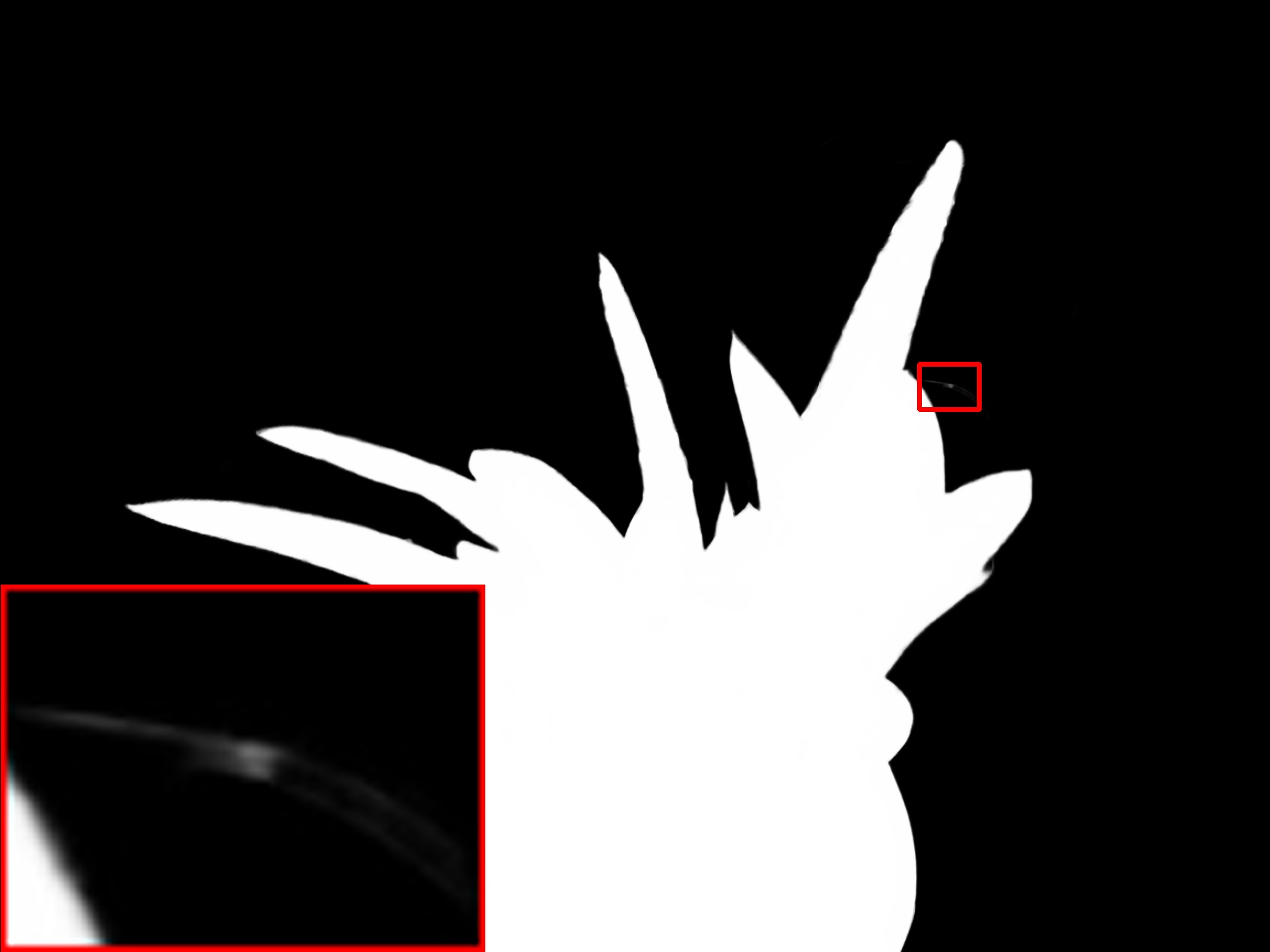} &
     \includegraphics[height=0.106\textwidth,width=0.142\textwidth]{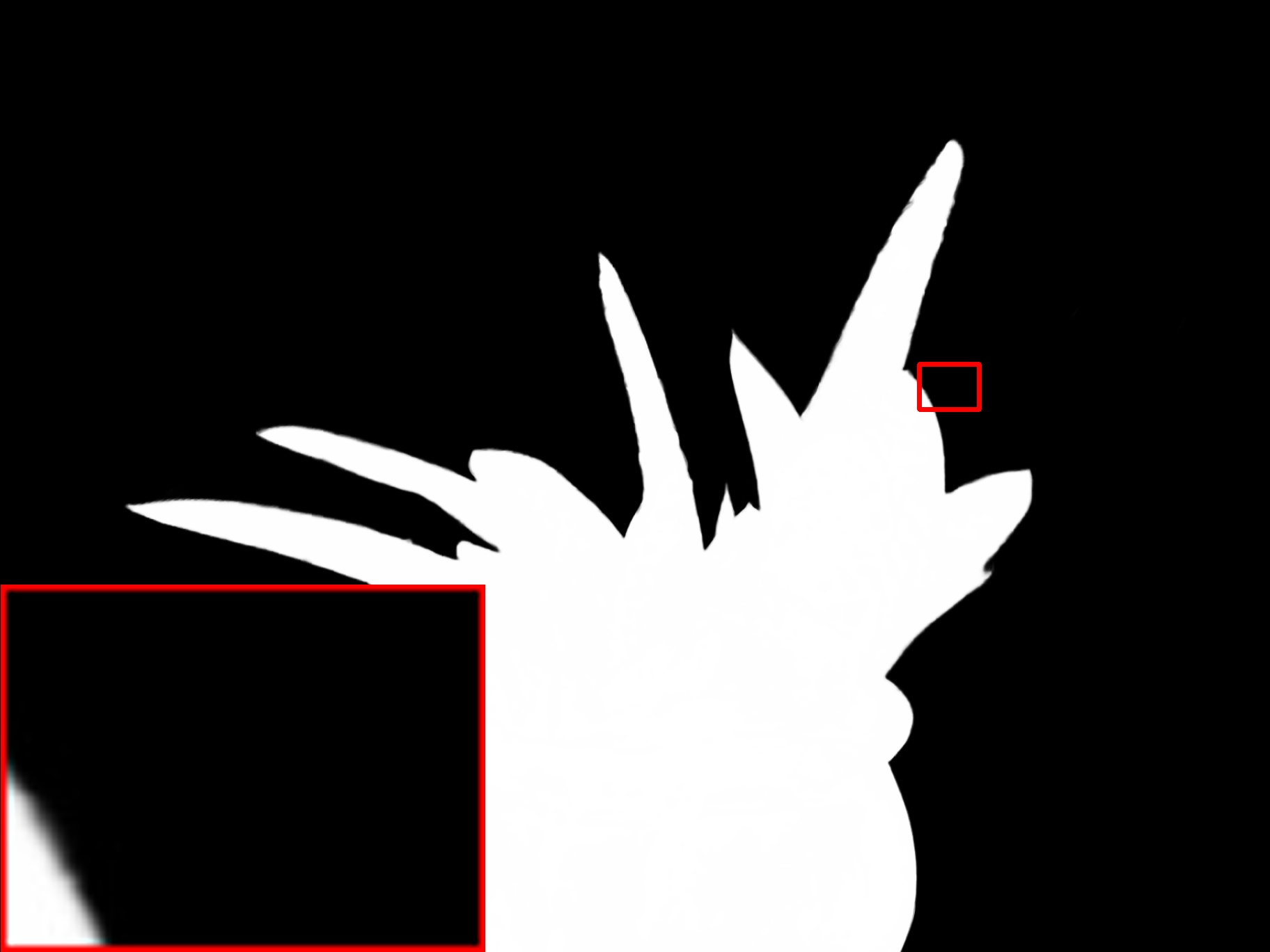} &
     \includegraphics[height=0.106\textwidth,width=0.142\textwidth]{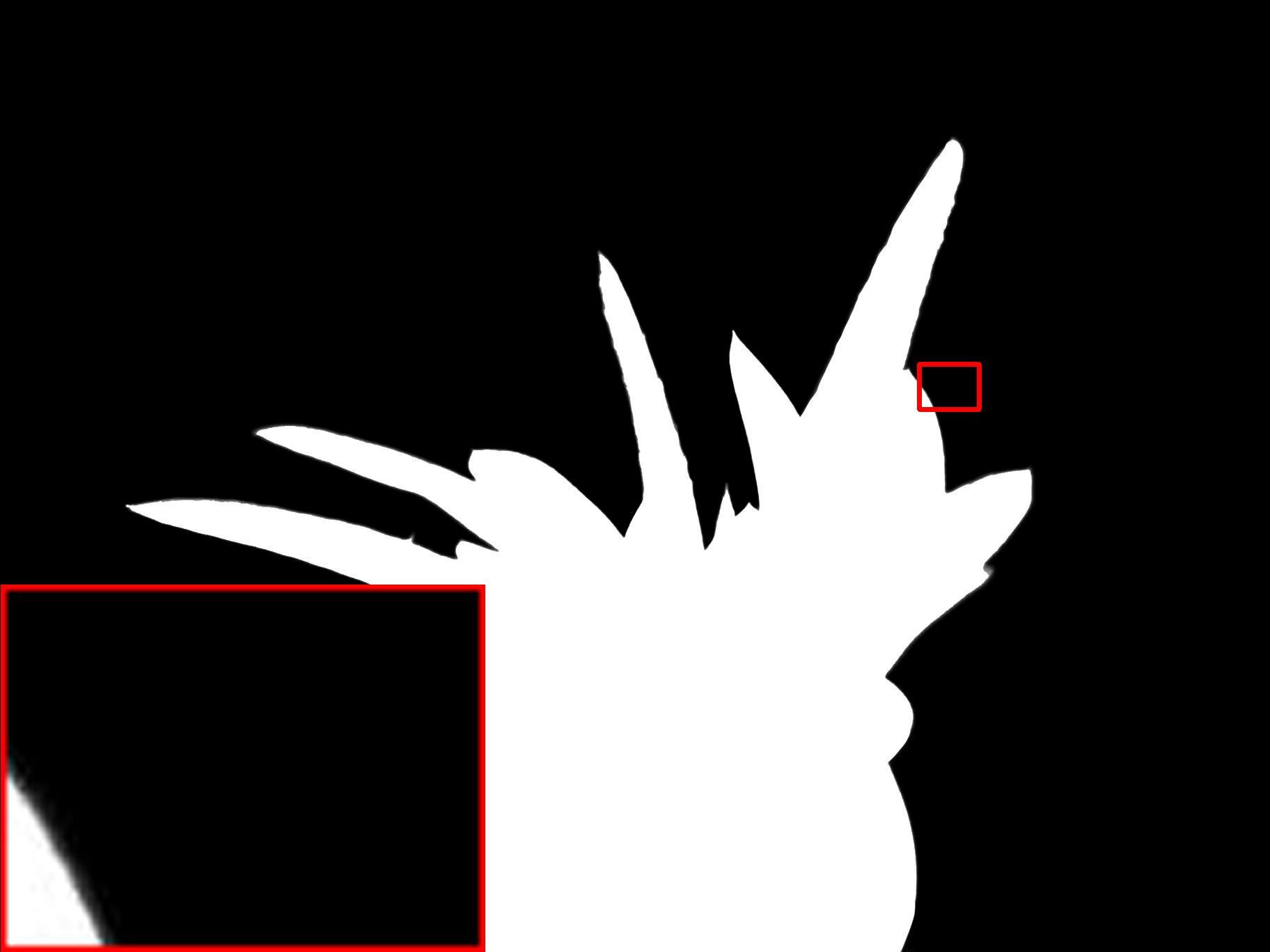}
     \\
     \includegraphics[trim={0 2.55cm 0 14cm},clip,width=0.142\textwidth]{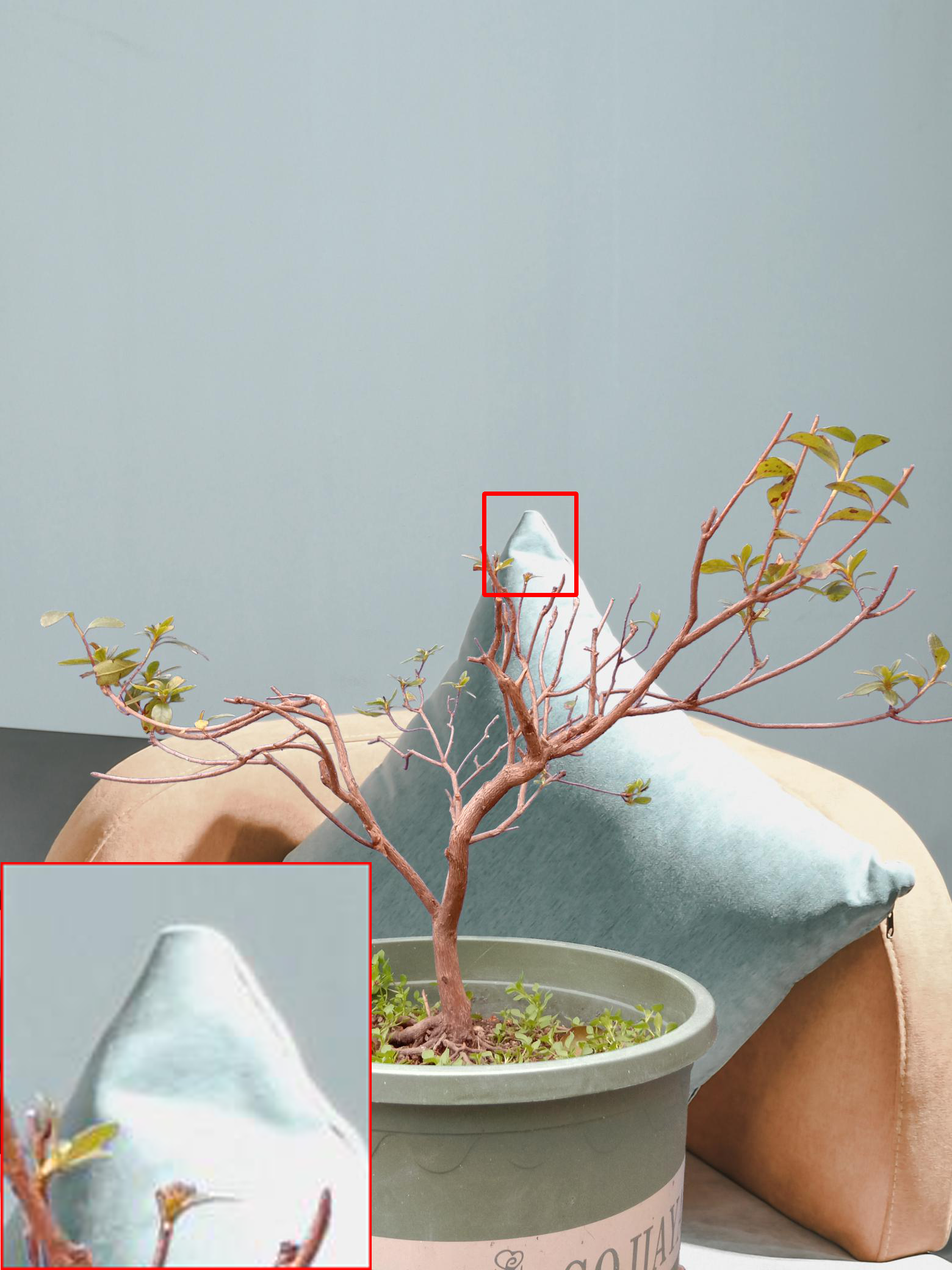} &
     \includegraphics[trim={0 2.55cm 0 14cm},clip,width=0.142\textwidth]{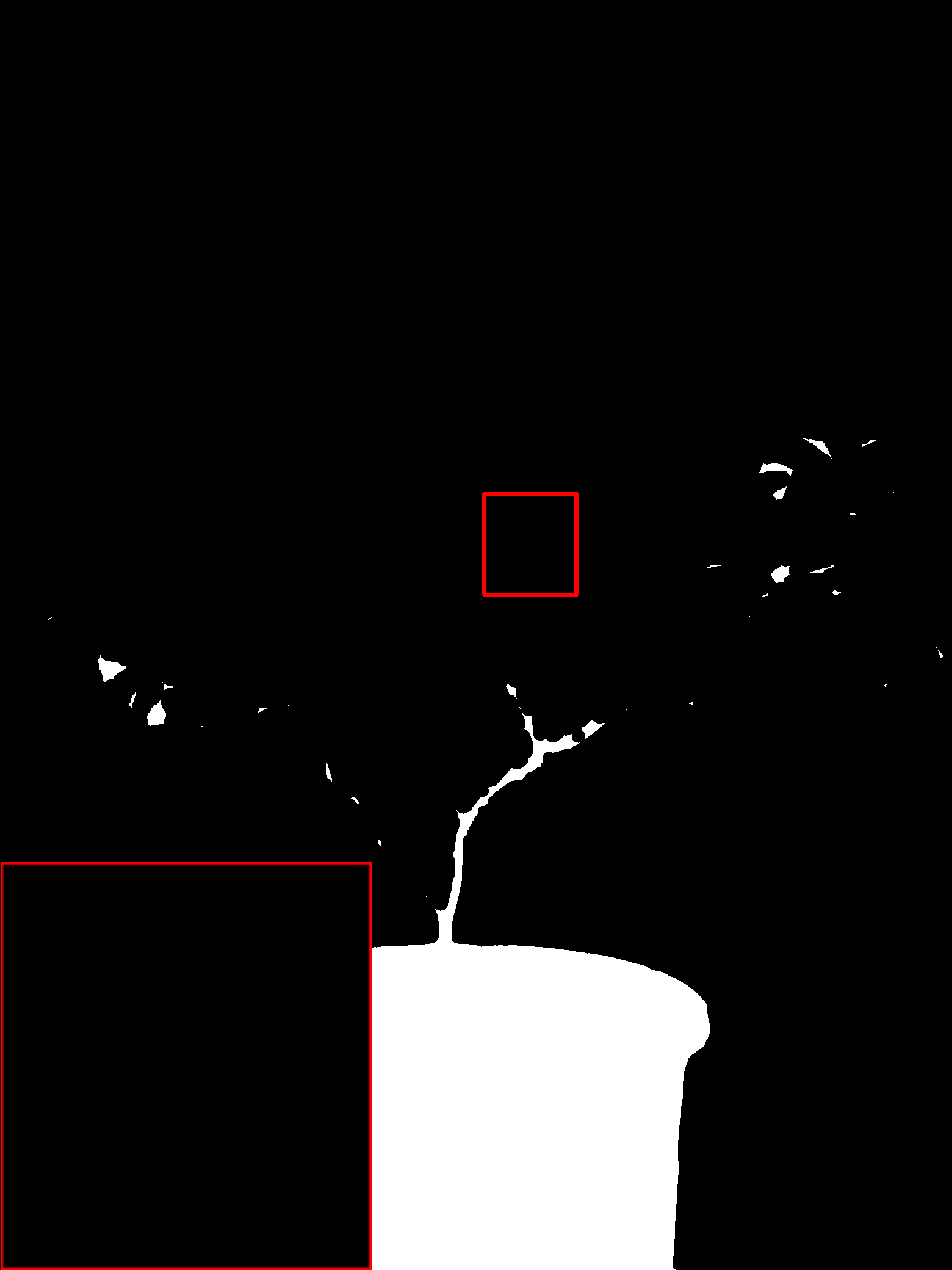} &
     \includegraphics[trim={0 2.55cm 0 14cm},clip,width=0.142\textwidth]{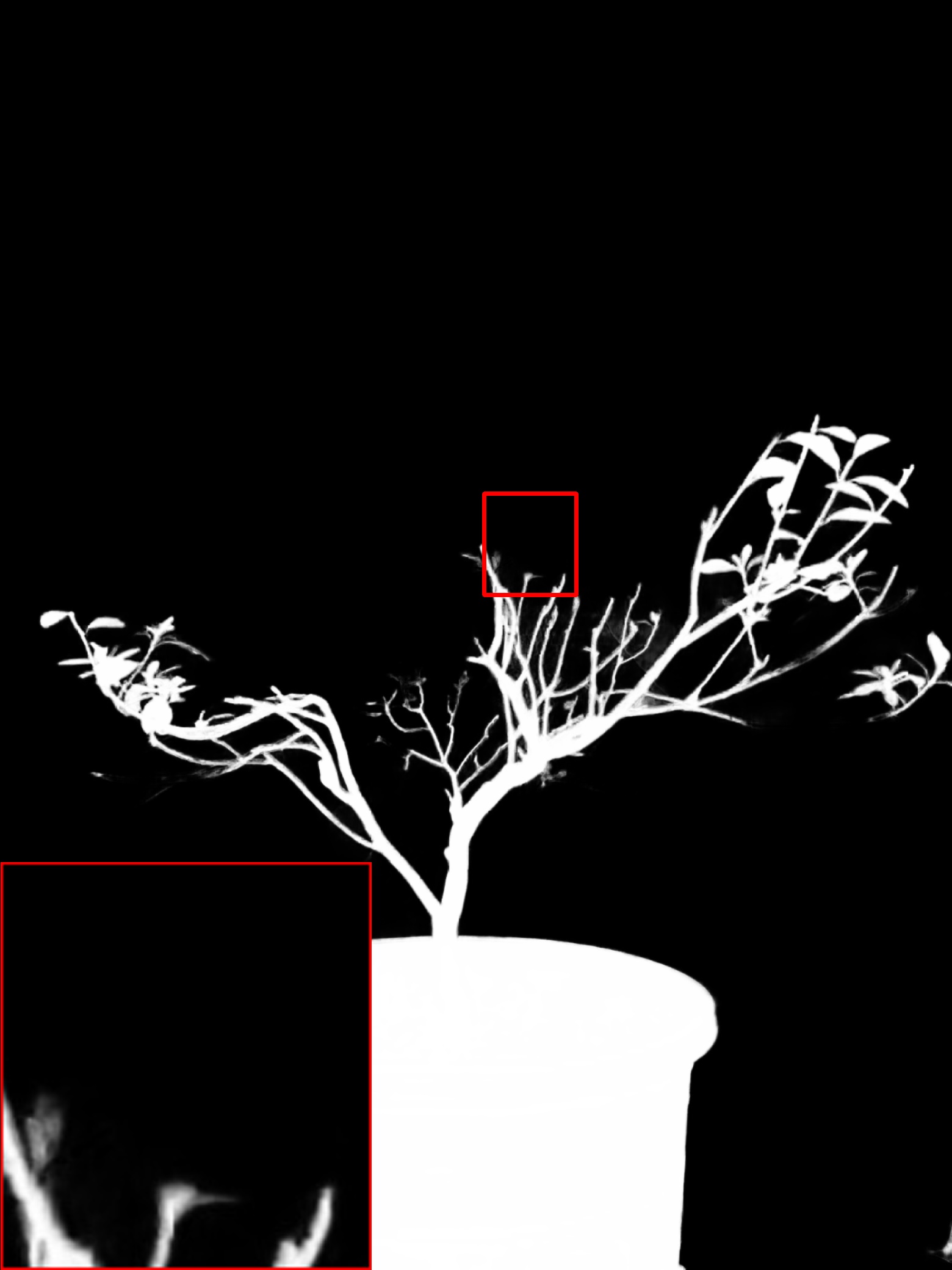} &
     \includegraphics[trim={0 2.55cm 0 14cm},clip,width=0.142\textwidth]{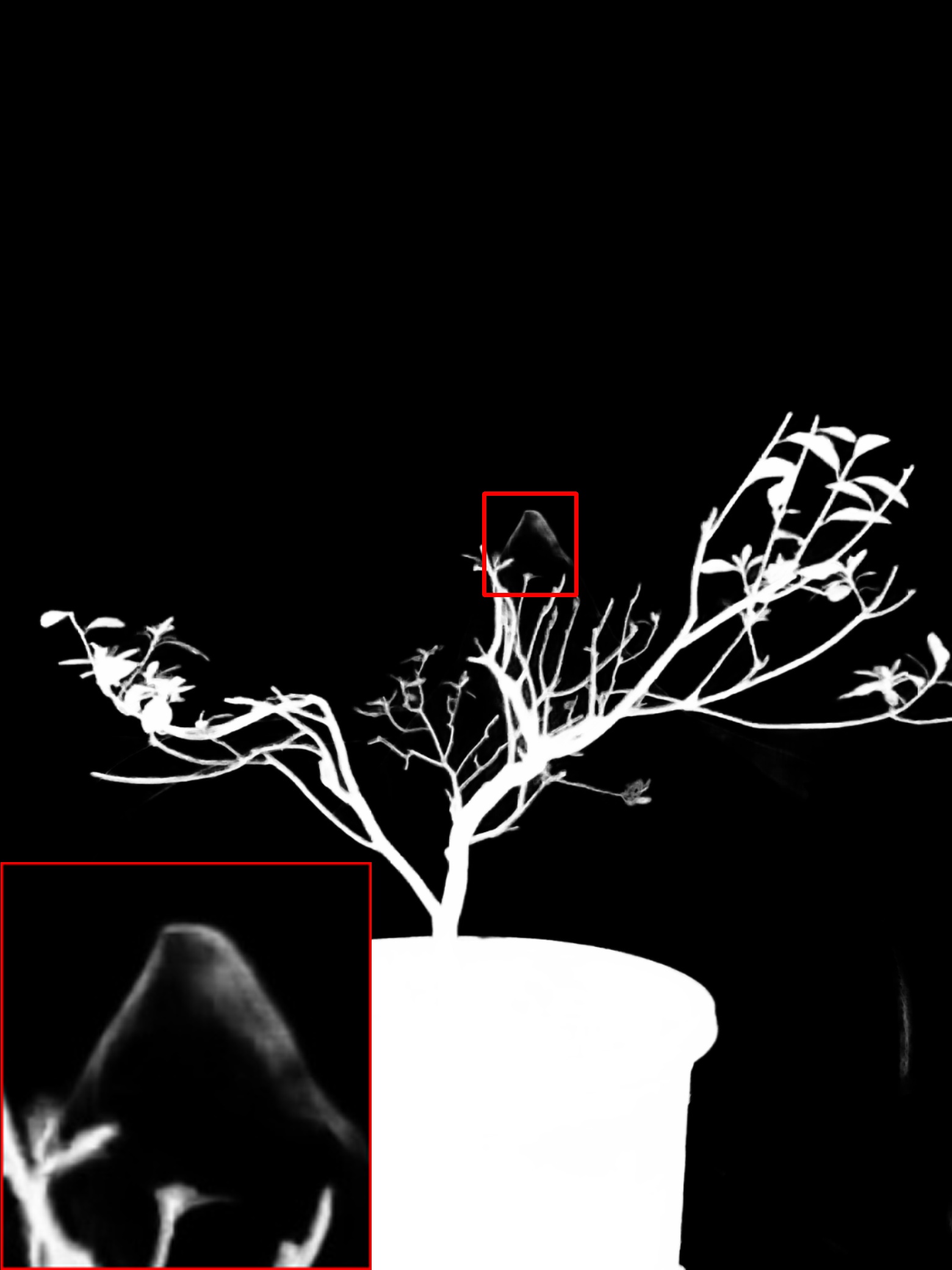} &
     \includegraphics[trim={0 2.55cm 0 14cm},clip,width=0.142\textwidth]{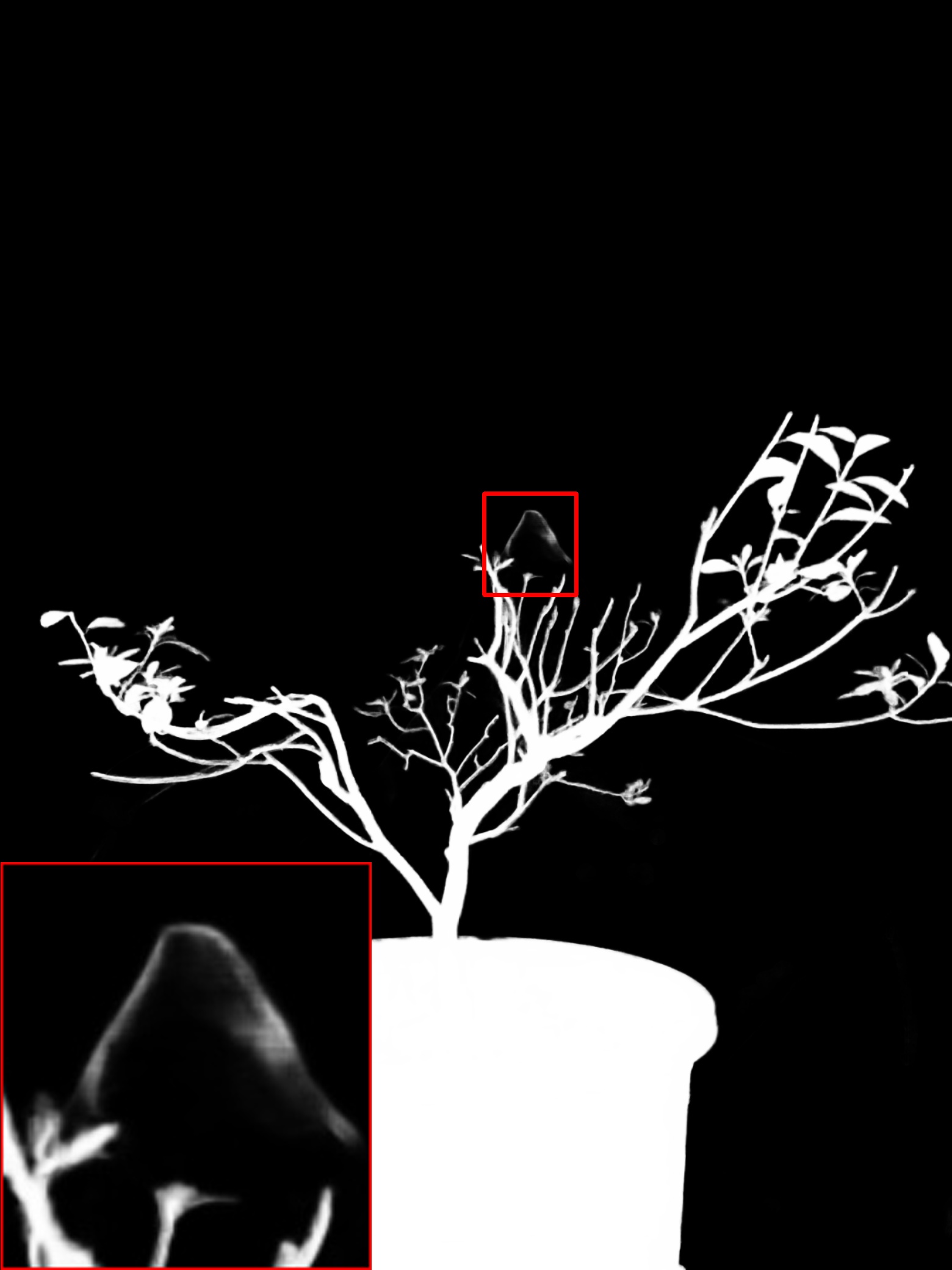} &
     \includegraphics[trim={0 2.55cm 0 14cm},clip,width=0.142\textwidth]{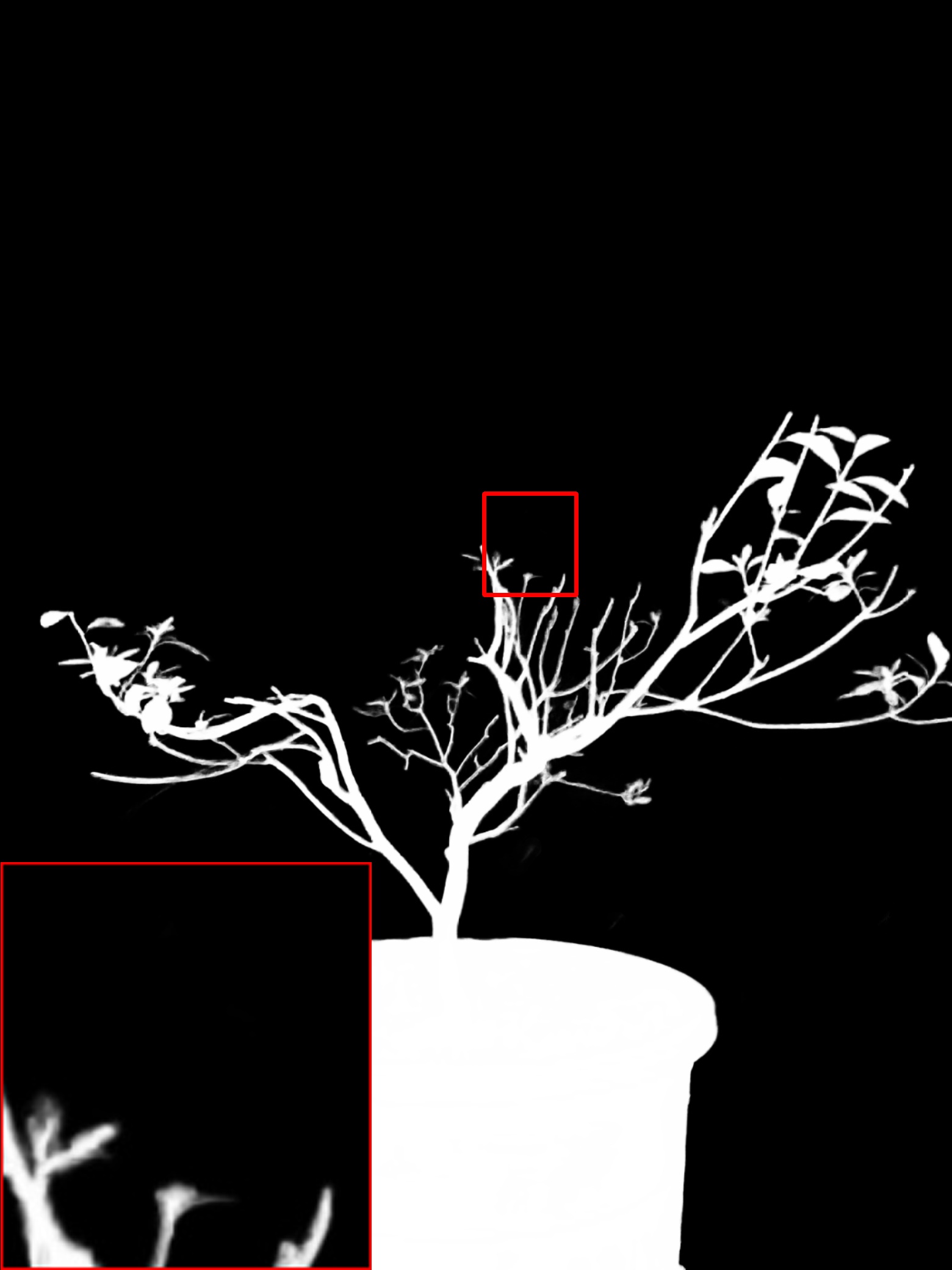} &
     \includegraphics[trim={0 2.55cm 0 14cm},clip,width=0.142\textwidth]{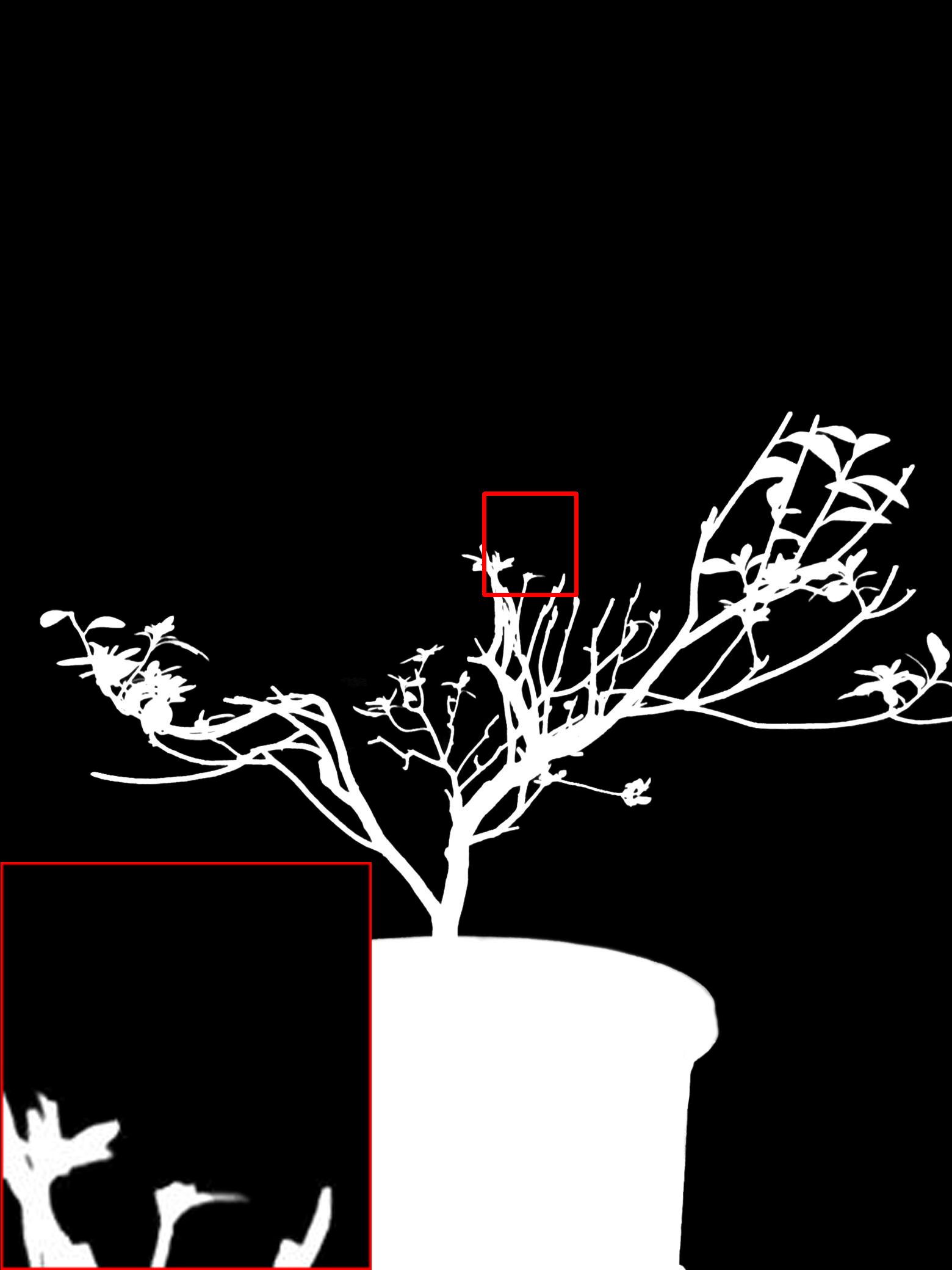}
     \\
     Image &
     Guidance &
     MGMatting &
     Ours(RASR) &
     Ours(RASR/IG) &
     Ours(RASR/IG/LD) &
     GT
  \end{tabular}
  
  \caption{The visual comparison results among different methods on Plant-Mat. RASR: real-world adaptive semantic representation. IG: with our IGDR module. LD: auxiliary learning with background line detection as Task 4.}
  \label{fig:plant_vis}
  \vspace{-1em} 
\end{figure*}

We evaluate our mask-guided method on the RWP~\cite{mgm_ref}, PPM-100~\cite{modnet_ref}, AM-2k~\cite{AM-2k}, AIM-500~\cite{aim}, and the proposed Plant-Mat benchmarks.  The RWP benchmark officially and publicly provides 636 real-world portraits with matting ground truth and coarse guidance masks in various scenes. We adhere to its original evaluation protocol, which includes metrics for both the entire image and detailed regions. The PPM-100~\cite{modnet_ref} provides 100 high-resolution real-world portraits with matting ground truth for evaluation. The AM-2k~\cite{AM-2k} provides 200 real-world animal images with matting ground truth for evaluation. The AIM-500~\cite{aim} provides 500 images with matting ground truth for evaluation, containing various real-world objects. We follow the evaluation protocol in the mask-guided method MG-wild~\cite{mgw} on  AIM-500~\cite{aim}, AM-2k~\cite{AM-2k}, and PPM-100~\cite{modnet_ref}. Our Plant-Mat benchmark provides 130 plant images, which can evaluate mask-guided methods for objects with  diverse and complex structures under clear backgrounds. To generate the coarse mask guidance for evaluation, we binarize the alpha matte of Plant-Mat with a threshold of 0.95 and then erode the binary mask with a 20$\times$20 kernel.

\textbf{Evaluation}. We follow previous mask-guided methods to evaluate the results by Sum of Absolute Differences (SAD), Mean Squared Error (MSE), Gradient error (Grad), and Connectivity error (Conn) using the official evaluation code~\cite{mgm_ref}.  Since mask-guided matting has shown great practicality compared to traditional trimap-based or guidance-free methods, we focus on the comparison of  mask-guided methods like MGMatting~\cite{mgm_ref} or MG-Wild~\cite{mgw}, but also report metrics for trimap-based methods~\cite{deepmatting,gca,indexnet,context} and a trimap-free method~\cite{lfm_ref,AM-2k,modnet_ref,rvm,shm,rethink_p3m}  as a reference. We denote our baseline model trained with only matting data as ``Matting only''. The model learning real-world adaptive semantic representation from Tasks 2 and 3 is denoted as ``RASR''. Our inconsistency-guided detail regularization (IGDR) module is denoted as ``IG''. The model trained with background line detection is denoted as ``LD''.

\subsection{Qualitative and quantitative comparisons}

As is shown in Tab.~\ref{tab:rwp},~\ref{tab:aim},~\ref{tab:am}, and ~\ref{tab:ppm}, our method outperforms SOTA mask-guided methods such as MGMatting~\cite{mgm_ref} and MG-Wild~\cite{mgw} on real-world matting benchmarks significantly. Specifically, on AIM-500~\cite{aim} with diverse real-world objects and scenes, our method, learning multiple auxiliary representations, outperforms the mask-guided method MG-Wild that solely learns its matting representation with the Mean Teacher mechanism for real-world generalization. Furthermore, our method significantly outperforms the SOTA MGMatting  employing  strong data augmentation proposed in~\cite{context} for real-world adaptation, on other high-quality real-world benchmarks including PPM-100~\cite{modnet_ref}, RWP~\cite{mgm_ref}, and AM-2k~\cite{AM-2k}. As for our Plant-Mat benchmark with complex object structures and clear backgrounds, our method outperforms MGMatting by a large margin.  For qualitative comparison  in Fig.~\ref{fig:illu_vis} and~\ref{fig:CompVis}, our method adapts to diverse and complex real-world scenes and objects,  performs better for objects with real shadows (Row 1, Fig.~\ref{fig:illu_vis} and Row 4, Fig.~\ref{fig:CompVis}) and complex structures like elongated cactus spines (Row 3, Fig.~\ref{fig:illu_vis}),  regularizes low-level detail refinement (Row 2, Fig.~\ref{fig:illu_vis}), and suppresses interference of background lines or textures better (Row 3, Fig.~\ref{fig:illu_vis} and Row 3, Fig.~\ref{fig:CompVis}). 

\subsection{Ablation analysis}
\textbf{Effectiveness of ``RASR'' and our IGDR module.} With our real-world adaptive semantic representation learned from more diverse and complex real-world segmentation data, ``RASR'' achieves 4.6 and 14.3 SAD improvements on ``Matting only'' for  PPM-100 (Tab.~\ref{tab:ppm_ablation}) and Plant-Mat (Tab.~\ref{tab:plant}), respectively. Besides metric improvements, ``RASR''  predicts better on real-world shadows (the right leg in Row 1, Fig.~\ref{fig:FinalVis}) and complex structures like elongated and irregular branches (Row 1 and 3, Fig.~\ref{fig:plant_vis}). Utilizing inconsistency between matting and semantic representations learned from ``RASR'', our inconsistency-guided detail regularization ``IG'' achieves 3.1 and 14.5 SAD improvements for  PPM-100 and Plant-Mat, respectively. As shown in Row 2, Fig.~\ref{fig:FinalVis}, ``IG'' regularizes detail refinement and avoids overfitting on low-level details in the transparent SARI on a completely opaque body.
\begin{figure}[ht]
  \centering
  \footnotesize
  \setlength{\tabcolsep}{0.5pt} 
  
  \begin{tabular}{cccc}
    \includegraphics[width=0.15\textwidth]{Figures/Final_Vis/box/4_im_03487_input.jpg} &
     \includegraphics[width=0.15\textwidth]{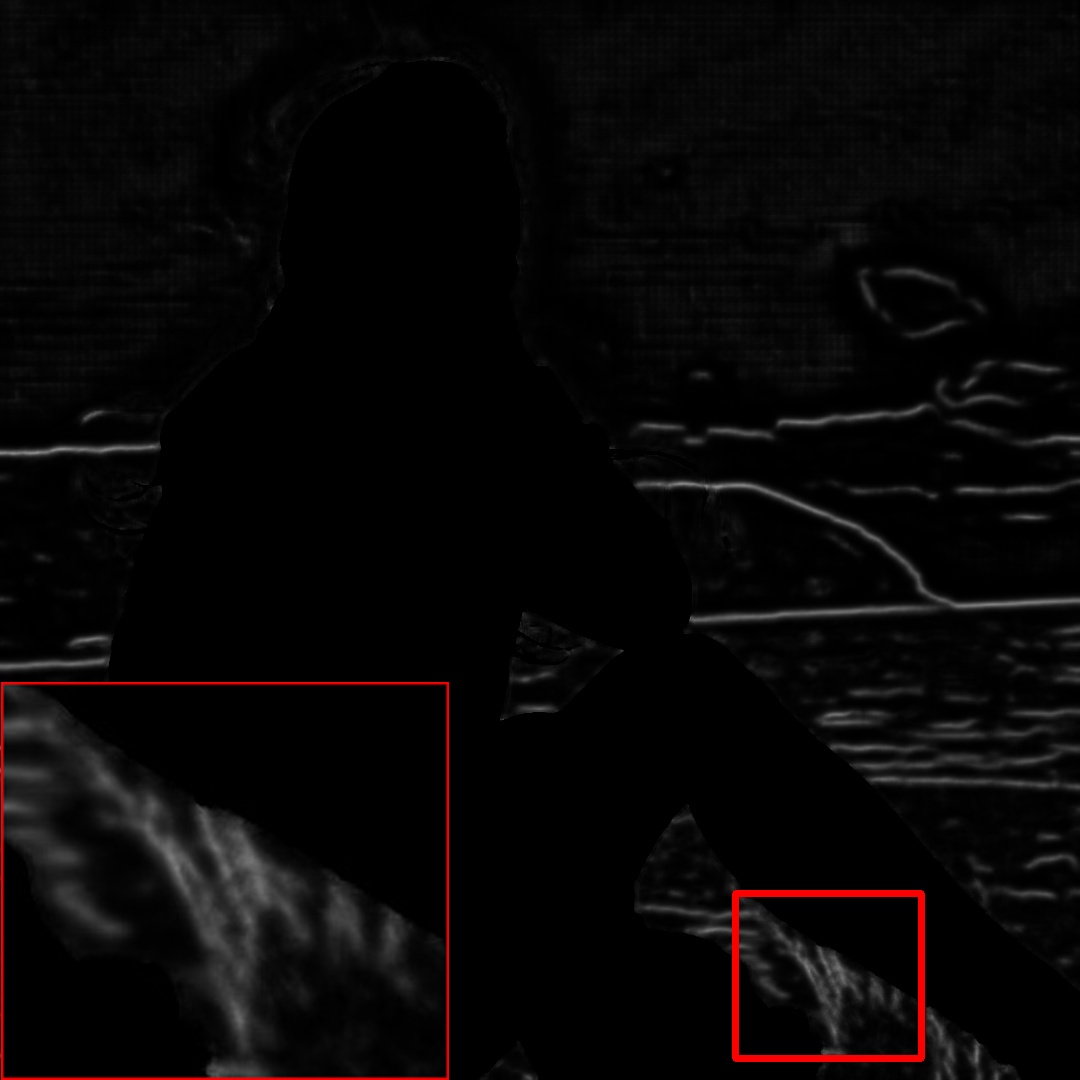} &
     \includegraphics[width=0.15\textwidth]{Figures/Final_Vis/box/4_line_03487_input.jpg}
     \\
     \includegraphics[height=0.1125\textwidth,width=0.15\textwidth]{Figures/plant_vis/box/2_im_plant_21_s.png} &
     \includegraphics[height=0.1125\textwidth,width=0.15\textwidth]{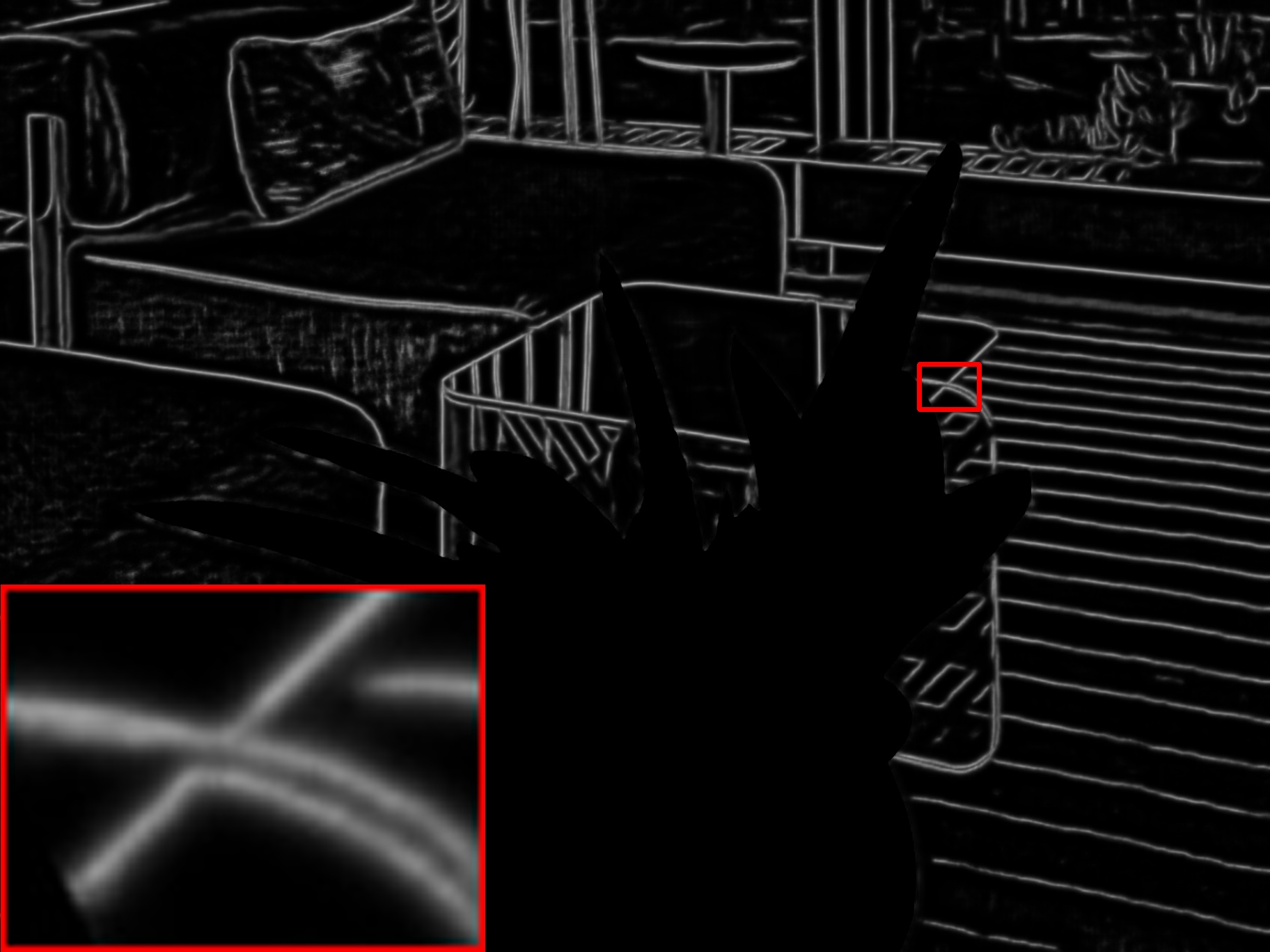} &
     \includegraphics[height=0.1125\textwidth,width=0.15\textwidth]{Figures/plant_vis/box/2_line_plant_21.jpg}
     \\
     Image &
     BG line prediction &
     Alpha prediction
  \end{tabular}
  \caption{Visualization for predictions of the background line and matting heads in our multi-task model. The auxiliary background line detection task learns representations that are aware of background lines, contributing to our matting predictions by suppressing background interference.}
  \label{fig:visline}
\end{figure}

\textbf{Effectiveness of our background line detection auxiliary task.} Learning with our novel background line detection auxiliary task, our model learns a discriminative representation to suppress background interference. With our background line detection task, ``LD''  improves 1.4 SAD and achieves 30.9 SAD on the PPM-100 benchmark (Tab.~\ref{tab:ppm_ablation}). For Plant-Mat with complex foreground structures over clear background lines, our ``LD'' performs better and  achieves 7.2 SAD improvements. For qualitative results, we use red boxes in Fig.~\ref{fig:FinalVis} and~\ref{fig:plant_vis} to zoom in on the effect of one place of background lines and textures in an image. For a real-world portrait in Row 3, Fig.~\ref{fig:FinalVis}, both MGMatting and our models without ``LD'' suffer from interference from background lines or textures of weeds, while our model with ``LD'' suppresses the background interference and predicts the foreground well. For a plant in Row 2, Fig.~\ref{fig:plant_vis}, both MGMatting and our models without ``LD'' suffer from background interference from the glass of a table, while ``LD'' also suppresses the interference. We visualize predictions of corresponding background lines in Fig.~\ref{fig:visline}. The predictions of background lines in Fig.~\ref{fig:visline} indicate that ``LD'' learns better representation to distinguish  the background lines or textures, which helps our model suppress  background interference. For plants in Row 3, Fig.~\ref{fig:plant_vis}, our model with ``LD'' simultaneously predicts the complex foreground structures and suppresses the interference from clear background textures well. More details can be found in the supplemental material.

\section{Conclusion}

In this paper, we propose a novel matting framework and model that learn different and effective representations through auxiliary learning and  adopt a novel inconsistency-guided detail regularization, to address challenges in mask-guided matting. By introducing auxiliary semantic segmentation and edge detection tasks and leveraging more accessible coarse segmentation annotations on real-world data, our model acquires a superior real-world adaptive semantic representation alongside matting representation, enabling it to adapt to complex real-world objects and scenes.  Utilizing the inconsistency between matting representation and semantic representation, our IGDR module regularizes the refinement of low-level details effectively. With our novel background line detection auxiliary task, our model learns discriminative representation to suppress background interference. In addition, we propose a plant matting dataset with complex object structures under proper and clear backgrounds, high-quality annotations, and look-natural composition to evaluate mask-guided matting methods. The quantitative and qualitative results on both established real-world matting  benchmarks and our Plant-Mat demonstrate the superiority of our proposed method.

\section*{Acknowledgments}
This paper is supported by NSFC, China (No. 62176155), Shanghai Municipal Science and Technology Major Project, China (2021SHZDZX0102), and OPPO Research Fund.


%

\bibliographystyle{IEEEtran}
\bibliography{aaai24}
\vspace{-5.5em}


\vfill

\end{document}